\documentclass{article}

\usepackage{microtype}
\usepackage{graphicx}
\usepackage{subfigure}
\usepackage{booktabs} 

\usepackage{hyperref}

\usepackage[accepted]{icml2023}

\usepackage{wrapfig}
\usepackage{csquotes}

\usepackage{amsmath}
\usepackage{amssymb}
\usepackage{mathtools}
\usepackage{amsthm}

\usepackage[capitalize,noabbrev]{cleveref}

\theoremstyle{plain}
\newtheorem{theorem}{Theorem}[section]
\newtheorem{assumption}{Assumption}[section]

\theoremstyle{definition}
\newtheorem{definition}{Definition}[section]
\theoremstyle{remark}

\usepackage[textsize=tiny]{todonotes}

\icmltitlerunning{Offline Equilibrium Finding}

\begin{document}

\twocolumn[
\icmltitle{Offline Equilibrium Finding}




\begin{icmlauthorlist}
\icmlauthor{Shuxin Li}{yyy}
\icmlauthor{Xinrun Wang}{yyy}
\icmlauthor{Youzhi Zhang}{sch}
\icmlauthor{Jakub \v{C}ern\'{y}}{yyy}
\icmlauthor{Pengdeng Li}{yyy}
\icmlauthor{Hau Chan}{comp}
\icmlauthor{Bo An}{yyy}
\end{icmlauthorlist}

\icmlaffiliation{yyy}{School of Computer Science and Engineering, Nanyang Technological University, Singapore}
\icmlaffiliation{comp}{University of Nebraska-Lincoln, USA}
\icmlaffiliation{sch}{Centre for Artificial Intelligence and Robotics, Hong Kong Institute of Science \& Innovation, Chinese Academy of Sciences, China}

\icmlcorrespondingauthor{Xinrun Wang}{xinrun.wang@ntu.edu.sg}

\icmlkeywords{Machine Learning, ICML}

\vskip 0.3in
]



\printAffiliationsAndNotice{}  

\begin{abstract}

Offline reinforcement learning (offline RL) is an emerging field that has recently begun gaining attention across various application domains due to its ability to learn strategies from earlier collected datasets. Offline RL proved very successful, paving a path to solving previously intractable real-world problems, and we aim to generalize this paradigm to a multiplayer-game setting. 
To this end, we introduce a problem of \textit{offline equilibrium finding} (OEF) and construct multiple types of datasets across a wide range of games using several established methods. To solve the OEF problem, we design a model-based framework that can directly apply any online equilibrium finding algorithm to the OEF setting while making minimal changes. The three most prominent contemporary online equilibrium finding algorithms are adapted to the context of OEF, creating three model-based variants: OEF-PSRO and OEF-CFR, which generalize the widely-used algorithms PSRO and Deep CFR to compute Nash equilibria (NEs), and OEF-JPSRO, which generalizes the JPSRO to calculate (Coarse) Correlated equilibria ((C)CEs). We also combine the behavior cloning policy with the model-based policy to further improve the performance and provide a theoretical guarantee of the solution quality.
Extensive experimental results demonstrate the superiority of our approach over offline RL algorithms and the importance of using model-based methods for OEF problems. We hope our work will contribute to advancing research in large-scale equilibrium finding.
\end{abstract}

\section{Introduction}
Game theory provides a universal framework for modeling interactions among cooperative and competitive players ~\cite{shoham2008multiagent}. The canonical solution concept is Nash equilibrium (NE), describing a situation when no player increases their utility by unilaterally deviating. However, computing NE in two-player or multi-player general-sum games is PPAD-complete~\cite{daskalakis2006complexity,chen2006settling}, which makes solving games both exactly and approximately difficult. The situation remains non-trivial even in two-player zero-sum games, no matter whether the players may perceive the state of the game perfectly (e.g., in Go~\cite{silver2016mastering}) or imperfectly (e.g., in poker~\cite{brown2018superhuman} or StarCraft II~\cite{vinyals2019grandmaster}). 
In recent years, learning algorithms have demonstrated their superiority in solving large-scale imperfect-information extensive-form games over traditional optimization methods, including linear or nonlinear programs. 
The most successful learning algorithms belong either to the line of research on counterfactual regret minimization (CFR)~\cite{brown2018superhuman}, or policy space response oracles (PSRO)~\cite{lanctot2017unified}. CFR is an iterative algorithm approximating NEs using repeated self-play.
Several sampling-based CFR variants~\cite{lanctot2009monte,gibson2012generalized} were proposed to solve large games efficiently. To scale up to even larger games, CFR could be embedded with neural network function approximation~\cite{brown2019deep,steinberger2019single,li2019double,agarwal2020optimistic}. The other algorithm, PSRO, generalizes the double oracle method~\cite{mcmahan2003planning,bosansky2014exact} by incorporating (deep) reinforcement learning (RL) methods as a best-response oracle~\cite{lanctot2017unified,muller2019generalized}. The neural fictitious self-play (NFSP) can be seen as a special case of PSRO~\cite{heinrich2015fictitious}. Both CFR and PSRO achieved great performance, especially in the more challenging type of large-scale imperfect-information extensive-form games like poker~\cite{brown2018superhuman, mcaleer2020pipeline}. In this paper, we also explicitly focus on this type of game.

One of the components that played a critical role in the success of CFR and PSRO is the existence of \emph{efficient and accurate simulators}. A simulator serves as an environment that allows an agent to collect millions to billions of trajectories for the training process.
The simulator may be encoded using rules as in different poker variants~\cite{lanctot2019openspiel}, or a video-game suite like StarCraft II~\cite{vinyals2017starcraft}.
However, in many real-world games such as football~\cite{kurach2020google,tuyls2021game} or table tennis~\cite{ji2021model}, constructing a sufficiently accurate simulator may not be feasible because of a multitude of complex factors affecting the game-play. These factors include the relevant laws of physics, environmental conditions (e.g., wind speed), or physiological limits of (human) bodies rendering certain actions unattainable. Therefore, the football teams or the table tennis players may resort to watching previous matches to improve their strategies, which semantically corresponds to \emph{offline equilibrium finding} (OEF).
Recent years have witnessed several (often domain-specific) attempts to formalize offline learning in the context of games. For example, the StarCraft II Unplugged~\cite{mathieu2021starcraft} offers a dataset of human game-plays in this two-player zero-sum symmetric game. Some concurrent works~\cite {cui2022offline,zhong2022pessimistic} investigate the necessary properties of offline datasets of two-player zero-sum Markov games to successfully infer their NEs. 

However, neither work considers the significantly more challenging field of multi-player games. To this end, we propose a more general problem -- \emph{offline equilibrium finding} (OEF) -- which aims to find the equilibrium strategy of the underlying game given a fixed offline dataset collected by an unknown behavior strategy. It is a big challenge since it needs to build the relationship between an equilibrium strategy and an offline dataset. To solve this problem, we introduce an environment model as the intermediary between them.
More specifically, our main contributions include 
i) proposing a novel problem -- OEF, and constructing OEF datasets from widely accepted game domains using different behavior strategies;
ii) proposing a model-based framework that can generalize any online equilibrium finding algorithm to the OEF setting by introducing an environment model;
iii) adapting several existing online equilibrium finding algorithms to the OEF setting for computing different equilibrium solutions, and providing a minimal dataset coverage assumption for the model-based framework to converge to the equilibrium strategy;
iv) combining the behavior cloning technique to further improve the performance, and providing the guarantee of the solution quality of our OEF algorithm;
v) conducting extensive experiments to verify the effectiveness of our algorithms. The experimental results substantiate the superiority of our method over model-based and model-free offline RL algorithms and the effectiveness and necessity of the model-based method for solving the OEF problem.

\vspace{-5pt}
\section{Rationale Behind OEF}
\label{rationale}
\vspace{-3pt}
In order to highlight the rationale behind the introduction of the OEF problem, we begin by providing a motivating scenario of the OEF problem and then explain why current algorithms cannot work for the OEF problem. Finally, we introduce the OEF problem and its challenges. A detailed overview of related works is provided in Appendix~\ref{app:related_works}.

\begin{figure}[t]
\centering
\includegraphics[width=0.325\textwidth]{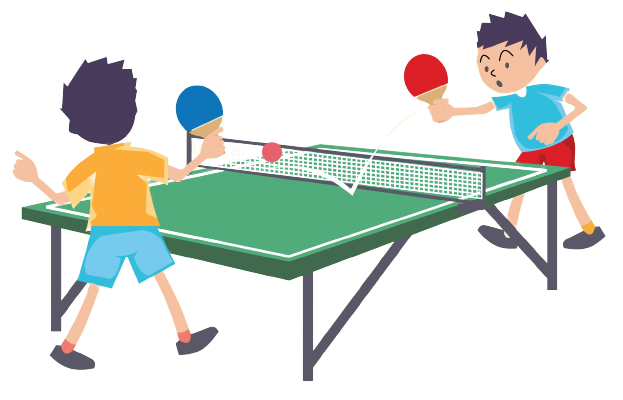}
\caption{The game of table tennis.}
\label{fig:motivation}
\vspace{-10pt}
\end{figure}

\textbf{Motivating Scenario.}  Assume that a table tennis player $A$ will play against player $B$ whom they never faced before (Figure \ref{fig:motivation}). 
What may player $A$ do to prepare for the match? Even though player $A$ knows the rules of table tennis, they remain unaware of specific circumstances of playing against player $B$, such as which moves or actions player $B$ prefers or their subjective payoff function. 
Without this detailed game information, self-play or other online equilibrium finding algorithms cannot work.
If player $A$ simply plays the best response strategy against player $B$'s previous strategy, the best response strategy may be exploited if player $B$ change his strategy.
Therefore, player $A$ has to \emph{watch the matches that player $B$ played against other players} to learn their style and compute the equilibrium strategy, which minimizes exploitation. This process corresponds to the proposed OEF methodology. 

\textbf{Why Naive Offline RL is not Enough?} 
From the above motivating scenario, the most straightforward method is the offline reinforcement learning (offline RL) algorithm. The offline RL aims to learn a good behavior policy from previously collected datasets which can achieve the highest utility for the agent \cite{levine2020offline}. However, the naive offline RL algorithm cannot solve the OEF problem since the offline RL can only maximize the utility of each player in the game independently which is not sufficient for computing the equilibrium strategy. In other words, if we use the offline RL algorithm to compute the optimal strategy for each player in the game, then these optimal strategies may be exploitable since every player can change their strategy instead of playing the behavior strategy computing from the offline dataset. To substantiate this claim, we conduct several offline RL algorithms under the OEF setting and these results show that the strategies computed by offline RL algorithms can be highly exploited. To further clarify the difference better the offline RL and the OEF, we prove that the dataset coverage assumption which is sufficient for computing the optimal strategy under the offline setting is not sufficient for computing the NE strategy under the OEF setting. The details of proof can be found in Appendix \ref{app:theoretical}. 

\textbf{Why Opponent Modeling (OM) \& Empirical Game-Theoretic Analysis (EGTA) is not Enough?}
Opponent modeling (OM) is used to predict the opponents' behavior strategies, in both single and multi-agent reinforcement learning~\cite{he2016opponent}. However, in the OEF setting, only predicting the opponents' behaviors is not enough since the opponents are not fixed but always best responding to the agent's strategy. 
The empirical game-theoretic analysis (EGTA) was proposed to reduce the complexity of large economic systems in electronic commerce~\cite{wellman2006methods}. It evolved into two main directions: strategic reasoning for simulation-based games~\cite{wellman2006methods} and evolutionary dynamical analysis of agent behavior inspired by evolutionary game theory~\cite{tuyls2018generalised}. 
The basic idea of EGTA is to use a sampled strategy to interact with a game simulator and estimate the empirical game from the simulation's results. It should consequently provide insights into the structure of the original game. 
However, in the OEF setting, only an offline dataset collected by an unknown strategy is available, while the game simulator and sampled strategy are not provided. Therefore, EGTA is not enough for computing the NE strategy in the OEF setting. More discussion can be found in Appendix \ref{app:FAQs}.

\textbf{Why Offline Equilibrium Finding?} In games with complex dynamics like table tennis games or football games~\cite{kurach2020google}, it is difficult to build a realistic simulator or learn the policy during playing the game. A remedy is to learn the policy from the historical game data. Therefore, we propose the \emph{offline equilibrium finding} (OEF) problem:
\begin{displayquote}
\emph{Given a fixed dataset $\mathcal{D}$ collected by an unknown behavior strategy $\sigma$, find an equilibrium strategy profile $\sigma^*$ of the underlying game.}
\end{displayquote}
The OEF problem is similar to the offline RL problem but poses several unique challenges: i) the canonical solution concept is the game-theoretic mixed NE, which requires an iterative procedure of computing best responses; ii) the games have at least two players who play against each other, which increases sensitivity to distribution shift and other uncertainties when compared with traditional offline RL; and iii) the distribution shifts of opponents' actions and the dynamic of the game are coupled, which brings difficulties to distinguish and address them.

\section{Preliminaries}

\subsection{Imperfect-Information Extensive-form Games}
An imperfect-information extensive-form game is represented as a tuple ($N, H, A, P, \mathcal{I}, u$)~\cite{shoham2008multiagent}, where $N = \{1,...,n\}$ is a set of players and $H$ is a set of histories (i.e., the possible action sequences). The empty sequence $\emptyset$ corresponds to a unique root node of a game tree included in $H$, and every prefix of a sequence in $H$ is also in $H$. $Z \subset H$ is the set of the terminal histories. $A(h) = \{a:(h,a) \in H\}$ is the set of available actions at any non-terminal history $h \in H$. $P$ is the player function. $P(h)$ is the player who takes an action at the history $h$, i.e., $P(h) \mapsto N \cup \{c\}$. $c$ denotes the ``chance player'', which represents stochastic events outside of the players' controls. If $P(h) = c$ then chance determines the action taken at history $h$. Information sets $\mathcal{I}_i$ form a partition over histories $h$ where player $i \in N$ takes action. Therefore, every information set $I_i \in \mathcal{I}_i$ corresponds to one decision point of player $i$ which means that $P(h_1) = P(h_2)$ and $A(h_1) = A(h_2)$ for any $h_1, h_2 \in I_i$. For convenience, we use $A(I_i)$ to represent the action set $A(h)$ and $P(I_i)$ to represent the player $P(h)$ for any $h \in I_i$. For $i \in N$, a utility function $u_i: Z \rightarrow \mathbb{R}$ specifies the payoff of player $i$ for every terminal history.

A player's behavior strategy $\sigma_i$ is a function mapping every information set of player $i$ to a probability distribution over $A(I_{i})$ and $\Sigma_{i}$ is the set of strategies for player $i$. A strategy profile $\sigma$ is a tuple of strategies, one for each player, ($\sigma_1, \sigma_2, ..., \sigma_n$), with $\sigma_{-i}$ referring to all the strategies in $\sigma$ except $\sigma_i$. Let $\pi^{\sigma}(h) = \prod_{i \in N \cup \{c\}}\pi^{\sigma}_i(h)$ be the probability of history $h$ occurring if all players choose actions according to $\sigma$. $\pi_i^{\sigma}(h)$ is the contribution of $i$ to this probability. Given a strategy profile $\sigma$, the value to player $i$ is the expected payoff of these resulting terminal nodes, $u_i(\sigma) = \sum_{z \in Z}\pi^\sigma(z)u_i(z)$. 

The canonical solution concept for imperfect information extensive form games is Nash equilibrium (NE). The strategy profile $\sigma^*$ forms an NE 
if 
\begin{equation}
    u_{i}(\sigma^*) = \max\nolimits_{\sigma_{i}'\in\Sigma_i} u_{i}(\sigma_{i}', \sigma_{-i}^*), \forall i\in N. \nonumber
\end{equation}
To measure of the distance between $\sigma$ and the NE strategy, we define $\textsc{NashConv}(\sigma)=\sum_{i\in N}\textsc{NashConv}_{i}(\sigma)$, where $\textsc{NashConv}_{i}(\sigma)=\max_{\sigma_{i}'} u_{i}(\sigma_{i}', \sigma_{-i}) - u_{i}(\sigma)$ for each player $i$. Except for the NE, some other solution concepts for extensive-form games exist conditional to different situations. For example, (Coarse) Correlated Equilibrium ((C)CE) is another popular solution concept for $n$-player general-sum extensive-form games. A CE strategy is a joint mixed strategy such that no player has the incentive to deviate from it. Coarse correlated equilibrium~\cite{moulin1978strategically} is a simpler solution concept that contains CE as a subset: NE $\subseteq$ CE $\subseteq$ CCE. A strategy profile is in CCE if no player wishes to deviate before receiving a recommended signal. Similar to NE, to measure the gap between a joint strategy $\sigma$ and the (C)CE, (C)CE Gap Sum is used to describe how close joint policies are to (C)CE~\cite{marris2021multi}. In this paper, we focus not only on the NE solution but also on the CCE solution. 


\subsection{Equilibrium Finding Algorithms}
\textbf{PSRO.} PSRO is initialized with a set of randomly-generated policies $\hat{\Sigma}_{i}$ for each player $i$. At each iteration of PSRO, a meta-game $M$ is built with all existing policies of players and then a meta-solver computes a meta-strategy, i.e., a distribution over policies of each player (e.g., Nash, $\alpha$-rank or uniform distributions). The joint meta-strategy for all players is denoted as $\boldmath{\alpha}$, where $\alpha_{i}(\sigma)$ is the probability that player $i$ takes $\sigma$ as their strategy. After that, an oracle computes at least one policy for each player, which is added to $\hat{\Sigma}_{i}$.
We note when computing the new policy for one player, all other players' policies and the meta-strategy are fixed, which corresponds to a single-player optimization problem and can be solved by DQN \cite{mnih2015human} or policy gradient reinforcement learning algorithms. NFSP can be seen as a special case of PSRO with uniform distributions as meta-strategies \cite{heinrich2015fictitious}. Joint Policy Space Response Oracles (JPSRO) is a novel extension to PSRO with full mixed joint policies to enable coordination among policies \cite{marris2021multi}. JPSRO is proven to converge to a (C)CE over joint policies in extensive-form games. 

\textbf{CFR.} CFR is a family of iterative algorithms for approximately solving large imperfect-information games. In every iteration, the whole game tree is traversed, and the counterfactual regret for every action $a$ in every information set $I$ is computed. The computation of the counterfactual regret value for one player's information set is related to the counterfactual value of the information set which is the expected value of the information set given that the player attempts to reach it. After traversing the game tree, to compute the strategy used for the next iteration, players use \textit{Regret Matching} to pick a distribution over actions in an information set proportional to the positive cumulative regret of those actions. In the next iteration, players use the new strategy to traverse the whole game tree. Then, we repeat these processes until convergence. Finally, in two-player zero-sum games, if both players' average regret is less than $\epsilon$, their average strategies over strategies in all iterations $(\overline{\sigma}^T_1,\overline{\sigma}^T_2)$ form a $2\epsilon$-equilibrium~\cite{waugh2009abstraction}. Most previous works focus on tabular CFR, where counterfactual values are stored in a table. Recent works adopt deep neural networks to approximate the counterfactual values and outperform their tabular counterparts~\cite{brown2019deep,steinberger2019single,li2019double,li2021cfr}.

\section{Algorithms for Offline Equilibrium Finding}

In real life, the offline dataset may be collected using an unknown behavior strategy. To simulate real-world cases, we focus on four types of datasets: expert dataset, learning dataset, random dataset, and hybrid dataset. Details of offline datasets, the relationship between the information set $I$ and game state $s_t$ can be found in Appendix~\ref{app:dateset}. Based on these offline datasets, we give the formal definition of the offline equilibrium finding problem. 
\begin{definition}[OEF]
Given a game's offline dataset $\mathcal{D} = (s_t, a, s_{t+1}, r_{t+1})$ where $s_t$ and $s_{t+1}$ refer to the game states, $a$ refers to the action played at $s_t$ and $r_{t+1}$ refers to the reward after performing action $a$ at $s_t$. The strategy used to collect the dataset $\mathcal{D}$ is unknown. The OEF problem is to find an approximate equilibrium strategy profile $\sigma^*$ that achieves a small gap between $\sigma^*$ and equilibrium, i.e., the $\textsc{NashConv}$ for NE and (C)CE Gap Sum for (C)CE, only based on $\mathcal{D}$. 
\end{definition}

Inspired by the offline RL \cite{chen2020bail,yu2020mopo}, there are two types of possible approaches for solving the OEF problem: model-free approach and model-based approach. The model-free approach needs to learn a policy \textit{directly} from the offline dataset. If the offline dataset is generated using the equilibrium strategy, we can easily learn the equilibrium strategy directly from the offline dataset by the behavior cloning technique. However, the strategy used to generate the offline dataset is unknown and when computing the equilibrium strategy, we cannot use the data of any two action tuples to determine which action tuple is closer to an equilibrium strategy since the equilibrium identification requires other action tuples to serve as references. Therefore, the model-free approach is not enough for solving the OEF problem since some data may be missing from the dataset and the model-free approach cannot measure the distance from the equilibrium strategy to guide the training.
Based on the above analysis, the model-based approach may be a more suitable choice. Therefore, to solve the OEF problem, we introduce a model-based framework, which can adapt any online equilibrium finding algorithm into the context of the offline setting. 


\subsection{Environment Model}
\label{env_model}
There are many model-based algorithms for offline single-agent RL. However, they cannot be directly applied to games for equilibrium findings. The main reason is their inherent reliance on the existence of no strategic opponents in the environment. It means that if we use these algorithms in the OEF setting, for any given opponent strategy, we would need to train a model for the player to compute the best response strategy. This process would be very time-consuming and highly computationally demanding. To sidestep this issue, we train an environment model for all players instead of using the single-agent model-based algorithm for every player. The trained environment model can capture the game information which is necessary for evaluating any action tuple. In this way, only one environment model needs to be trained and all players can share this environment model to compute the equilibrium strategy.

\begin{figure}[t]
    \vspace{4pt}
    \centering
    \includegraphics[width=0.7\linewidth]{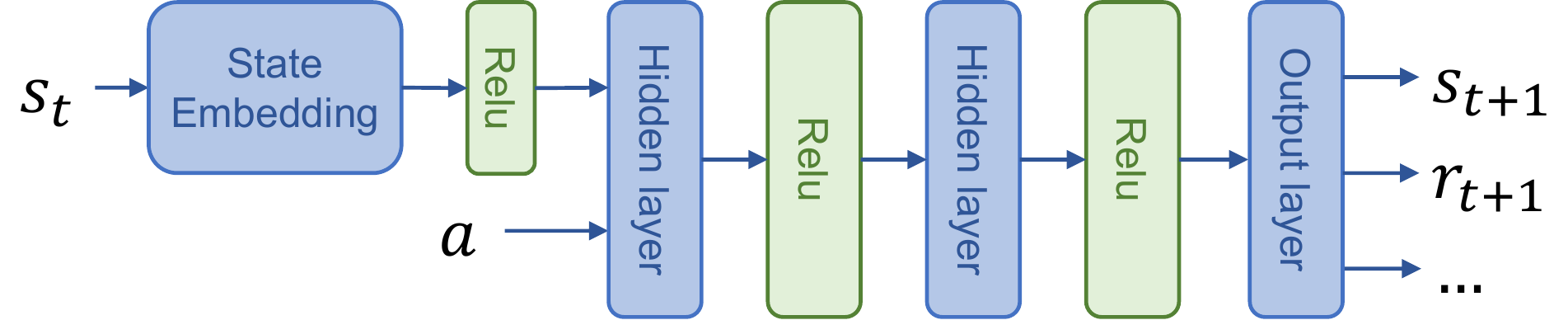}
    \caption{Environment model structure.}
    \label{fig:mb_model}
    \vspace{-14pt}
\end{figure}

Then we move to introduce how to train an environment model based on an OEF dataset. The deep neural network is adopted as the environment model due to its powerful generalization abilities. Figure \ref{fig:mb_model} shows the structure of the environment model. The environment model $E$ parameterized by $\theta_e$ takes the game state $s_t$ and action $a$ of the player, who played at the state $s_t$, as inputs and outputs the next game state $s_{t+1}$, the rewards $r_{t+1}$ of all players, and other information such as the next legal action set and whether the game ends. To deal with the chance player in the game, we predict whether the next state is a chance node or not, and if the next state is a chance node, then we sample an action according to the predicted legal action set to play. We take the stochastic gradient descent (SGD) as the optimizer to perform parameter updates. Any loss function that satisfies the conditions of Bregman divergence \citep{banerjee2005clustering} can be employed. We use the mean squared error loss which can be defined below. Finally, we train the environment model by performing mini-batch SGD iterations. 
\begin{align}
    \mathcal{L}_{env} &= \mathbb{E}_{D} [MSE((s_{t+1}, r_{t+1}), E(s_t, a;\theta_e))] \nonumber. 
\end{align}

\subsection{Model-Based Framework}
Once the environment model is well trained, the model can provide enough game information for equilibrium computation. Based on the trained environment model, we propose a general model-based framework, which can generalize any online equilibrium finding algorithm to the context of the OEF setting by replacing the actual environment with the well-trained environment model. To illustrate how to generalize existing online equilibrium finding algorithms to the context of the OEF setting, we instantiate three model-based algorithms: Offline Equilibrium Finding-Policy Space Response Oracles (OEF-PSRO) and Offline Equilibrium Finding-Deep CFR (OEF-CFR) algorithms, which generalize PSRO and Deep CFR to compute NEs, and Offline Equilibrium Finding-Joint Policy Space Response Oracles (OEF-JPSRO), which generalize JPSRO to compute (C)CEs.  


In PSRO or JPSRO, a meta-game is represented as an empirical game starting with a single policy (uniform random) and iteratively enlarged by adding new policies (oracles) that approximate the best responses to the meta-strategies of other players. It is clear that when computing the best response policy oracle, interactions with the environment are required to gain game information. In the OEF setting, only an offline dataset is provided, and directly applying PSRO or JPSRO is not feasible. In OEF-PSRO and OEF-JPSRO, we use the trained environment model to replace the actual environment to provide the game information. It is common knowledge that when computing the best response policy using DQN or other RL algorithms, the next state and reward based on the current state and the action are required. Our environment model can provide such information. The model can also offer additional information for approximating the missing entries in the meta-game matrix using the same manner. 
Deep CFR is a variant of CFR that uses neural networks to approximate counterfactual regret values and average strategies. During this approximation, the game tree has to be partially traversed to arrive at the counterfactual regret value. This process requires an environment to provide the necessary game information. Akin to OEF-PSRO, also in OEF-CFR, we use the trained environment model to replace the actual environment. During the traversal, the environment needs to identify the next game state and utility for the terminal game state, for which we employ our trained environment model. These algorithms are described in detail in Appendix~\ref{app:implementation}.

To analyze the performance of the model-based framework, we provide a minimal dataset coverage assumption over the offline dataset to guarantee to converge to the equilibrium strategy of the underlying game in the OEF setting, which is represented by the following theorem. The proof can be found in Appendix \ref{app:theoretical}.
\begin{assumption} (Uniform Coverage)
    For all state $s_t$ and all actions $a_t \in A(s_t)$, all state-action pairs $(s_t, a_t, s_{t+1})$ are covered by the dataset. 
\end{assumption}
\begin{theorem}
    The uniform coverage assumption over the offline dataset is the minimal dataset coverage assumption which is sufficient for our model-based algorithm to converge to the equilibrium strategy in the OEF setting.
\end{theorem}

\subsection{Combination Method: BC+MB}
The model-based framework has the convergence guarantee only on the uniform dataset coverage assumption. However, offline datasets may not satisfy the uniform dataset coverage. To make up for the deficiency of the model-based framework on these datasets, we introduce the behavior cloning (BC) technique and combine the behavior cloning technique and model-based framework by assigning different weights to trained policies using these two methods. The details of the BC technique can be found in Appendix~\ref{app:implementation}.

Let us introduce the combination method BC+MB, i.e., how to combine these two trained policies. Let $\alpha$ be the weight of BC policy. Then the weight of the MB policy is $1-\alpha$. We preset 11 weight assignment plans, i.e., $\alpha\in\{0, 0.1, 0.2, ..., 0.9, 1\}$. Next, we use these 11 weight assignment plans to combine these two policies to get a set of final policies. Finally, we test these combination policies in the real game to get the best final policy according to the exploitability value. 
\begin{algorithm}[t]
\caption{General Framework of OEF}
\label{oef}
\begin{algorithmic}[1]
\STATE {\bfseries Input:} an offline dataset $D$
\STATE Train an environment model $E$ based on $D$;
\STATE Train a policy ${\pi}^{mb}$ on $E$ using MB algorithm;
\STATE Train a policy ${\pi}^{bc}$ based on $D$ using BC technique;
\STATE Select   $\alpha$ based on the exploitability of the real game;
\STATE Combine ${\pi}^{bc}$ and ${\pi}^{mb}$ with   $\alpha$ to get policy ${\pi}$;
\STATE {\bfseries Output:} the policy $\pi$
\end{algorithmic}
\end{algorithm}
\begin{figure}[t]
\centering
\includegraphics[width=0.75\linewidth]{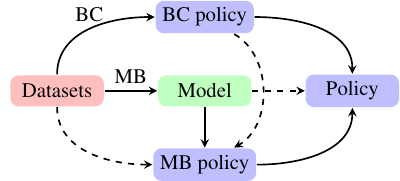}
\caption{The flow of OEF algorithms.}
\label{fig:oef_flow}
\vspace{-10pt}
\end{figure}
We present the general procedure of the BC+MB combination algorithm in Algorithm \ref{oef}. Given the offline dataset, we first train an environment model $E$ according to the method introduced in Section \ref{env_model}. Then based on the trained environment model, we can get the MB policy using the model-based algorithm which is the generalized algorithm from the online equilibrium finding algorithm under the model-based framework. To get the BC policy, we directly apply the behavior cloning technique on the offline dataset. Finally, we combine these two policies, i.e., the BC policy and the MB policy by assigning proper weights to these two policies to obtain the final policy. We represent the combination method as our OEF algorithm in the remainder of the paper.

To analyze the performance of the combination method BC+MB, we provide the guarantee of the solution quality for our combination method BC+MB, which is represented using the following theorem. The proof of the theorem and more analysis about the relationship between the dataset coverage and our algorithms can be found in Appendix~\ref{app:theoretical}.   
\begin{theorem}
Assuming that the environment model and the behavior cloning policy are well-trained, under the offline dataset $\mathcal{D}_{\sigma}$ generated using $\sigma$, BC+MB can get an equal or better strategy than $\sigma$.
\end{theorem}

Figure \ref{fig:oef_flow} shows the whole structure of our combination algorithm. Note that several dashed lines represent several unexplored options for future research: i) whether we can learn an MB policy with the regularization of the BC policy, as well as interact with the dataset, and ii) if we can use the learned model to get the proper weights when combining the two policies.

\section{Experiments}\label{sec:experiments}
To evaluate the performance of our algorithms, we conduct the following experiments: i) we conduct two offline RL algorithms on the OEF setting to verify their performance; ii) we conduct experiments on different offline datasets to evaluate the performance of our algorithms in computing NEs under the OEF setting; and iii) we conduct experiments on two three-player games to assess the performance of our algorithm in computing CCEs under the OEF setting. 

\subsection{Experimental Setting}\label{sec:experiments_setting}
OpenSpiel\footnote{\url{https://github.com/deepmind/open_spiel}} is an extensive collection of environments and algorithms for research in general reinforcement learning and search/planning in games \cite{lanctot2019openspiel}. It is widely accepted and implements many different games. We use it as our experimental platform and opt for Kuhn poker, Leduc poker, Liar's Dice, and Phantom Tic-Tac-Toe, which are all widely used in previous works~\cite{lisy2015online,brown2019solving}, as experimental domains. 
To get the OEF datasets, we generate three datasets for every game as introduced in Appendix~\ref{app:dateset} and mix the random and the expert datasets in different proportions to get hybrid datasets. Then we conduct our experiments on these offline datasets. NashConv (exploitability) is adopted for measuring the strategy of how close to NEs, and (C)CE Gap Sum is used as a measurement of the closeness to (C)CEs. 
All results are averaged over three seeds, and the error bars are also reported. Only selected results are shown here. The rest experiment results, ablation study, and parameter setting can be found in Appendix~\ref{app:results}.

\subsection{Comparison with Offline RL}
\begin{figure}[t]
\centering
\subfigure[Dataset Analysis]{
\includegraphics[width=0.21\textwidth]{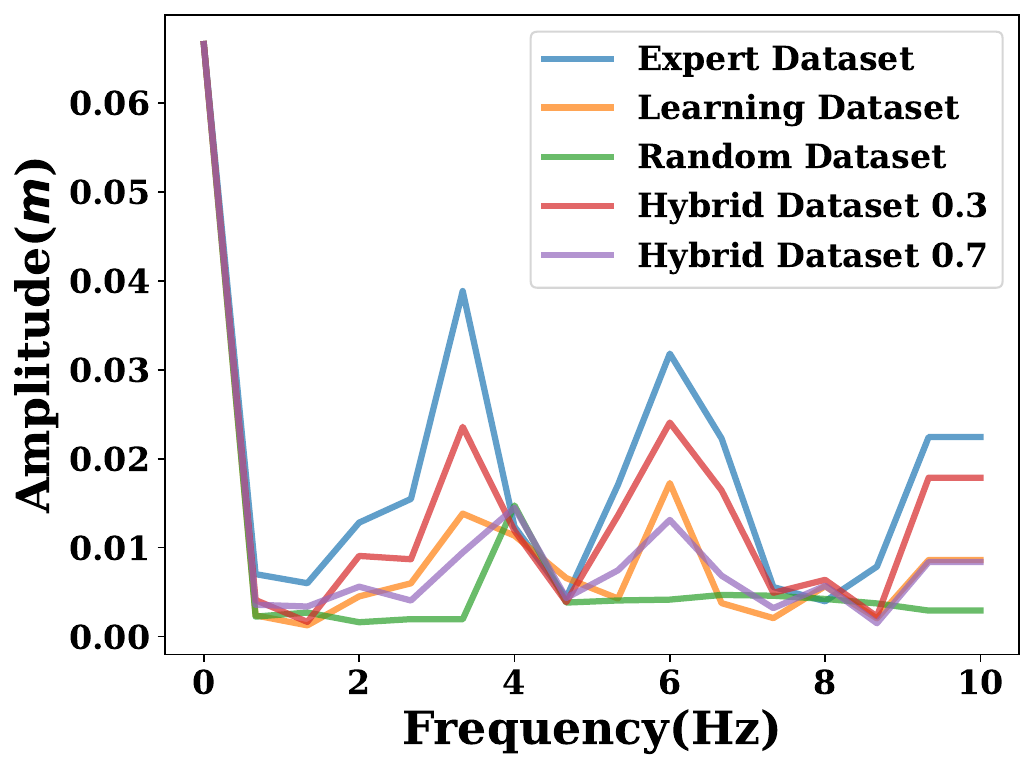}
\label{dataset_1}}
\subfigure[Offline RL]{
\includegraphics[width=0.21\textwidth]{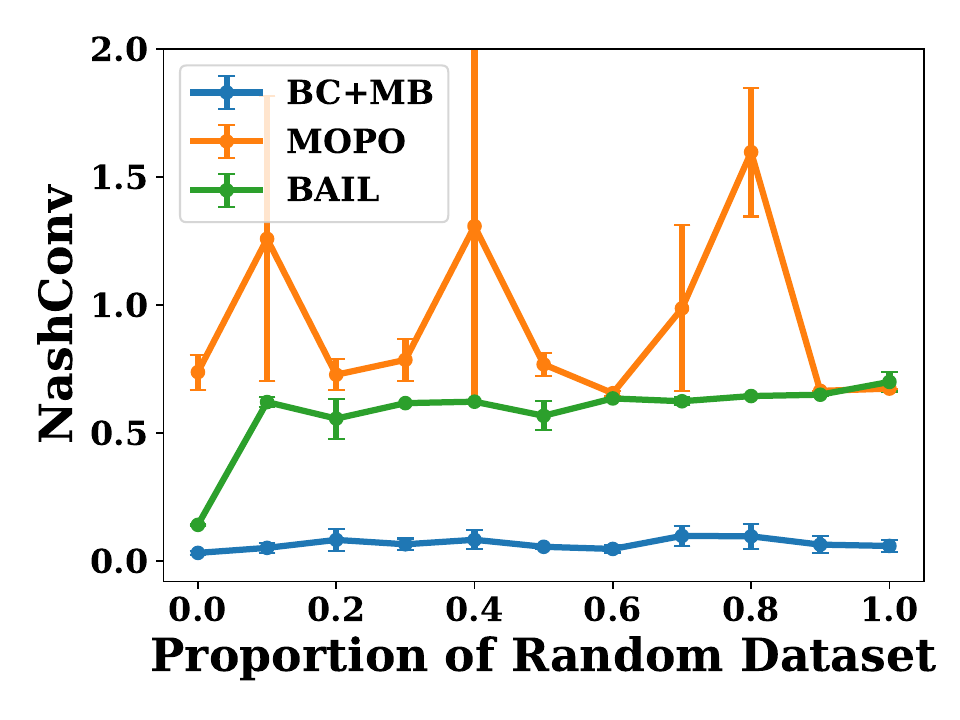}
\label{offline_1}}
\subfigure[Behavior Cloning]{
\includegraphics[width=0.21\textwidth]{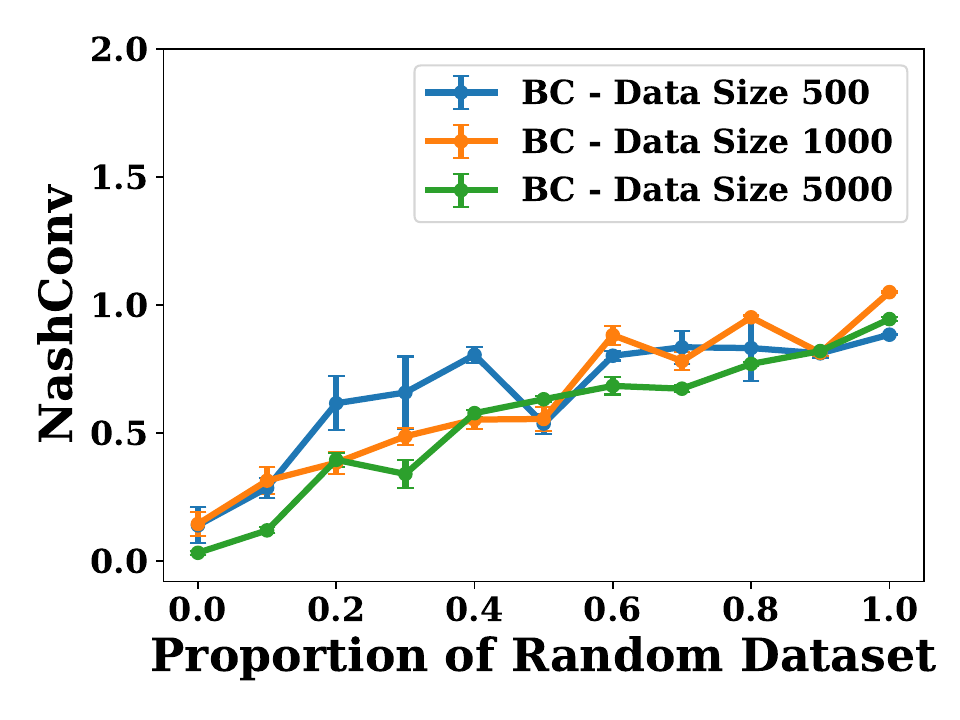} \label{bc_1}}
\subfigure[Model Based Method]{
\includegraphics[width=0.21\textwidth]{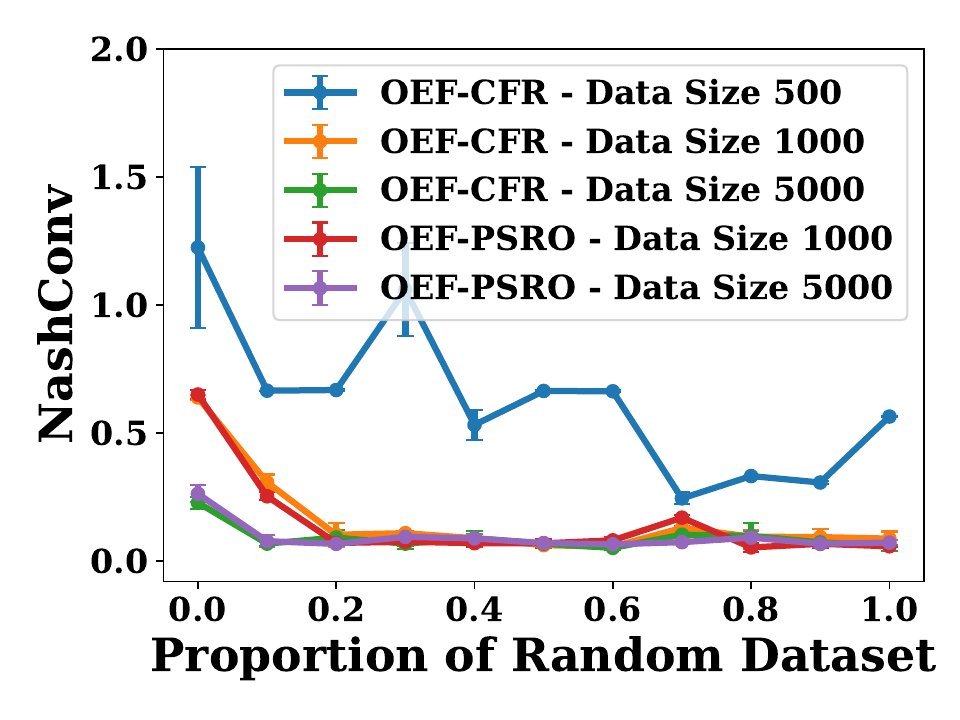}\label{mb_1}}
\subfigure[BC+MB]{
\includegraphics[width=0.21\textwidth]{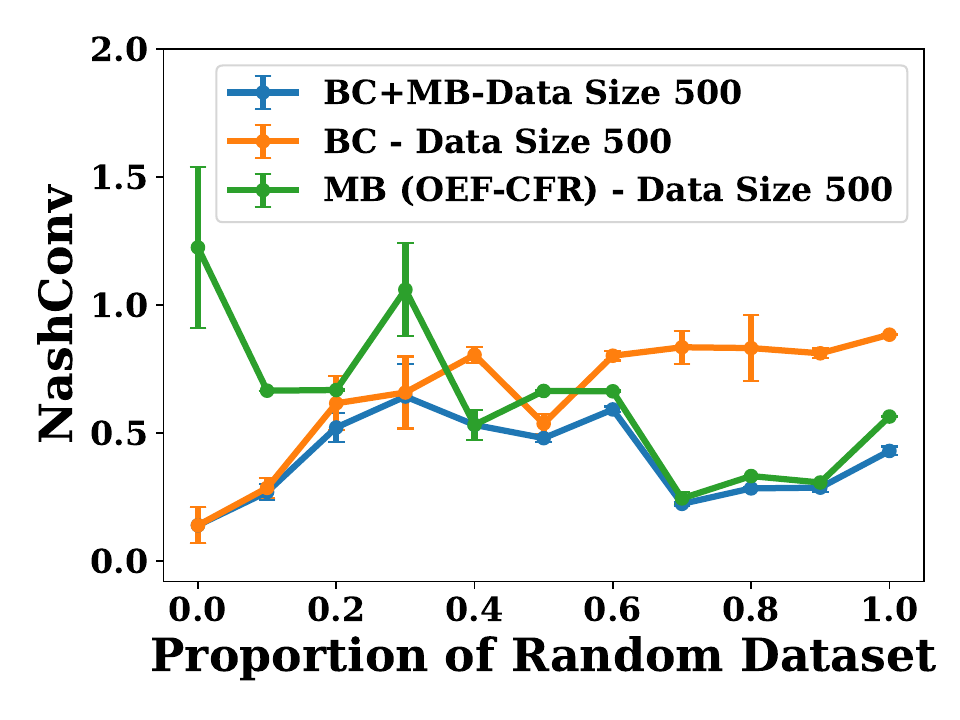}\label{bc_mb_1_1}}
\subfigure[BC+MB]{
\includegraphics[width=0.21\textwidth]{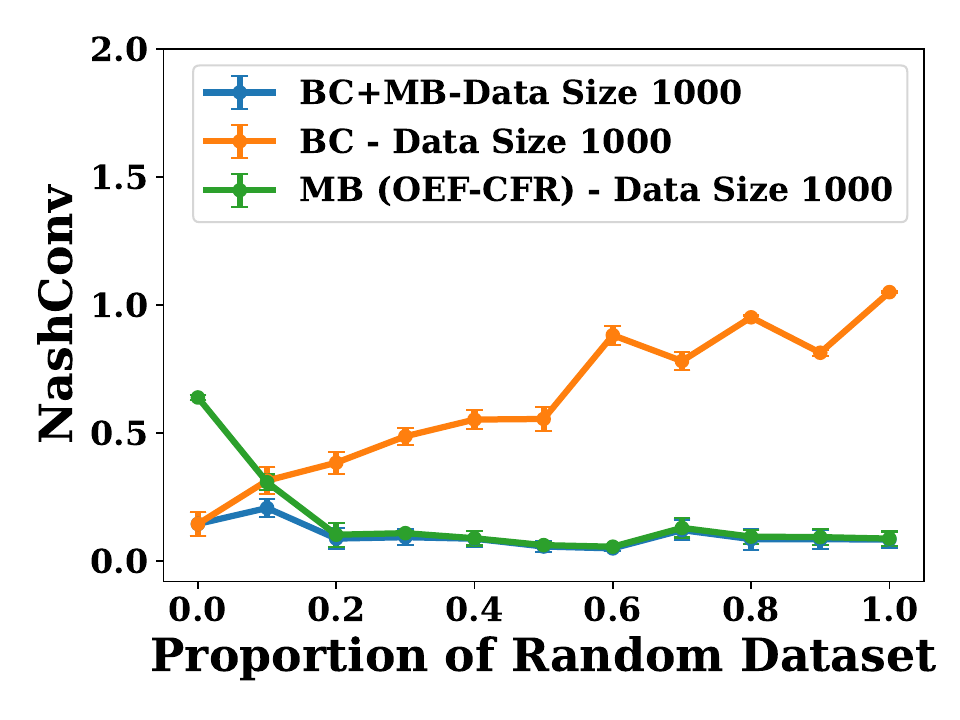}\label{bc_mb_1_2}}
\subfigure[Weight]{
\includegraphics[width=0.21\textwidth]{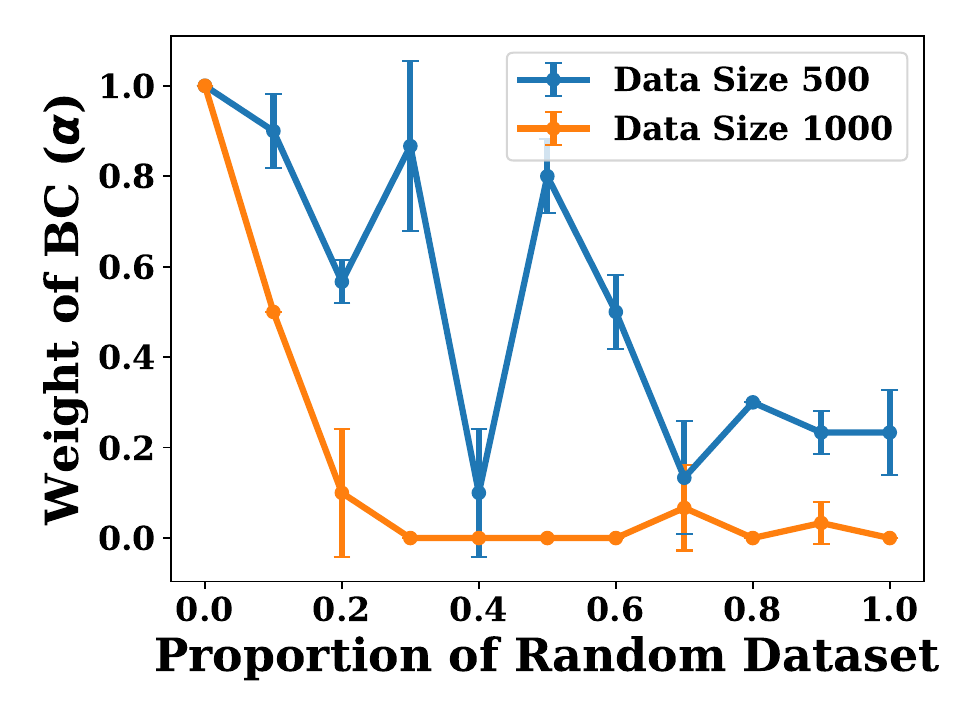}\label{weight_1}}
\subfigure[Learning Dataset]{
\includegraphics[width=0.21\textwidth]{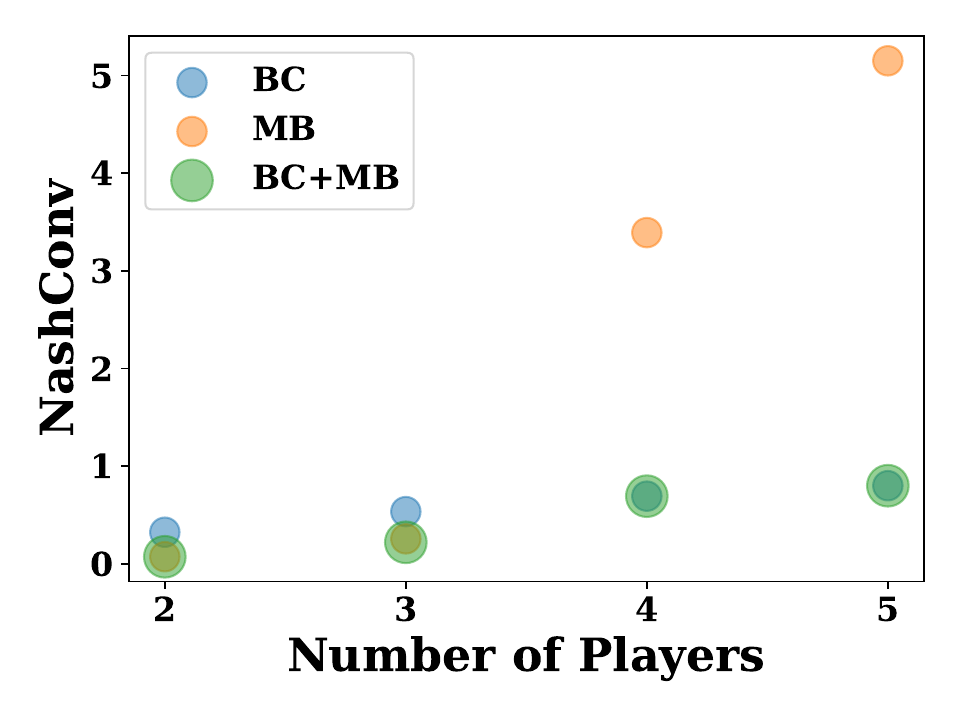}\label{learning_1}}
\caption{Experimental results on Kuhn poker.}
\vspace{-10pt}
\end{figure}

\begin{figure}[t]
\centering
\subfigure[Dataset Analysis]{
\includegraphics[width=0.21\textwidth]{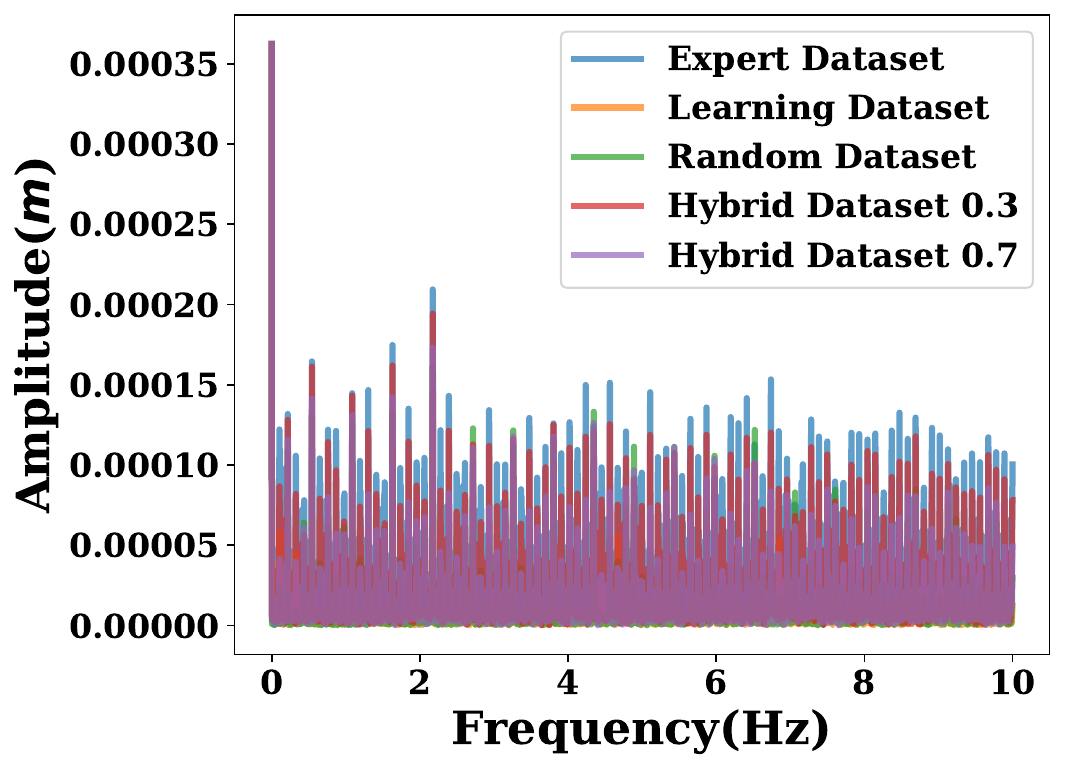}
\label{dataset_2}}
\subfigure[Offline RL]{
\includegraphics[width=0.21\textwidth]{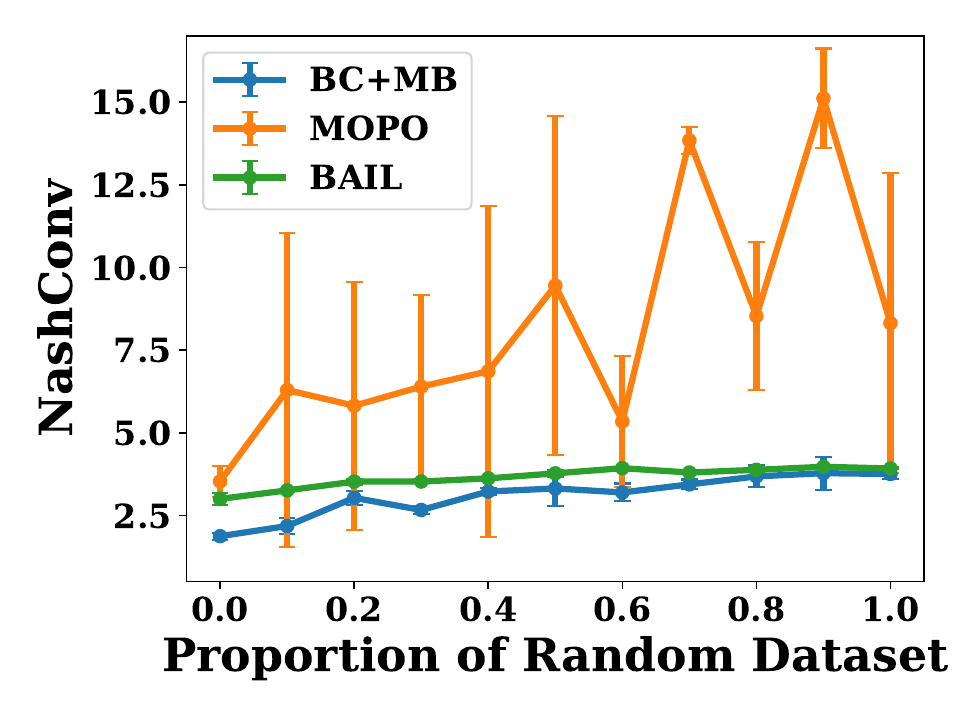}
\label{offline_2}}
\subfigure[Behavior Cloning]{
\includegraphics[width=0.21\textwidth]{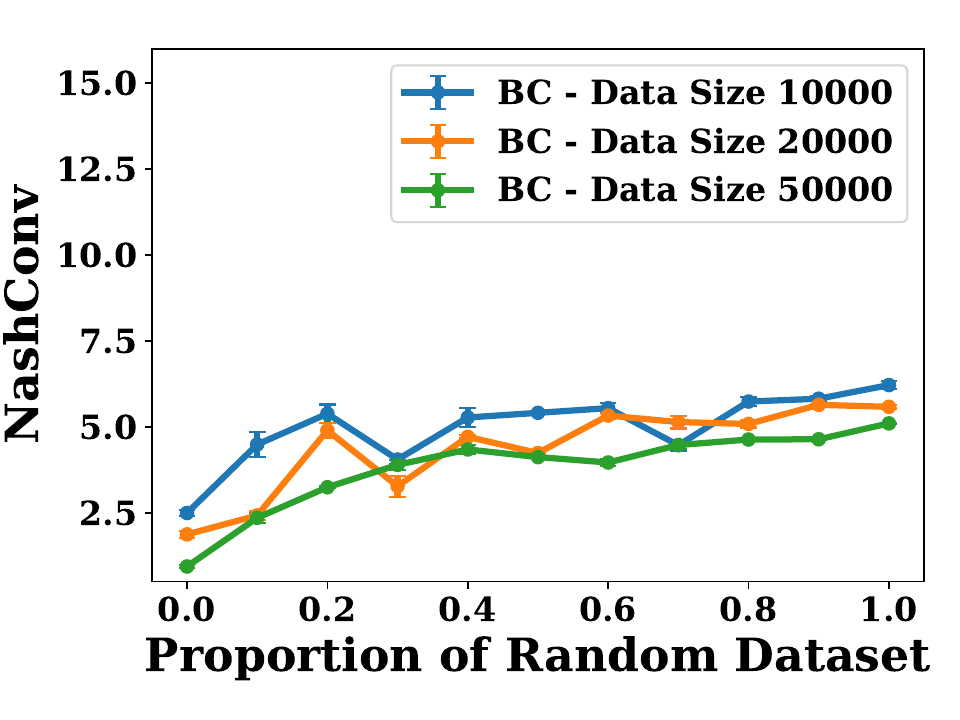}
\label{bc_2}}
\subfigure[Model Based Method]{
\includegraphics[width=0.21\textwidth]{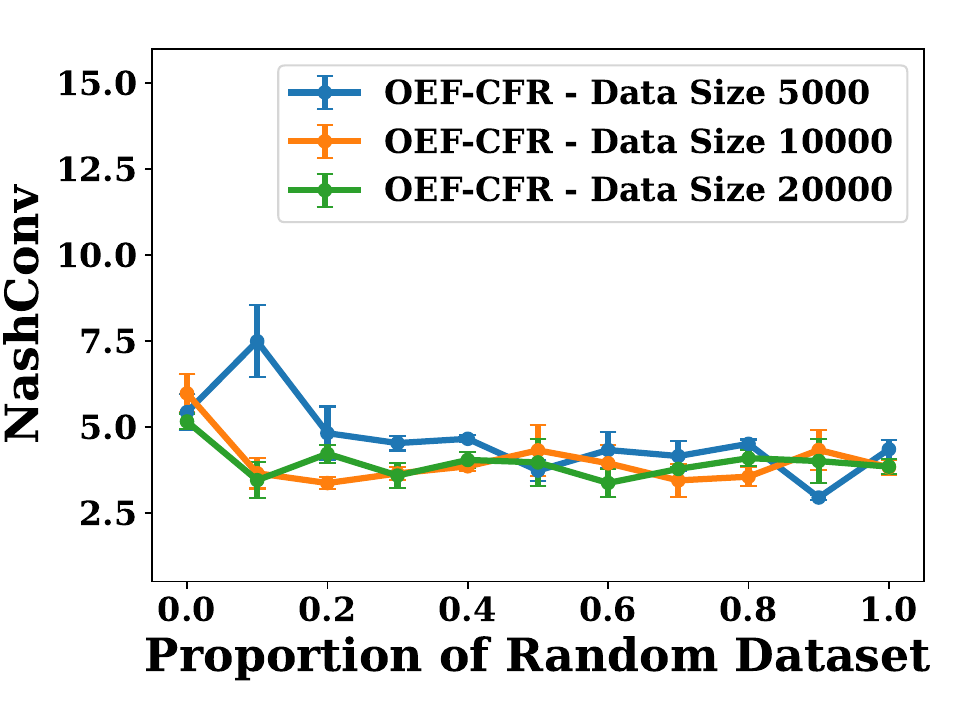}
\label{mb_2}}
\subfigure[BC+MB]{
\includegraphics[width=0.21\textwidth]{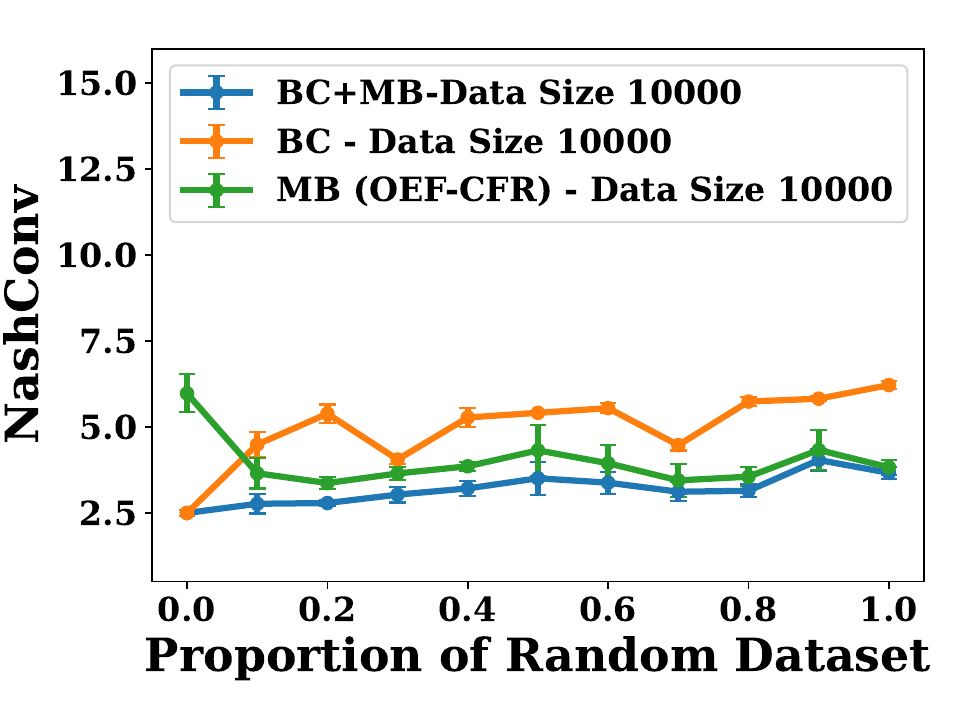}
\label{bc_mb_2_1}}
\subfigure[BC+MB]{
\includegraphics[width=0.21\textwidth]{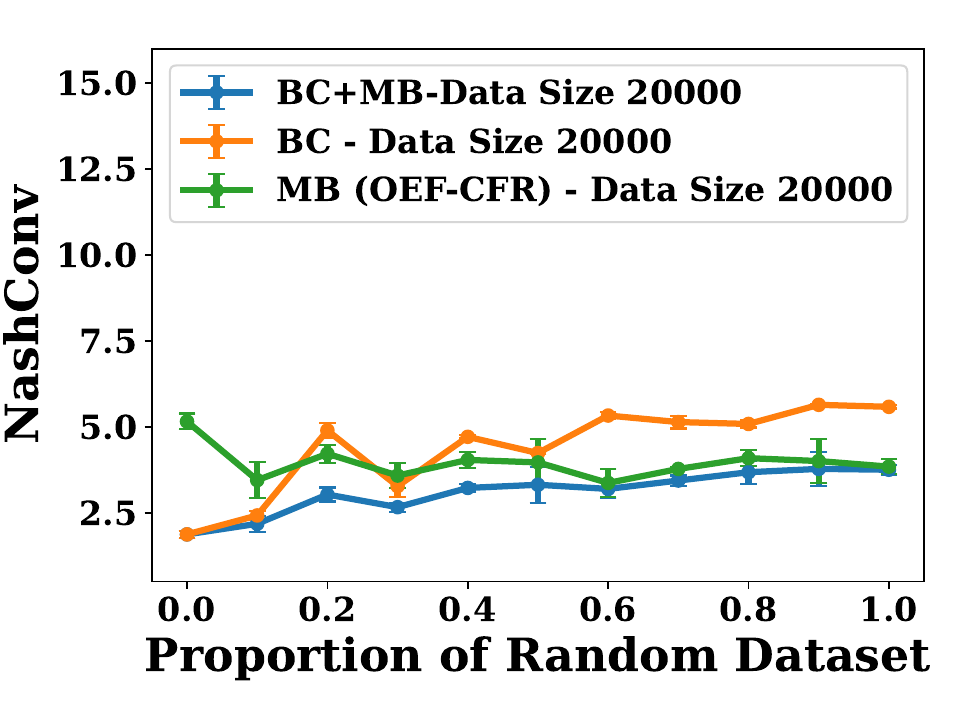}
\label{bc_mb_2_2}}
\subfigure[Weight]{
\includegraphics[width=0.21\textwidth]{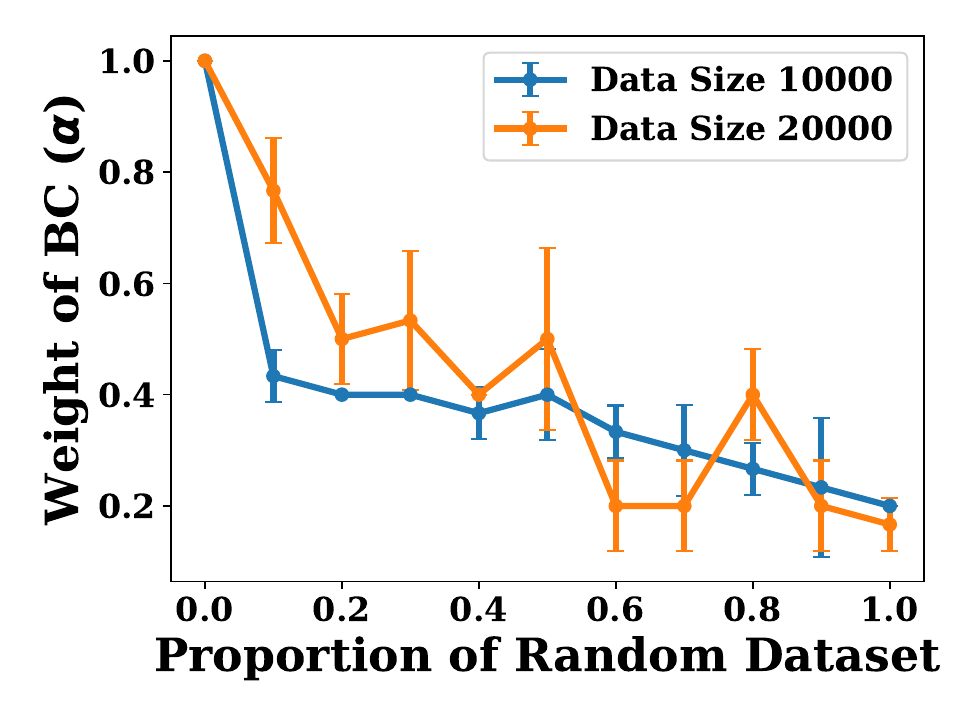}\label{weight_2}}
\subfigure[Learning Dataset]{
\includegraphics[width=0.21\textwidth]{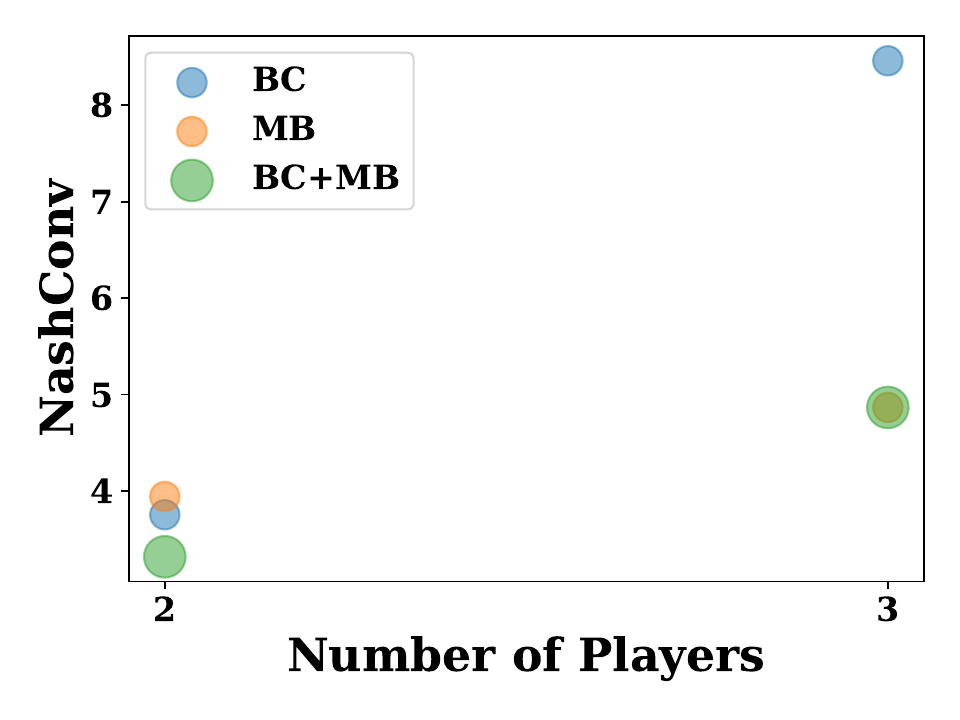}\label{learning_2}}
\caption{Experimental results on Leduc poker.}
\vspace{-10pt}
\end{figure}

In this section, we empirically show that naive offline RL algorithms are not enough for solving the OEF problem. To this end, we select one model-based offline RL algorithm -- Model-based Offline Policy Optimization (MOPO)~\cite{yu2020mopo} and one model-free offline RL algorithm -- Best-Action Imitation Learning (BAIL)~\cite{chen2020bail} as the representative of offline RL algorithms. 
Figures \ref{offline_1} and \ref{offline_2} show the comparison results of two-player Kuhn poker and two-player Leduc poker games. 
The x-axis represents the proportion of random data in the hybrid dataset. If the ratio is zero, the dataset equals the expert dataset, and if the percentage is one, it indicates that the hybrid dataset is the random dataset. 
We can find that in these hybrid datasets, our algorithm performs better than these two offline RL algorithms. It also shows that the performance of the MOPO algorithm varies widely regardless of the type of the dataset.
Compared to the MOPO algorithm, the performance of the BAIL algorithm is somewhat related to the quality of the dataset. 
However, neither of these offline RL algorithms can get a strategy profile close enough to the equilibrium strategy, which may be due to the players' policies being optimized independently. It verifies the claim that offline RL is not enough for the OEF setting.

\subsection{Computation of Nash Equilibrium}
Before running our OEF algorithms, we first analyze the OEF datasets. We count the frequency of leaf nodes and perform Fourier transform on them. Figures \ref{dataset_1} and \ref{dataset_2} show the results for different datasets. It can be found that the expert dataset has the highest amplitude which means that the expert dataset has more high-frequency data. In contrast, the random dataset has the lowest amplitude which means that the frequency of the data in the random dataset is similar. And the learning dataset and other hybrid datasets are intermediate between the expert dataset and the random dataset, which is consistent with our intuition. More analysis results of datasets can be found in Appendix~\ref{app:dataset visualization}.


We first run the behavior cloning technique and model-based algorithms solely on several games based on these hybrid datasets to assess their performance in computing the Nash equilibrium strategy. The NashConv is adopted to measure the distance to the NE strategy. Figures \ref{bc_1} and \ref{bc_2} show the results of BC on two-player Kuhn poker and Leduc poker games. We can see that as the proportion of the random dataset increases, the performance of BC policy decreases in these two games. The finding is consistent with our intuition that the BC technique can mimic the behavior strategy in the dataset and performs well only on datasets containing more expert data. It also shows that with the increase in the size of offline data, the performance evolves more stable while the improvement is not very significant. Therefore, the performance of BC policy depends on the quality of datasets, i.e., the quality of the behavior policy used to generate the dataset. Figures \ref{mb_1} and \ref{mb_2} show the results of the MB framework. From Figure \ref{mb_1}, we find that different model-based algorithms (OEF-CFR and OEF-PSRO) can get almost the same results. It indicates that the performance of the MB framework mainly depends on the quality of the trained environment model, and we can use either algorithm to calculate MB policy. Another finding is that with the increase in the size of offline data, the performance becomes better. It indicates that if the dataset includes enough data, the trained environment model is closer to the actual environment (test environment). 
From the above results, we can find that the BC policy performs poorly in the random dataset and performs well in the expert dataset. The MB framework performs slightly poorly in the expert dataset and performs well in the random dataset. More theoretical analysis can be found in Appendix \ref{app:theoretical}.

Then we move to perform our OEF algorithm -- BC+MB on these offline datasets to evaluate its performance. Figures \ref{bc_mb_1_1}-\ref{bc_mb_1_2} and \ref{bc_mb_2_1}-\ref{bc_mb_2_2} show the results of our OEF algorithm on two-player Kuhn and Leduc poker games. We also plot the results of BC and MB methods for comparison. We found that our OEF algorithm performs better than both BC and MB methods in all cases, which indicates that the combination is useful. The weights of BC policy ($\alpha$) which make these combined policies perform best on these datasets are shown in Figures \ref{weight_1} and \ref{weight_2}. As the proportion of the random dataset decreases, the weight of the BC policy in the final policy increases. It adheres to the intuition that BC policy performs well when the dataset includes many expert data. In that case, the weight of the BC policy in the final policy is high. We also test our OEF algorithm on poker games with different players under the learning datasets which can be viewed as datasets produced by unknown strategies. Figures \ref{learning_1} and \ref{learning_2} show that our OEF algorithm outperforms other methods in all games including multi-player Kuhn poker and Leduc poker. It indicates that given a dataset generated by an unknown strategy, our OEF algorithm can always get a better approximate equilibrium strategy. More experimental results on other games can be found in Appendix~\ref{app:results}.

\subsection{Computation of Coarse Correlated Equilibrium}
\begin{figure}[t]
\centering
\subfigure[Dataset Analysis]{
\includegraphics[width=0.21\textwidth]{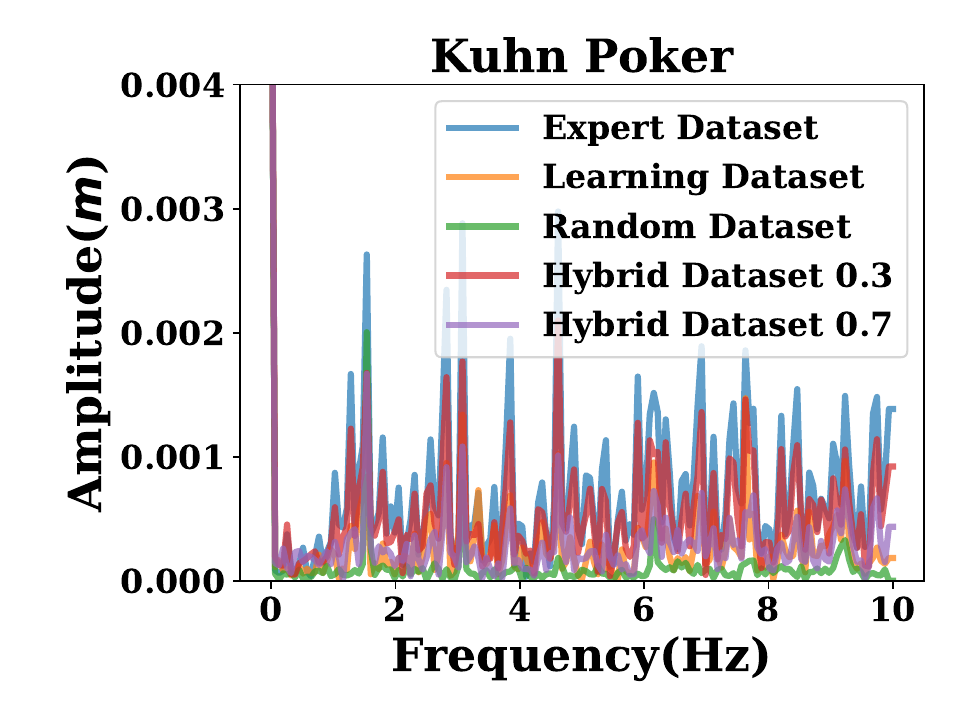}}
\subfigure[OEF-JPSRO]{
\includegraphics[width=0.21\textwidth]{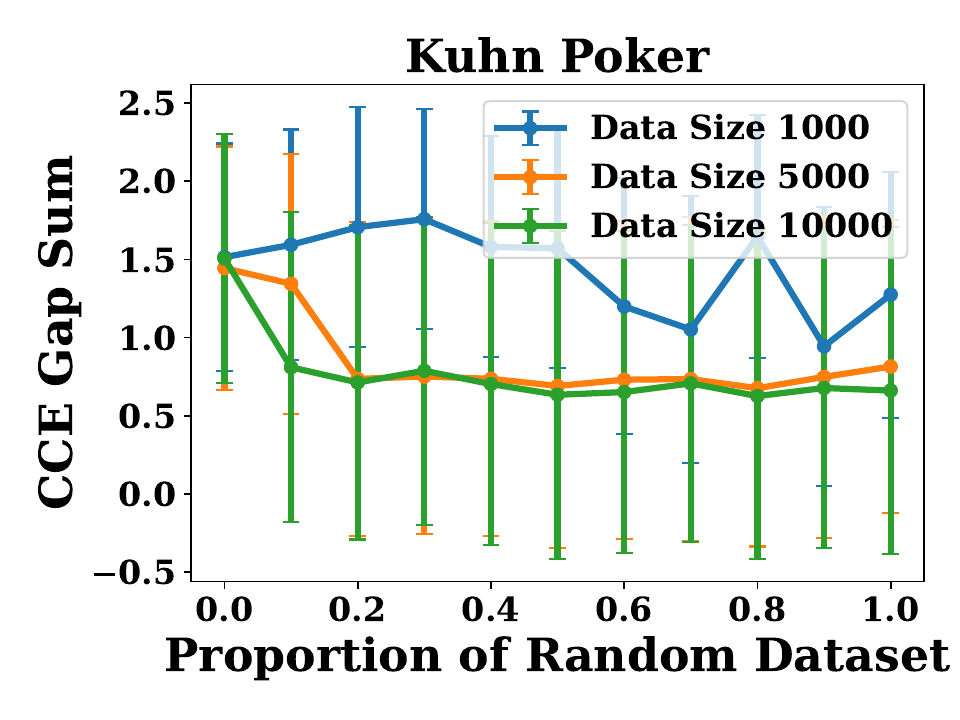}}
\subfigure[Dataset Analysis]{
\includegraphics[width=0.21\textwidth]{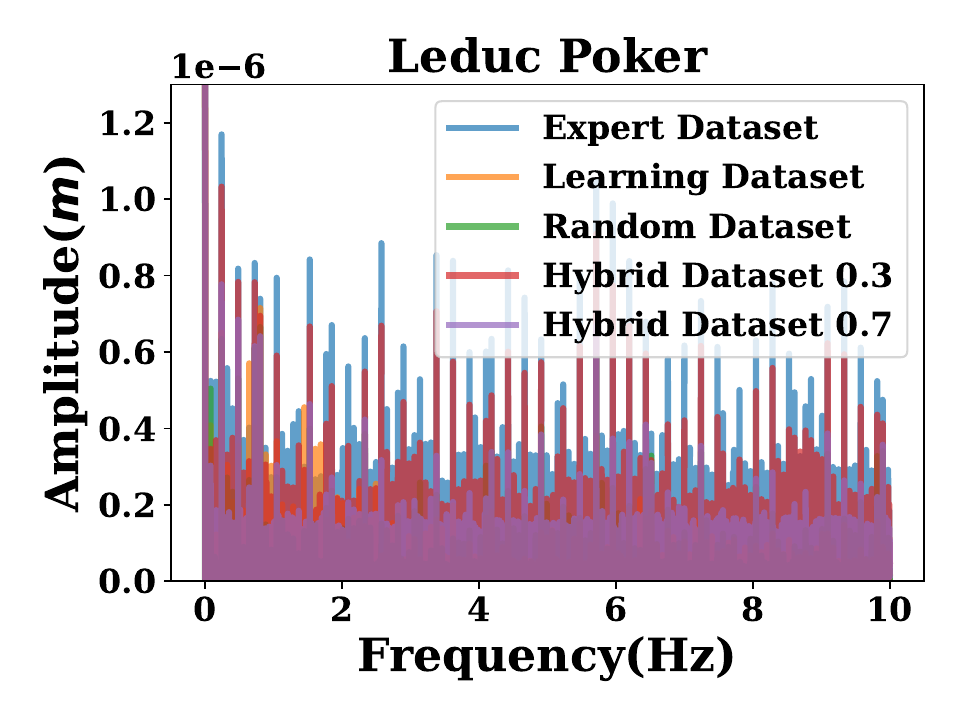}}
\subfigure[OEF-JPSRO]{
\includegraphics[width=0.21\textwidth]{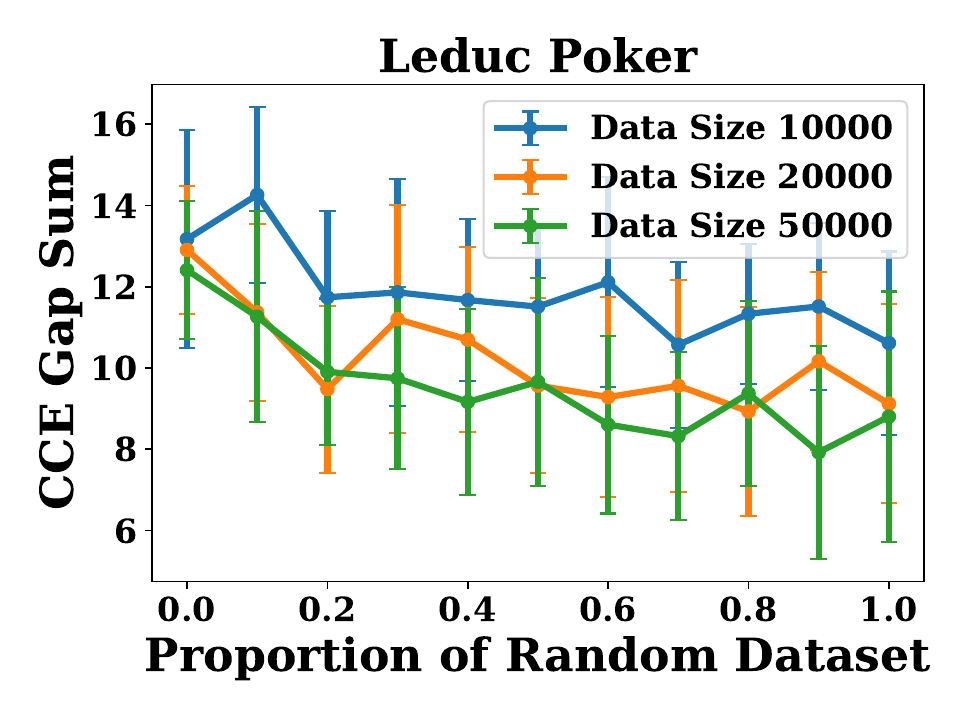}}
\caption{Experimental results on multi-player games.}
\label{jpsro}
\vspace{-10pt}
\end{figure}

To evaluate the performance in computing the CCE strategy, we perform the OEF-JPSRO algorithm on two three-player poker games under hybrid datasets and the CCE Gam Sum is adopted to measure the gap to the CCE. Here, we did not perform the behavior cloning technique since the offline dataset is collected using an independent strategy of every player instead of a joint strategy. In multiplayer games, although there is no guarantee to converge to NE, we can still use PSRO with $\alpha$-rank as the meta-solver to get a good strategy (low exploitability) to generate the expert dataset. Figure \ref{jpsro} shows the results of three-player Kuhn and Leduc poker games. The analysis of datasets shows the same result as before. As the size of the used offline data increases, we find that the performance of OEF-JPSRO improves. It further verifies that the performance of the model-based framework mainly depends on the trained environment model and its importance in solving the OEF problem. 



\section{Conclusion}
\label{sec:con}
We initiated an investigation of offline equilibrium finding (OEF), i.e., equilibrium finding on offline datasets, and constructed OEF datasets from widely-used games using three data-collecting strategies. To solve the OEF problem, we proposed a model-based framework that can generalize any online equilibrium finding algorithm with mere changes by introducing an environment model. Specifically, we adapted several existing online equilibrium finding algorithms to the OEF setting for computing different equilibrium solutions. To further improve the performance, we combined the behavior cloning technique with the model-based framework. Experimental results demonstrated that our algorithm performs better than existing offline RL algorithms and the model-based method is necessary for the OEF setting. We hope our efforts may open new directions in equilibrium findings and accelerate the research in game theory.

\textbf{Future works.} There are several limitations of this work that we intend to tackle in the future. First, the games we considered are rather smaller and large-scale games like Texas Hold'em poker~\cite{brown2018superhuman} were postponed till future work. Second, the types of generated offline datasets are limited. For future work, we plan to collect datasets using large-scale games and connect our library to StarCraft~II Unplugged~\cite{mathieu2021starcraft}. We will also include more data-collecting strategies (e.g., bounded rational agents) as well as additional human expert data\footnote{\url{http://poker.cs.ualberta.ca/irc_poker_database.html}} to diversify the provided datasets. 

\textbf{Negative Societal Impacts.} This work has no foreseeable negative societal impacts.




\bibliography{biblio}
\bibliographystyle{icml2023}

\newpage
\appendix
\onecolumn
\section{Frequently Asked Questions}
\label{app:FAQs}
\textbf{Q1: What is the impact of this work?}

Offline RL bridges the RL with real-world applications. We expect our offline equilibrium finding setting can open new directions in equilibrium finding and pave a path to solving real-world problems using game theory. More importantly, the offline RL algorithm can not directly apply to the OEF setting. Offline RL aims to compute the optimal strategy from the single agent perspective while this optimal strategy may be exploitable in the game setting. In this case, the Nash Equilibrium (NE) strategy may be a more suitable solution since NE consists of non-exploitable strategies. Therefore, OEF is important for getting a more robust strategy for some real-world problems.

\textbf{Q2: How to connect the example scenario with offline equilibrium finding?}

In the example scenario, to obtain a larger reward, Player A tends to apply the best strategy (i.e., the best response against the previous policy of Player B). However, this best strategy may be exploited by Player B if he changes his strategy accordingly. Therefore, Player A has to learn more game information by observing the replays (e.g., actions and preferences of Player B). To avoid being exploited as much as possible, the optimal solution for Player A is to choose the Nash equilibrium strategy of the underlying game. We have added more descriptions of this example scenario in the revision. 

\textbf{Q3: Why OEF is important and is more difficult than offline cooperative multi-agent RL?}

Using OEF algorithms, tailored to adversarial environments, is extremely important in strictly competitive games, such as security games. This setting is fundamentally different from offline multi-agent RL, which in general focuses on cooperation (instead of strict competition) between the agents. For example, consider the class of pursuit-evasion games, in which the pursuer (defender) chases the evader (attacker). Here, we cannot make any assumptions about the strategy of the attacker beforehand, since the attacker is strategic and learning. Using any vanilla offline RL algorithm to learn the optimal strategy for the defender based purely on historical data may result in a huge utility loss since the optimal strategy for the defender may be exploitable. In other words, the attacker may switch to the best response against the computed strategy of the defender instead of sticking to their past behavior estimated from the data. For this reason, attaining Nash Equilibrium (NE) may be a more suitable solution since NE consists of non-exploitable strategies.  

To be more specific, traditional offline RL focuses on learning the optimal strategy, i.e., obtaining the highest utility, for an agent acting in a dynamic environment modeled as a single MDP, which does not depend on the actions of other agents. In contrast, in two-player games, the dynamic for one player depends not only on the environment but also on the strategy of the opponent. In other words, the MDP a player acts in games is determined by both the game and the fixed strategy of the opponent, and hence a change in the opponent's strategy instigates a corresponding change of the MDP. This makes computing the best strategy for the defender against a strategic opponent using offline RL significantly more difficult. The framework of OEF we introduced provides methods for computing a player's NE strategy which is their optimal strategy against the strategic opponent (i.e., the worst case for the player).

\textbf{Q4: What are the differences between OEF and EGTA?}

1) As described in~\cite{wellman2006methods}, EGTA takes the game simulator as the fundamental input and performs strategic reasoning through interleaved simulation and game-theoretic analysis. Therefore, \textbf{the game simulator is required in EGTA.} In contrast, under the OEF setting, only the offline dataset is available and the game simulator is not required.

2) The estimated game model (empirical game) in EGTA is built based on the simulation's results, which are obtained by performing \textbf{known strategies} on the simulator. In contrast, in the OEF setting, the offline dataset is generated with an \textbf{unknown strategy}. In our work, although we use different behavior strategies to generate several offline datasets, we did not utilize these behavior strategies when performing our OEF algorithm.
Therefore, our proposed approach is different from EGTA. It is more challenging to find the equilibrium strategy in our OEF setting.

\textbf{Q5: What are the novelties of the proposed OEF algorithm -- BC+MB?}

We are the first ones to propose an empirical algorithm for solving the OEF problem. We introduce an environment model to propose a model-based framework that can generalize any existing online equilibrium finding algorithm to the context of the OEF setting. Due to the performance limitations of the model-based framework on some offline datasets, we combine a model-free algorithm -- behavior cloning technique with the model-based framework to improve the performance. Unlike these offline RL algorithms, it belongs to the model-based algorithms or model-free algorithms. Our algorithm combines the advantages of model-based and model-free to efficiently solve the OEF problem.

\clearpage
\section{Related Work Overview}
\label{app:related_works}
\textbf{Offline Reinforcement Learning (Offline RL).} Offline RL is a \textit{data-driven} paradigm that learns exclusively from static datasets of previously collected interactions, making it feasible to extract policies from large and diverse training datasets~\cite{levine2020offline}. This paradigm can be extremely valuable in settings where online interaction is impractical, either because data collection is expensive or dangerous (e.g., in robotics~\cite{singh2021reinforcement}, education~\cite{singla2021reinforcement}, healthcare~\cite{liu2020reinforcement}, and autonomous driving~\cite{kiran2021deep}). Therefore, efficient offline RL algorithms have a much broader range of applications than online RL and are particularly appealing for real-world applications~\cite{prudencio2022survey}. Due to its attractive characteristics, there have been a lot of recent studies. Here, we can divide the research of Offline RL into two categories: model-based and model-free algorithms. 

Model-free algorithms mainly use the offline dataset directly to learn a good policy. When learning the strategy from an offline dataset, we have two types of algorithms: actor-critic and imitation learning methods. Those actor-critic algorithms focus on implementing policy regularization and value regularization based on existing reinforcement learning algorithms. \citet{haarnoja2018soft} propose soft actor-critic (SAC) by adding an entropy regularization term to the policy gradient objective. This work mainly focuses on policy regularization. 
For the research of value regularization, an offline RL method named Constrained Q-Learning (CQL)~\cite{kumar2020conservative} learns a lower bound of the true Q-function by adding value regularization terms to its objective. 
Another line of research on learning a policy is imitation learning which mimics the behavior policy based on the offline dataset. \citet{chen2020bail} propose a method named Best-Action Imitation Learning (BAIL), which fits a value function, then uses it to select the best actions. Meanwhile, \citet{siegel2020keep} propose a method that learns an Advantage-weighted Behavior Model (ABM) and uses it as a prior in performing Maximum a-posteriori Policy Optimization (MPO)~\cite{abdolmaleki2018maximum}. It consists of multiple iterations of policy evaluation and prior learning until they finally perform a policy improvement step using their learned prior to extracting the best possible policy. 

Model-based algorithms rely on the offline dataset to learn a dynamics model or a trajectory distribution used for planning. The trajectory distribution induced by models is used to determine the best set of actions to take at each given time step. \citet{kidambi2020morel} propose a method named Model-based Offline Reinforcement Learning (MOReL), which measures their model’s epistemic uncertainty through an ensemble of dynamics models. Meanwhile, \citet{yu2020mopo} propose another method named Model-based Offline Policy Optimization (MOPO), which uses the maximum prediction uncertainty from an ensemble of models. Concurrently, \citet{matsushima2020deployment} propose the BehaviorREgularized Model-ENsemble (BREMEN) method, which learns an ensemble of models of the behavior MDP, as opposed to a pessimistic MDP. In addition, it implicitly constrains the policy to be close to the behavior policy through trust-region policy updates. More recently, \citet{yu2021combo} proposed a method named Conservative Offline Model-Based policy Optimization (COMBO), a model-based version of CQL. The main advantage of COMBO concerning MOReL and MOPO is that it removes the need for uncertainty quantification in model-based offline RL approaches, which is challenging and often unreliable. However, these above Offline RL algorithms can not directly apply to the OEF problem, which we have described in Section~\ref{rationale} and experimental results empirically verify this claim.   

\textbf{Empirical Game Theoretic Analysis (EGTA).} Empirical Game Theoretic Analysis is an empirical methodology that bridges the gap between game theory and simulation for practical strategic reasoning ~\cite{wellman2006methods}. In EGTA, game models are iteratively extended through a process of generating new strategies based on learning from experience with prior strategies. The strategy exploration problem~\cite{jordan2010strategy} that how to efficiently assemble an efficient portfolio of policies for EGTA is the most challenging problem in EGTA.

\citet{schvartzman2009stronger} deploy tabular RL as a best-response oracle in EGTA for strategy generation. They also build the general problem of strategy exploration in EGTA and investigate whether better options exist beyond best-responding to an equilibrium~\cite{schvartzman2009exploring}. Investigation of strategy exploration was advanced significantly by the introduction of the Policy Space Response Oracle (PSRO) framework~\cite{lanctot2017unified} which is a flexible framework for iterative EGTA, where at each iteration, new strategies are generated through reinforcement learning. Note that when employing NE as the meta-strategy solver, PSRO reduces to the double oracle (DO) algorithm~\cite{mcmahan2003planning}. In the OEF setting, only an offline dataset is provided, and there is no accurate simulator. In EGTA, a space of strategies is examined through simulation, which means that it needs a simulator, and the policies are known in advance. Therefore, techniques in EGTA cannot directly apply to OEF. 

\textbf{Opponent Modeling (OM) in Multi-Agent Learning.} Opponent modeling algorithm is necessary for multi-agent settings where secondary agents with competing goals also adapt their strategies, yet it remains challenging because policies interact with each other and change~\cite{he2016opponent}. One simple idea of opponent modeling is to build a model each time a new opponent or group of opponents is encountered~\cite{zheng2018deep}. However, it is infeasible to learn a model every time. A better approach is to represent an opponent’s policy with an embedding vector. \citet{grover2018learning} use a neural network as an encoder, taking the trajectory of one agent as input. Imitation learning and contrastive learning are also used to train the encoder. Then, the learned encoder can be combined with RL by feeding the generated representation into the policy or/and value network. DRON~\cite{he2016opponent} and DPIQN~\cite{hong2017deep} are two algorithms based on DQN, which use a secondary network that takes observations as input and predicts opponents’ actions. However, if the opponents can also learn, these methods become unstable. So it is necessary to take the learning process of opponents into account. 

\citet{foerster2017learning} propose a method named Learning with Opponent-Learning Awareness (LOLA), in which each agent shapes the anticipated learning of the other agents in the environment. Further, the opponents may still be learning continuously during execution. Therefore, \citet{al2017continuous} propose a method based on a meta-policy gradient named Mata-MPG. It uses trajectories from current opponents to perform multiple meta-gradient steps and constructs a policy that favors updating the opponents. Meta-MAPG~\cite{kim2021policy} extends this method by including an additional term that accounts for the impact of the agent’s current policy on the future policies of opponents, similar to LOLA. \citet{yu2021model} propose model-based opponent modeling (MBOM), which employs the environment model to adapt to various opponents. In the OEF setting, our goal is to compute the equilibrium strategy based on the offline dataset. Applying opponent modeling is not enough for calculating the equilibrium strategy in the OEF setting since the opponent will always best respond to the agent.

\textbf{Equilibrium Finding Algorithms.} The contemporary state-of-the-art algorithms for solving imperfect-information extensive-form games may be roughly divided into two groups: no-regret methods derived from CFR, and incremental strategy-space generation methods of the PSRO framework. 

For the first group, CFR is a family of iterative algorithms for approximately solving large imperfect-information games. Let $\sigma^t_i$ be the strategy used by player $i$ in round $t$. We define $u_i(\sigma, h)$ as the expected utility of player $i$ given that the history $h$ is reached, and then all players act according to strategy $\sigma$ from that point on. Let us define $u_i(\sigma, h\cdot a)$ as the expected utility of player $i$ given that the history $h$ is reached and then all players play according to strategy $\sigma$ except player $i$ who selects action $a$ in history $h$. Formally, $u_i(\sigma,h)=\sum_{z \in Z}\pi^\sigma(h,z)u_i(z)$ and $u_i(\sigma, h\cdot a)=\sum_{z \in Z}\pi^\sigma(h \cdot a,z)u_i(z)$.
The \textit{counterfactual value} $v_i^\sigma(I)$ is the expected value of information set $I$ given that player $i$ attempts to reach it. This value is the weighted average of the value of each history in an information set. The weight is proportional to the contribution of all players other than $i$ to reach each history. Thus, $v_i^\sigma(I) = \sum_{h\in I} \pi^\sigma_{-i}(h)\sum_{z \in Z}\pi^\sigma(h,z)u_i(z)$.
For any action $a \in A(I)$, the counterfactual value of action $a$ is $v_i^\sigma(I,a)= \sum_{h\in I} \pi^\sigma_{-i}(h)\sum_{z \in Z}\pi^\sigma(h \cdot a,z)u_i(z)$.
The \textit{instantaneous regret} for action $a$ in information set $I$ of iteration $t$ is $r^t(I,a)=v_{P(I)}^{\sigma^t}(I,a)-v_{P(I)}^{\sigma^t}(I)$. The \textit{counterfactual regret} for action $a$ in $I$ of iteration $T$ is $R^T(I,a) =\sum_{t=1}^{T}r^t(I,a)$. 
In vanilla CFR, players use \textit{Regret Matching} to pick a distribution over actions in an information set proportional to the positive cumulative regret of those actions. Formally, in iteration $T+1$, player $i$ selects action $a\in A(I)$ according to probabilities
\begin{equation}
\sigma^{T+1}(I,a)=
\begin{cases}
\frac{R^T_{+}(I,a)}{\sum_{b \in A(I)}R^T_{+}(I,b)} & \text{if $\sum\limits_{b \in A(I)}R^T_{+}(I,b) >0$,} \\
\frac{1}{|A(I)|}& \text{otherwise,} 
\end{cases}\nonumber
\end{equation}
where $R^T_{+}(I, a)=\max\{R^T(I, a),0\}$ because we are concerned about the cumulative regret when it is positive only.
If a player acts according to regret matching in $I$ on every iteration, then in iteration $T$, $R^T(I) \leq \Delta_i\sqrt{|A_i|}\sqrt{T}$ where $\Delta_i= \max_z u_i(z)-\min_zu_i(z)$ is the range of utilities of player $i$. Moreover, 
$R^T_i \leq \sum_{I\in \mathcal{I}_i}R^T(I) \leq |\mathcal{I}_i|\Delta_i\sqrt{|A_i|}\sqrt{T}$. 
Therefore, $\lim_{T \rightarrow \infty}\frac{R^T_i}{T} = 0$. In two-player zero-sum games, if both players' average regret  $\frac{R^T_i}{T} \leq \epsilon$, their average strategies $(\overline{\sigma}^T_1,\overline{\sigma}^T_2)$ form a $2\epsilon$-equilibrium~\cite{waugh2009abstraction}. 
Some variants are proposed to solve large-scale imperfect-information extensive-form games. Some sampling-based CFR variants~\cite{lanctot2009monte,gibson2012generalized,schmid2019variance} are proposed to effectively solve large-scale games by traversing a subset of the game tree instead of the whole game tree. With the development of deep learning techniques, neural network function approximation is also applied to the CFR algorithm. Deep CFR~\cite{brown2019deep}, Single Deep CFR~\cite{steinberger2019single}, and Double Neural CFR~\cite{li2019double} are algorithms using deep neural networks to replace the tabular representation in the CFR algorithm. 

For the second group, PSRO~\cite{lanctot2017unified} is a general framework that scales Double Oracle (DO)~\cite{mcmahan2003planning} to large extensive-form games via using reinforcement learning to compute the best response strategy approximately. To make PSRO more effective in solving large-scale games, Pipeline PSRO (P2SRO)~\cite{mcaleer2020pipeline} is proposed by parallelizing PSRO with convergence guarantees. Extensive-Form Double Oracle (XDO)~\cite{mcaleer2021xdo} is a version of PSRO where the restricted game allows mixing population strategies not only at the root of the game but every information set. It can guarantee to converge to an approximate NE in a number of iterations that are linear in the number of information sets, while PSRO may require a number of iterations exponential in the number of information sets. Neural XDO (NXDO) as a neural version of XDO learns approximate best response strategies through any deep reinforcement learning algorithm. Recently, Anytime Double Oracle (ADO)~\cite{mcaleer2022anytime}, a tabular double oracle algorithm for 2-player zero-sum games is proposed to converge to a Nash equilibrium while decreasing exploitability from one iteration to the next. Anytime PSRO (APSRO) as a version of ADO calculates best responses via reinforcement learning algorithms. Except for NEs, other equilibrium solution concepts, for example, (Coarse) Correlated equilibrium ((C)CE) is considered. Joint Policy Space Response Oracles (JPSRO)~\cite{marris2021multi} is proposed for training agents in n-player, general-sum extensive-form games, which provably converges to (C)CEs. 
The excellent performance of these equilibrium finding algorithms depends on the existence of efficient and accurate simulators. However, constructing a sufficiently accurate simulator may not be feasible or very expensive. In this case, we may resort to offline equilibrium finding (OEF) where the equilibrium strategy is computed based on the previous game data.

\clearpage
\section{Datasets}
\label{app:dateset}
In this section, we describe four types of data sets and how to collect them.

\textbf{Learning dataset.} This type of dataset is collected when learning the Nash equilibrium strategy. When solving a game using some existing equilibrium finding algorithm, the players have to interact with the game environment. During the course of learning, we may gather these intermediate interaction game data and store them as a learning dataset. In contrast to the random dataset, the players' strategies gradually improve as we get closer to Nash equilibria.   

\textbf{Expert dataset.} This type of dataset is collected using an NE strategy. The motivation behind the dataset is that when learning a game, we often prefer to observe more experienced players at play. We simulate the expert players using the NE strategy and collect the interaction data. We follow a similar methodology as with the random dataset. First, we compute the NE strategies using any existing equilibrium finding algorithm. As a second step, the Nashian players repeatedly interact in the game. Finally, we gather the generated data and store them as the expert dataset. 
In multi-player or general-sum games, although CFR or PSRO cannot converge to NE, we also apply these algorithm for collecting the expert dataset. Although there is no guarantee, we can still get a good strategy using these algorithms, for example, PSRO with $\alpha$-rank as the meta-solver \cite{muller2019generalized} can get pretty good strategy (low exploitability) under general-sum many-player games.

\textbf{Hybrid dataset.} In addition to the three types of datasets mentioned above, we also consider hybrid datasets consisting of random and expert interactions mixed in different ratios to simulate offline datasets generated by unknown strategies. 

We collect the data from player engagements in the most frequently used benchmarking imperfect-information extensive-form games in contemporary research on the equilibrium finding. These games include poker games (two-player and multi-player Kuhn poker, two-player and multi-player Leduc poker), Phantom Tic-Tac-Toe, and Liar's Dice. 

\begin{figure}[ht]
\centering
\includegraphics[width=0.9\textwidth]{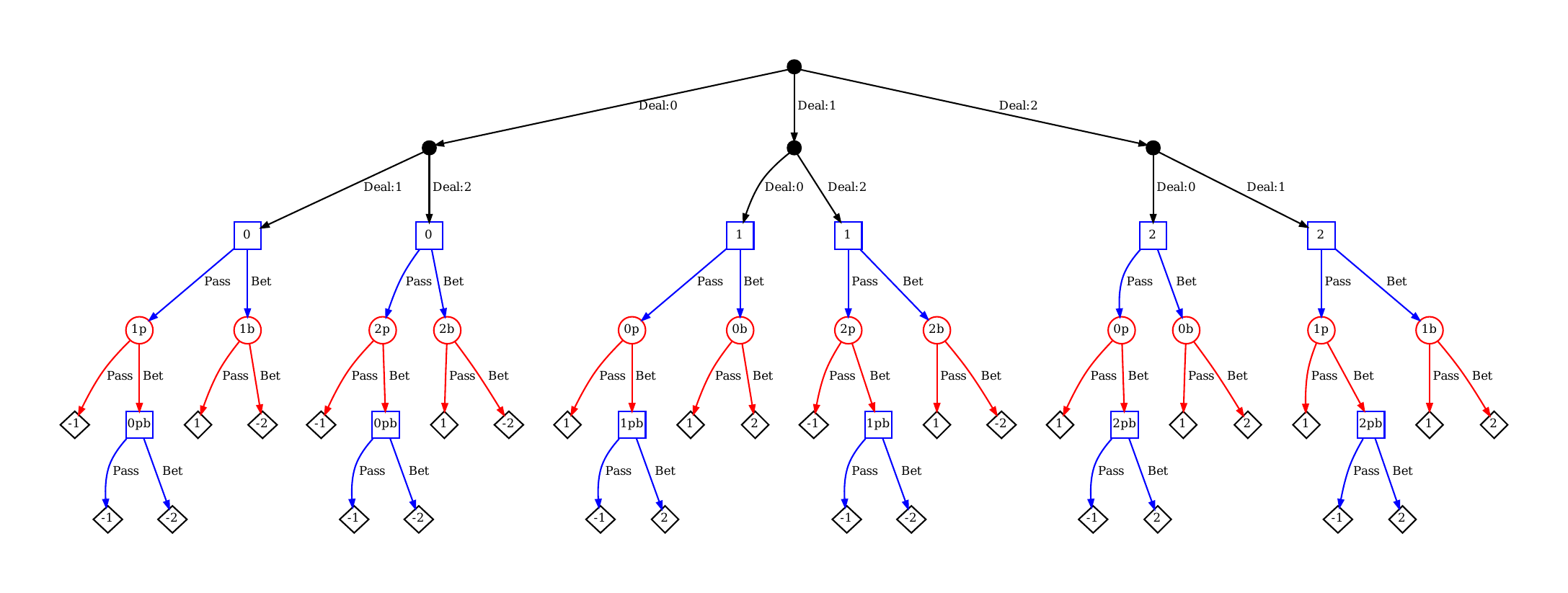}
\caption{2-player Kuhn poker}
\label{fig:2_kuhn}
\end{figure}
\section{Visualization of Datasets}
\label{app:dataset visualization}
In this section, we describe the visualization methods of the datasets. First, we plot the game tree. Figure \ref{fig:2_kuhn} shows an example of the game tree. Then, we traverse the game tree using depth-first search (DFS) and index each leaf node according to the DFS results. Finally, we count the frequency of the leaf node in each dataset. The reason why we only count the frequency of the leaf node is that every leaf node determines one sampled trajectory from the root node. And the dataset is sampled using a specified strategy. Therefore the frequency of the leaf node can reflect the distribution of the dataset. In the main paper, we use the Fourier transform to analyze the dataset since the Fourier transforms on the frequency can be used to analyze the high-frequency data in the dataset. In other words, the higher amplitude value means that there are more high-frequency data in the dataset. Here, we provide more methods to visualize these datasets.

\begin{figure}[ht]
\centering
\subfigure[Kuhn Poker-2 Player]{
\includegraphics[width=0.23\textwidth]{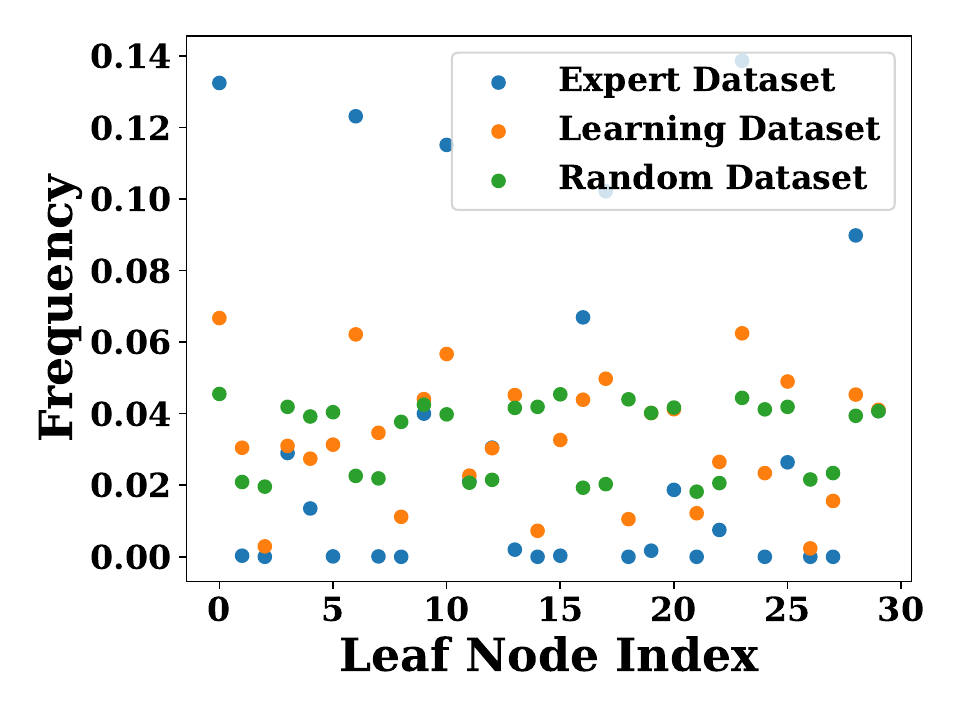}}
\subfigure[Kuhn Poker-3 Player]{
\includegraphics[width=0.23\textwidth]{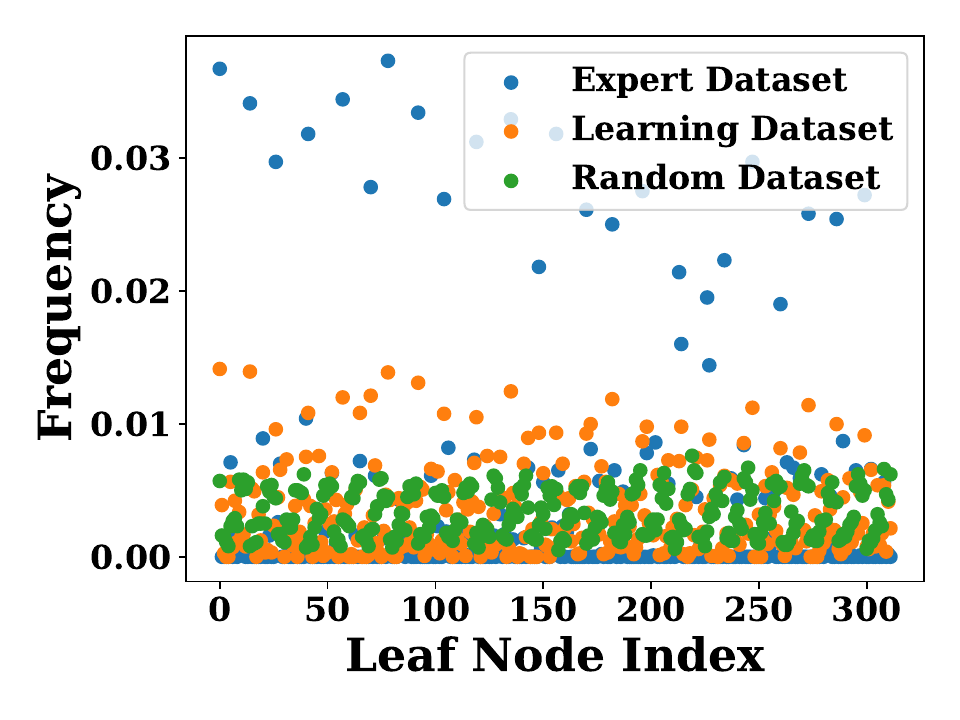}}
\subfigure[Kuhn Poker-4 Player]{
\includegraphics[width=0.23\textwidth]{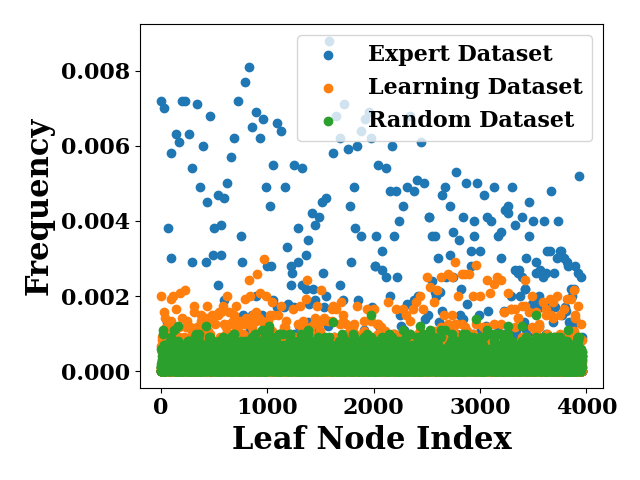}}
\subfigure[Kuhn Poker-5 Player]{
\includegraphics[width=0.23\textwidth]{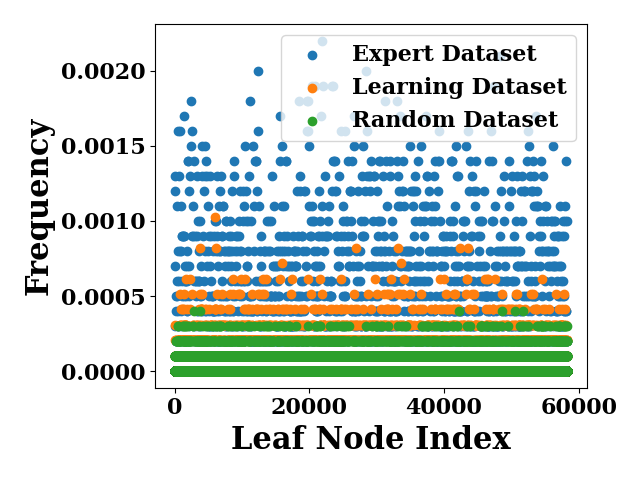}}
\subfigure[Leduc Poker-2 Player]{
\includegraphics[width=0.25\textwidth]{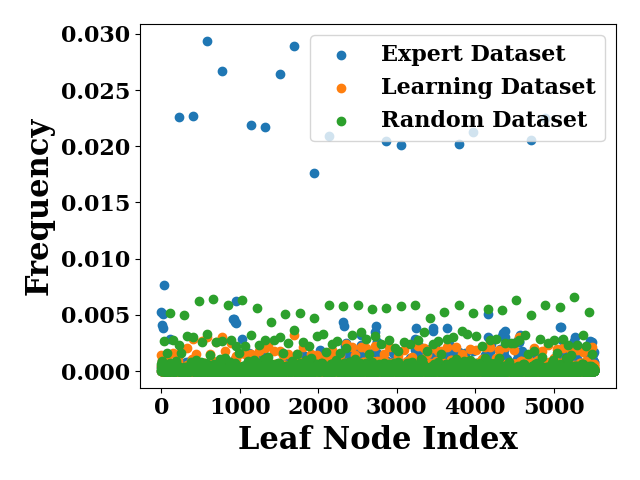}}
\subfigure[Leduc Poker-3 Player]{
\includegraphics[width=0.25\textwidth]{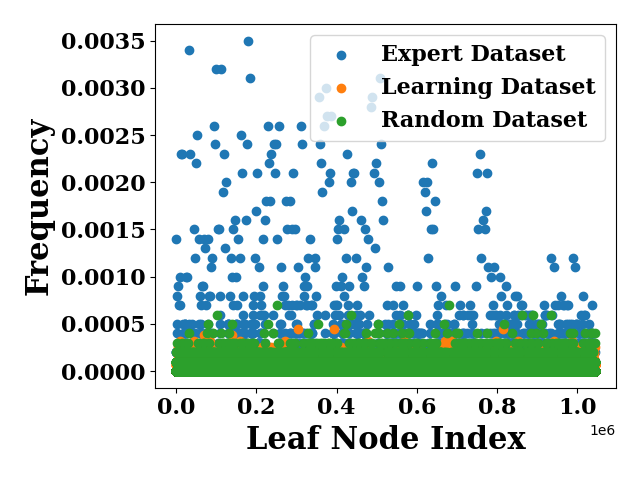}}
\subfigure[Liars Dice-2 Player]{
\includegraphics[width=0.25\textwidth]{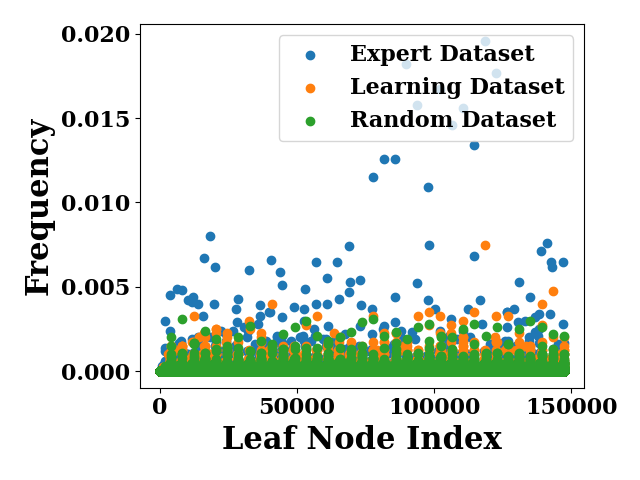}}
\caption{DataSet Visualization}
\label{visual_1}
\centering
\subfigure[Kuhn Poker-2 Player]{
\includegraphics[width=0.23\textwidth]{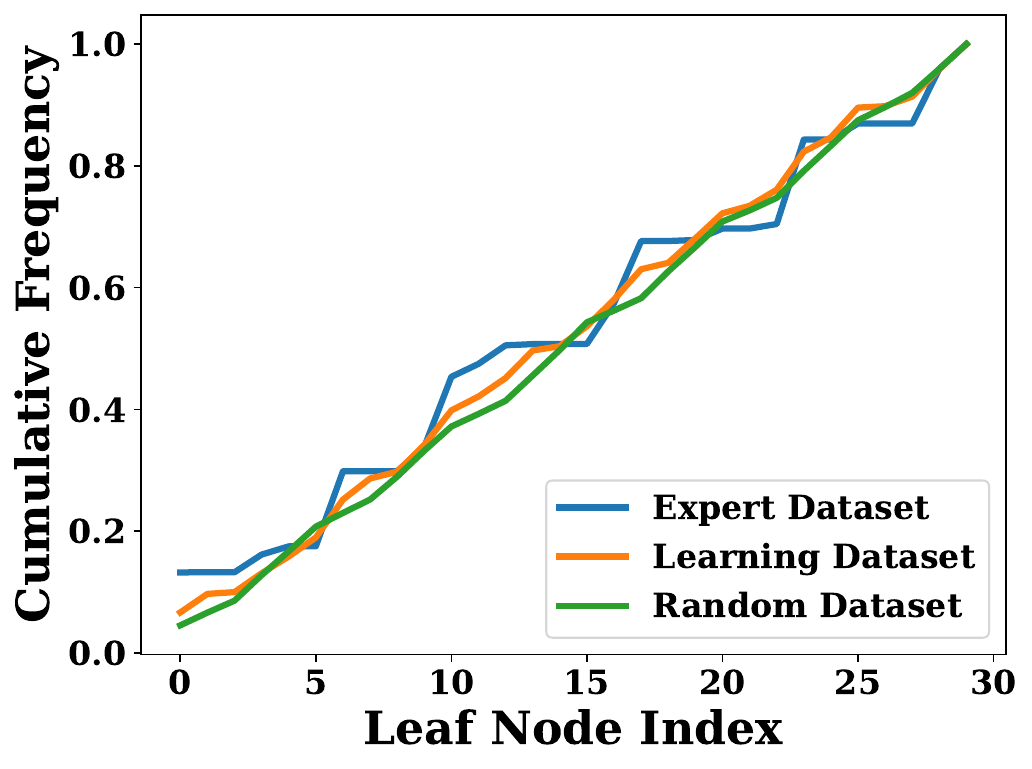}}
\subfigure[Kuhn Poker-3 Player]{
\includegraphics[width=0.23\textwidth]{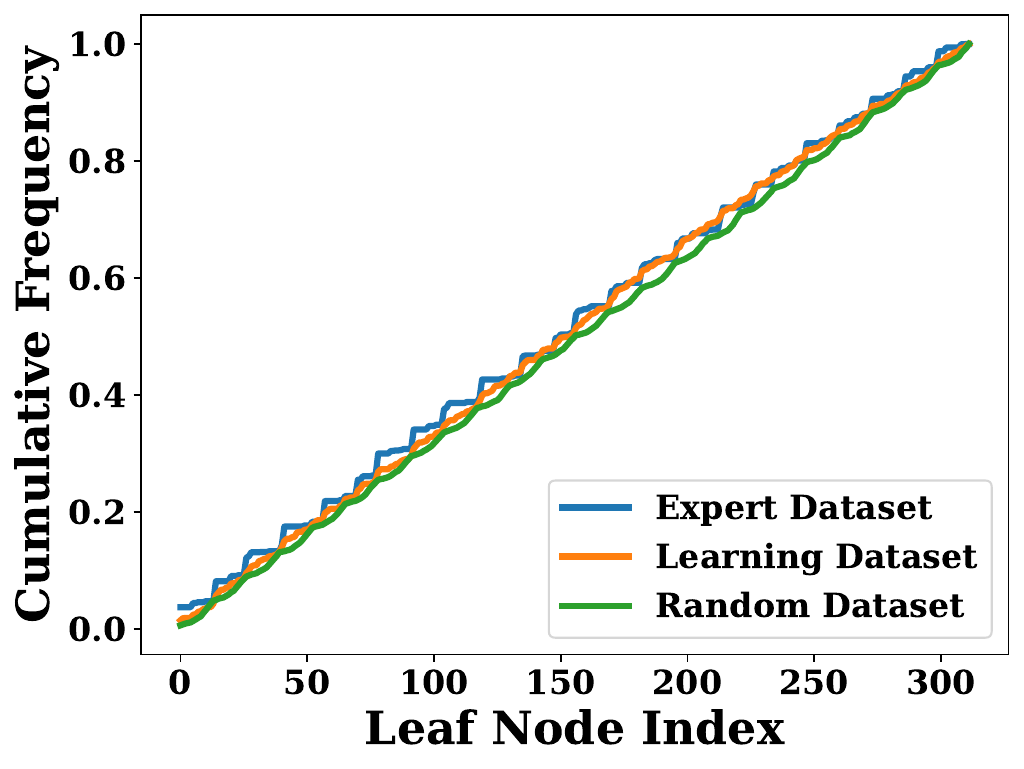}}
\subfigure[Kuhn Poker-4 Player]{
\includegraphics[width=0.23\textwidth]{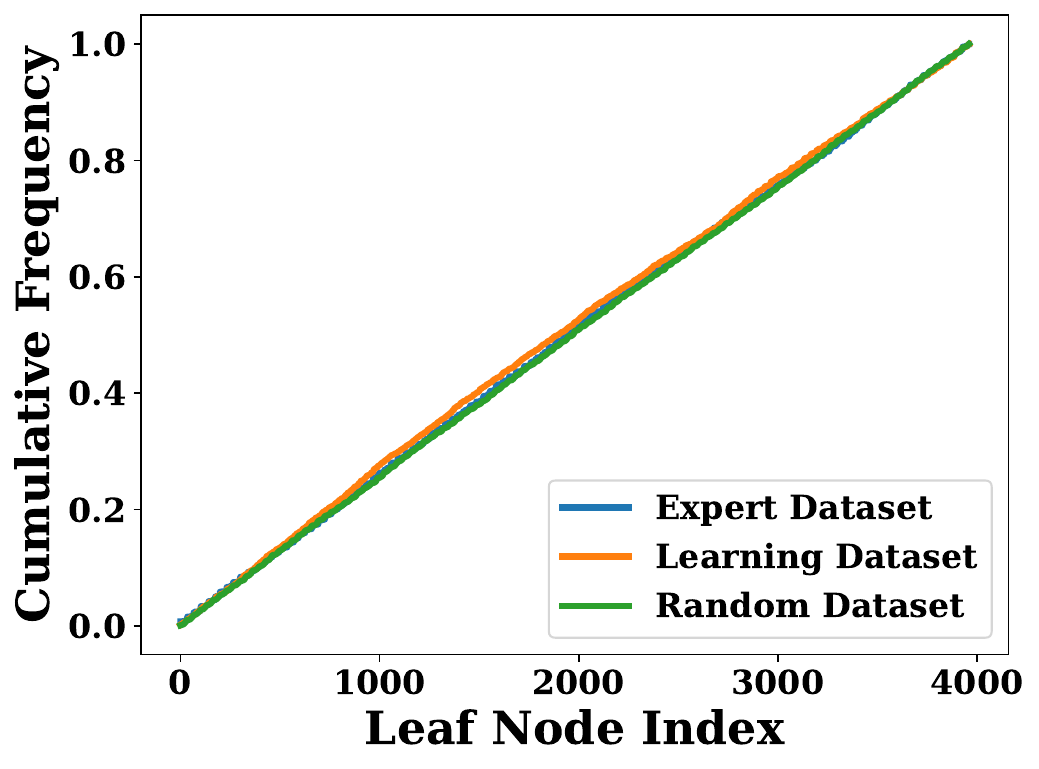}}
\subfigure[Kuhn Poker-5 Player]{
\includegraphics[width=0.23\textwidth]{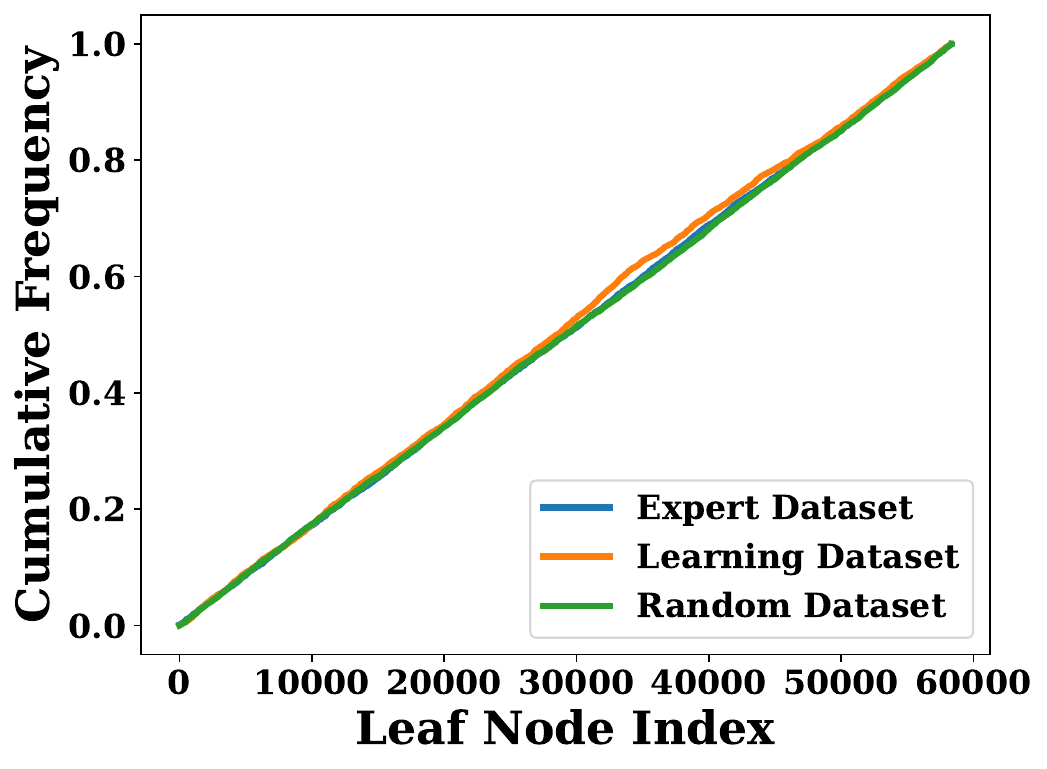}}
\subfigure[Leduc Poker-2 Player]{
\includegraphics[width=0.25\textwidth]{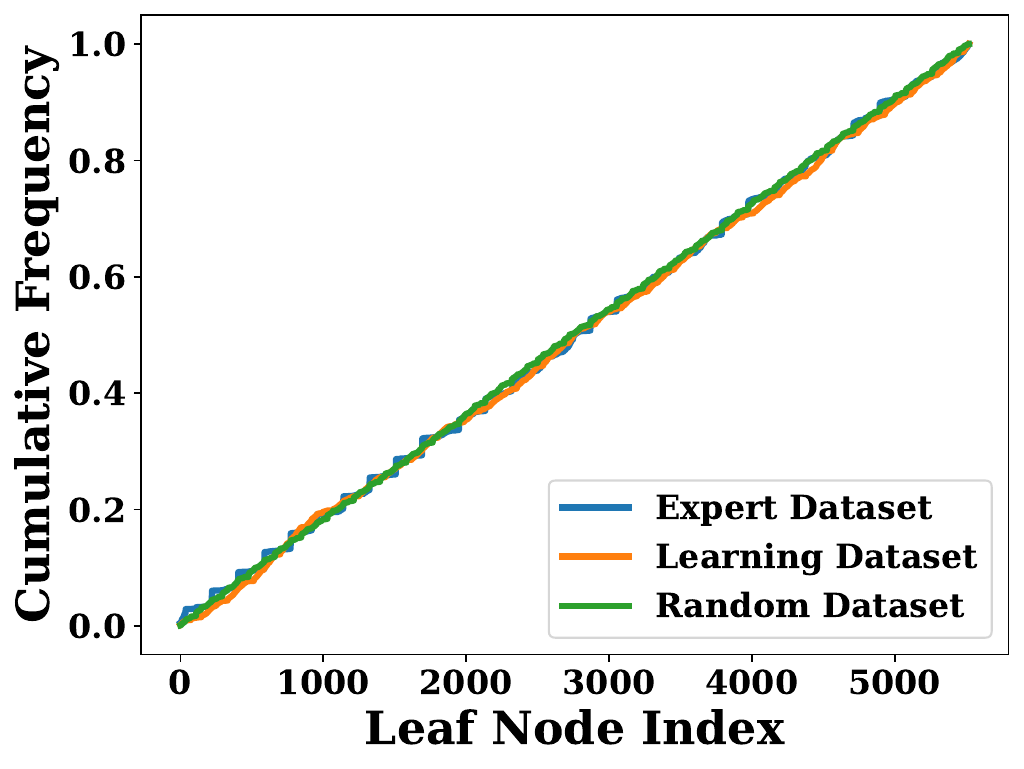}}
\subfigure[Leduc Poker-3 Player]{
\includegraphics[width=0.25\textwidth]{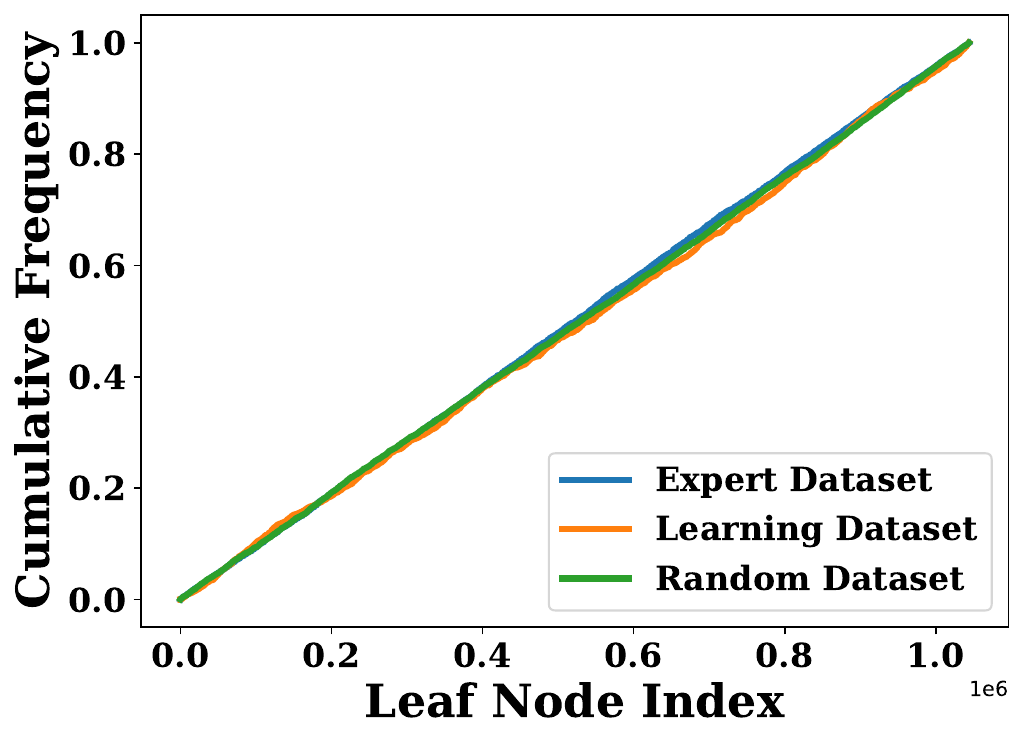}}
\subfigure[Liars Dice-2 Player]{
\includegraphics[width=0.25\textwidth]{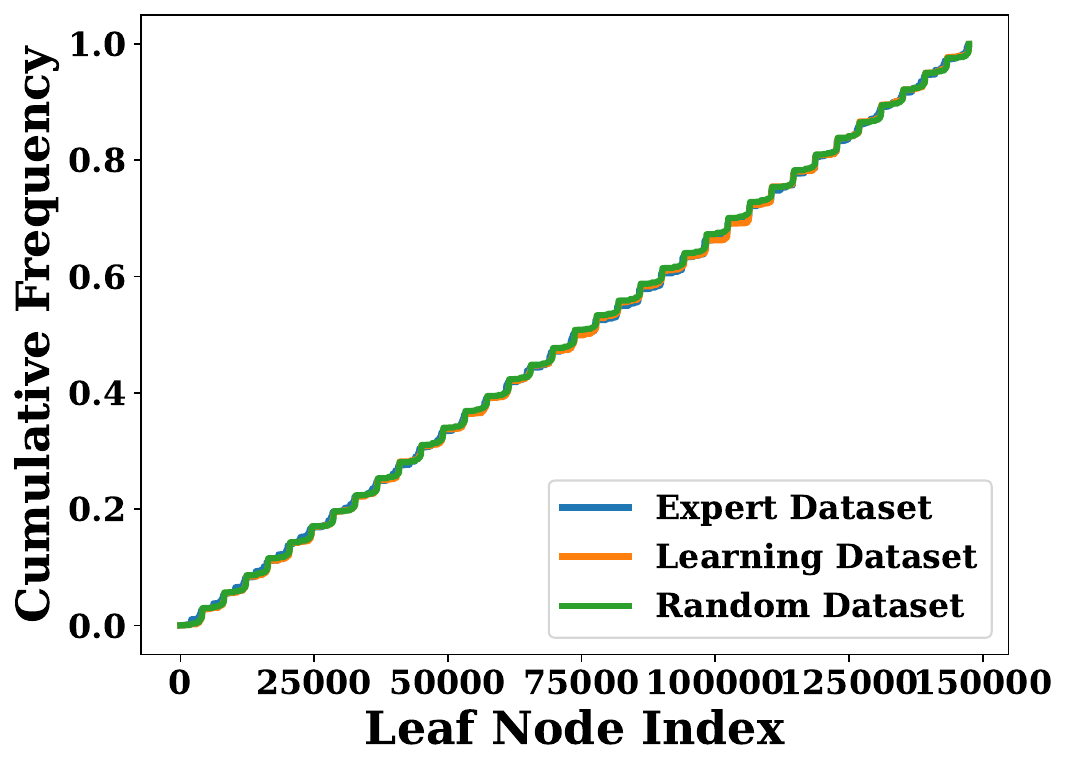}}
\caption{DataSet Visualization}
\label{visual_2}
\end{figure}

Figure \ref{visual_1} show the frequency of leaf node in datasets. We can find that in the random dataset, the frequency of leaf nodes is almost uniform, while in the expert dataset, the frequency distribution of leaf nodes is uneven. The distribution of the learning dataset is between the expert dataset and the random dataset. Figure \ref{visual_2} shows the cumulative frequency of leaf nodes. These figures also exhibit the same result. 

\clearpage
\section{Theoretical Analysis}
\label{app:theoretical}
The concurrent works \cite{cui2022offline, zhong2022pessimistic} investigate the necessary properties of offline datasets of two-player zero-sum Markov games to successfully infer their NEs. To do this, they proposed several dataset coverage assumptions. Following their assumptions \cite{cui2022offline}, we also define some hypotheses on the dataset coverage under our OEF setting and provide extensive analysis about how the dataset coverage influences computing the equilibrium under the OEF setting. Our results are mainly for computing Nash equilibrium in extensive-form games. 

\begin{figure}[h]
        \centering
\includegraphics[width=0.5\textwidth]{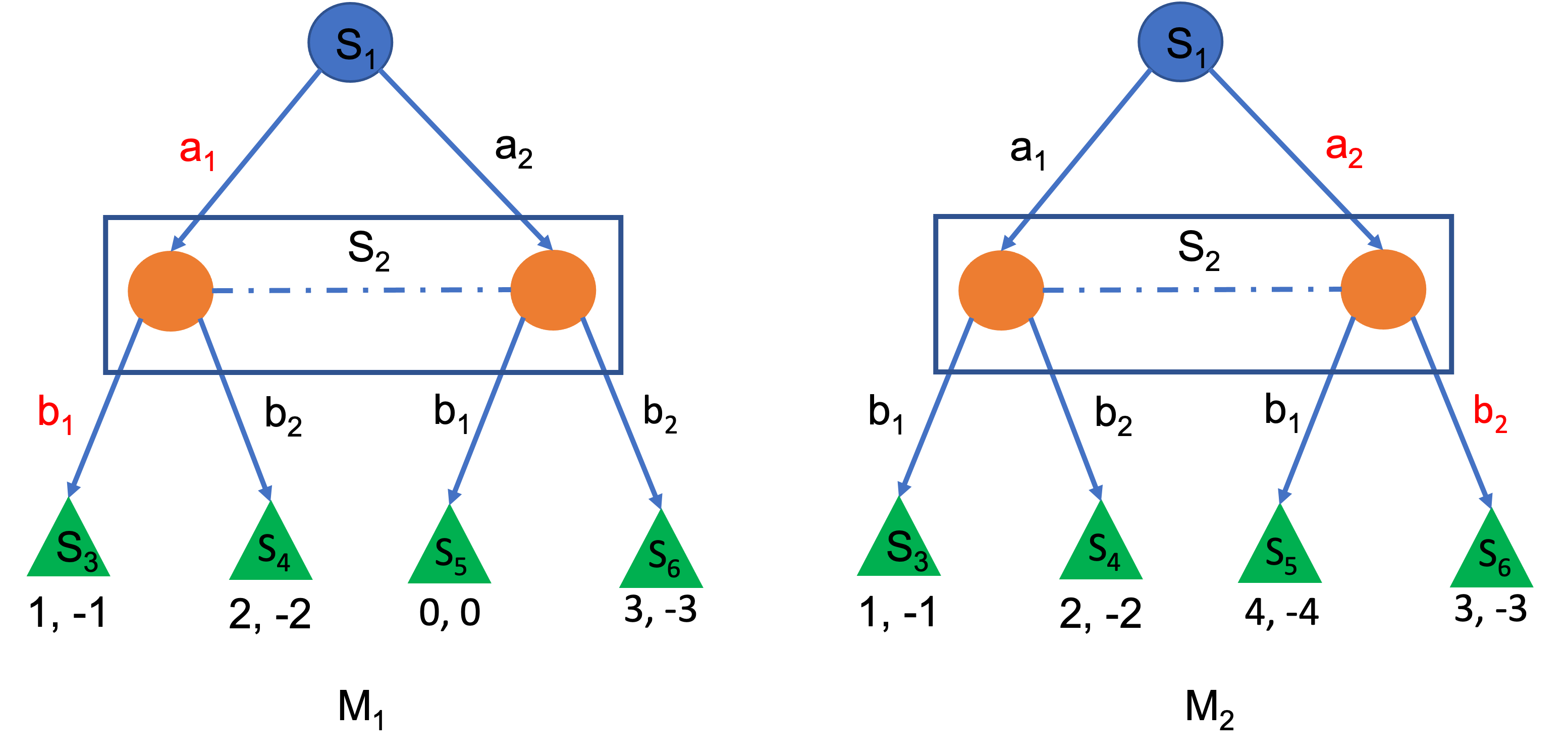}
        \caption{Example of two-player extensive-form game}
        \label{example}
\end{figure}

As demonstrated in offline RL papers \cite{rashidinejad2021bridging,xie2021bellman}, a coverage condition over the optimal policy is sufficient for the offline learning of MDPs. Therefore, it is straightforward to extend this coverage condition to our offline equilibrium finding settings. The following assumption shows this extended coverage condition.
\begin{assumption}(Single Strategy Coverage) The Nash equilibrium strategy $\sigma^*$ is covered by the dataset. 
\end{assumption}
Then there is a question of whether the single strategy coverage assumption over the offline dataset is also sufficient for computing NE strategy under the OEF setting. The answer is no and we use the following theorem to explain the reason. 
\begin{theorem}
    Single strategy coverage assumption over offline dataset is not sufficient for computing an NE strategy. 
\label{theorem_single}
\end{theorem}
\begin{proof}
    We provide a counter-example to prove this theorem. Here, we consider two two-player extensive-form games $M_1$ and $M_2$, which are represented in Figure \ref{example}. 

    We can easily find that the NE of the game $M_1$ is strategy profile $\sigma^{1} = (\sigma^1_1, \sigma^1_2) = (\{S_1: a_1\}, \{S_2: b_1\})$, i.e., player 1 plays $a_1$ at information set $S_1$ and player 2 plays $b_1$ at information set $S_2$. The NE of the game $M_2$ is strategy profile $\sigma^{2} = (\sigma^2_1, \sigma^2_2) = (\{S_1: a_2\}, \{S_2: b_2\})$. Now we consider an offline dataset $D$ which is generated using a strategy profile $\sigma_D$ and the $\sigma_D$ is set to be the uniform distribution on the strategy profiles $\sigma^1$ and $\sigma^2$. 
    
    The dataset $D$ covers strategy profile $\sigma^1$ and $\sigma^2$. Therefore, the dataset $D$ satisfies the single strategy coverage assumption for these two games $M_1$ and $M_2$. However, it is impossible for any algorithm to distinguish these two extensive-form games only based on the dataset $D$ since these two games are both consistent on the dataset $D$. 

    Therefore, the single strategy converges assumption over the offline dataset is not sufficient for computing an NE strategy. 
\end{proof}
From the above proof, we know that the single strategy coverage assumption over the dataset is sufficient for computing the optimal strategy under the offline RL setting while it is not sufficient for computing an NE strategy under the OEF setting. The intuition behind this theorem is that in an offline RL setting, we can easily use the data of two actions to decide which action is better, whereas, in an OEF setting, we cannot use data from only two action pairs to know which action pair is closer to NE, because identifying NE requires other action pairs as inferences. 
Based on this analysis, \cite{cui2022offline} et al. provide a minimal coverage assumption over the dataset which is sufficient for computing an NE strategy in the two-player zero-sum Markov games. 

\begin{assumption}
    (Unilateral Coverage) For all strategy $\sigma_{i}$, ($\sigma_{i}, \sigma_{-i}^*$) for all player $i$ are covered by the dataset, where $\sigma^*=(\sigma_1^*, ..., \sigma_n^*)$ is the NE strategy. 
\end{assumption}
\begin{assumption}
    (Deterministic Unilateral Coverage) For all deterministic strategy $\sigma_{i}$, ($\sigma_{i}, \sigma_{-i}^*$) for all player $i$ are covered by the dataset, where $\sigma^*=(\sigma_1^*, ..., \sigma_n^*)$ is the NE strategy. 
\end{assumption}
We can easily find that deterministic unilateral coverage assumption is equivalent to unilateral coverage assumption. The intuition behind this finding is that any mixed strategy can be represented by a combination of several deterministic strategies. Therefore, if all the deterministic strategies are covered by the dataset, then all mixed strategies are also covered. Based on this finding, in the following proof, we only consider all deterministic strategies.

\citet{cui2022offline} have proved that unilateral coverage assumption is the minimal assumption which is sufficient for computing an NE strategy in the two-player zero-sum Markov games. However, this conclusion is not hold for our model-based framework in computing the equilibrium strategy under the OEF setting. In other words, under the OEF setting, our model-based algorithm cannot guarantee coverage to the equilibrium strategy of the underlying game based on the dataset satisfying the unilateral coverage assumption. 
\begin{theorem}
    The unilateral coverage assumption over the offline dataset is not sufficient for our model-based algorithm to converge to the equilibrium strategy of the underlying game in the OEF setting.
    \label{unilateral}
\end{theorem}
\begin{proof}
    We prove it by providing a counter-example. Here, we consider an imperfect-information extensive-form game $M_3$, which is represented in Figure \ref{example_2}. We can easily find that the NE strategy of game $M_3$ is the strategy profile $\sigma^*=(\sigma_1, \sigma_2)=(\{S_1: a_1\}, \{S_2: b_1\})$. 
    \begin{figure}[h]
        \centering
        \includegraphics[width=0.7\textwidth]{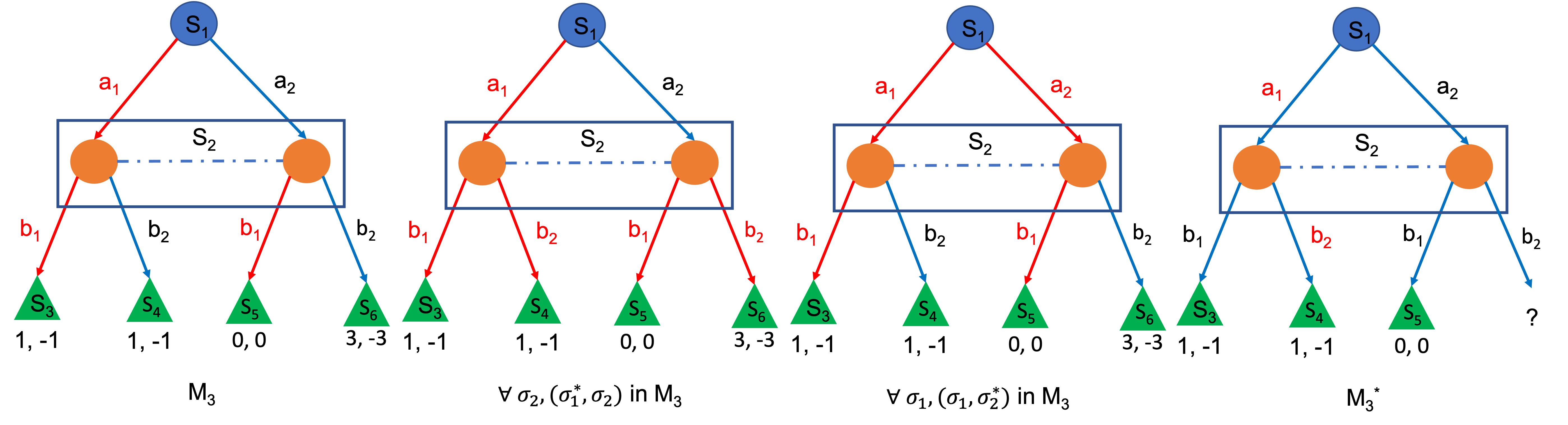}
        \caption{Example of two-player extensive-form game}
        \label{example_2}
    \end{figure}

    To build an offline dataset satisfying the unilateral coverage assumption, the dataset needs to cover $(\sigma_1^*, \sigma_2)$ for all $\sigma_2$ and $(\sigma_1, \sigma_2^*)$ for all $\sigma_1$. We show the state-action pairs covered by these strategy profiles in Figure \ref{example_2}. These red lines show these covered state-action pairs. It means that the dataset satisfying the unilateral coverage assumption would cover these state-action pairs. When applying our model-based framework, the first step is to train an environment model based on the offline dataset. Assume that the environment model can be trained well which means that the environment model can precisely represent all these state-action pairs in the dataset. Therefore, the game represented by the trained environment model would be $M_3^*$ in Figure \ref{example_2}. Note that there are missing data in the game. Although our trained environment model can give approximate results for these missing data, it may result in a different equilibrium strategy. For example, if the missing value in $M_3^*$ is $(0, 0)$ or $(-1, 1)$, then the strategy profile $\sigma=(\sigma_1, \sigma_2^{'})=(\{S_1: a_1\}, \{S_2: b_2\})$ would be the NE strategy of game $M_3^*$. However, the strategy profile $\sigma$ is not the NE strategy for the original game $M_3$. Therefore, the unilateral coverage assumption over the offline dataset is not sufficient for our model-based framework to converge to the NE strategy of the underlying game.
\end{proof}

Therefore, the unilateral coverage assumption is not sufficient for our model-based framework to converge to the equilibrium strategy. To guarantee the convergence of our model-based framework, we provide a minimal dataset coverage assumption for our model-based algorithm to converge to the equilibrium strategy of the underlying game under the OEF setting.
\begin{assumption} (Uniform Coverage)
    For all state $s_t$ and all actions $a_t \in A(s_t)$, all state-action pairs $(s_t, a_t, s_{t+1})$ are covered by the dataset. 
    \label{uniform}
\end{assumption}
\begin{theorem}
    The uniform coverage assumption over the offline dataset is the minimal dataset coverage assumption which is sufficient for our model-based algorithm to converge to the equilibrium strategy in the OEF setting.
    \label{uniform_theorem}
\end{theorem}
\begin{proof}
    From the example in the proof of Theorem \ref{unilateral}, we find that a slight violation of the uniform coverage assumption will impede the computation of the NE strategy using our model-based algorithm. In other words, any state-action pair that is not covered by the dataset would impede the restructure of the game using our environment model. 

    Once the dataset satisfies the uniform coverage, then it covers all the state-action pairs in the game which is enough for training the environment model. It means that the environment model would be the same as the underlying game of the dataset. Then applying our model-based equilibrium finding algorithm on the trained environment model definitely can converge to the equilibrium strategy of the underlying game in the OEF setting.
\end{proof}

Here, we proved that the uniform dataset coverage assumption is sufficient for our model-based framework to converge to the equilibrium strategy. From the proof of Theorem \ref{unilateral}, we find that the game represented by the dataset satisfying the unilateral coverage assumption may be a part of the original game (here, we call the game in the dataset subgame). However, the non-uniqueness of the equilibrium in the subgame would result in the failure to find the equilibrium strategy of the underlying game using our model-based framework. The following theorem provides more analysis of the unilateral coverage assumption in the OEF setting.

\begin{theorem}
    Under the assumption that the equilibrium strategy profile of the game represented by the dataset is unique, the unilateral coverage assumption would be the minimal assumption over the offline dataset which is sufficient for computing an NE strategy in the OEF setting.
\end{theorem}
\begin{proof}
    Firstly, we prove that a slight violation of the unilateral coverage assumption will impede the computation of the Nash equilibrium strategy. We can reuse the example game $M_1$ in the proof of Theorem \ref{theorem_single} and consider another dataset $D$ which is generated using strategy profile $\sigma_D$ and $\sigma_D$ is set to be the uniform distribution on these three deterministic strategy profiles $\sigma^{1} = (\sigma^1_1, \sigma^1_2) = (\{S_1: a_1\}, \{S_2: b_1\})$, $\sigma^{2} = (\sigma^2_1, \sigma^2_2) = (\{S_1: a_2\}, \{S_2: b_1\})$ and $\sigma^{3} = (\sigma^1_1, \sigma^2_2) = (\{S_1: a_2\}, \{S_2: b_2\})$. Since the NE strategy of game $M_1$ is strategy profile $\sigma^{1} = (\sigma^1_1, \sigma^1_2) = (\{S_1: a_1\}, \{S_2: b_1\})$, we can find that only the deterministic strategy profile $\sigma^{4} = (\sigma^2_1, \sigma^1_2) = (\{S_1: a_2\}, \{S_2: b_1\})$ is not covered by the dataset $D$ compared with the dataset satisfying the unilateral coverage assumption. Then the game generated by the dataset is represented in Figure \ref{fig:my_label}.
    \begin{figure}[h]
        \centering
        \includegraphics[width=0.6\textwidth]{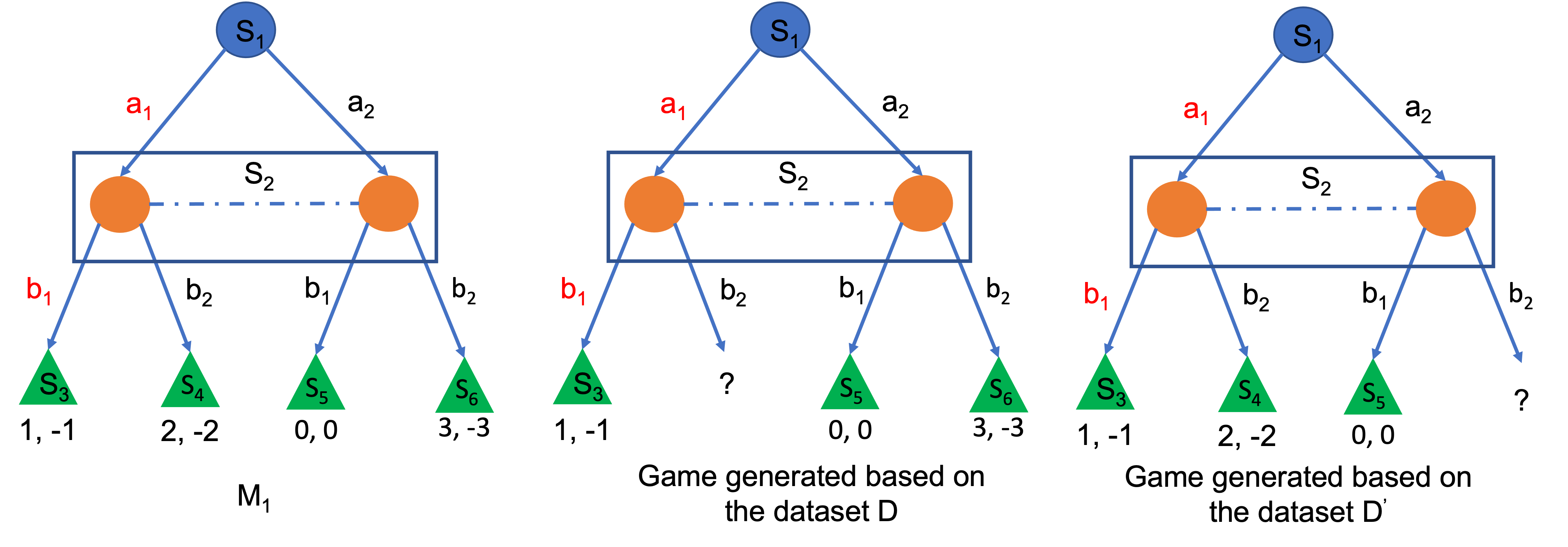}
        \caption{Example game}
        \label{fig:my_label}
    \end{figure}
    we can find that the game generated based on the dataset has the unique equilibrium strategy $\sigma^{1} = (\sigma^1_1, \sigma^1_2) = (\{S_1: a_1\}, \{S_2: b_1\})$. Therefore, the dataset $D$ satisfies the assumption the game generated based on the dataset has a unique equilibrium and slightly violates the unilateral coverage assumption. 
    However, we find that the different missing data values in the game generated based on the dataset would result in a different equilibrium strategy. For example, if the missing value in the game generated based on the dataset is (0, 0), then the equilibrium strategy profile of the game would be $\sigma^* =(\sigma_1, \sigma_2) = \{S_1: \{a_1: 0.75, a_2: 0.25\}, S_2: \{b_1: 0.75, b_2: 0.25\}\}$, which is not the equilibrium strategy of the original game. Therefore, a slight violation of the unilateral coverage assumption will impede the computation of the equilibrium strategy.  
    

    Then we prove that the unilateral coverage assumption is sufficient for computing an NE strategy in the OEF setting under the unique equilibrium assumption. Recall the definition of NE strategy, the strategy profile $\sigma^*$ forms an NE strategy if $u_i(\sigma^*) \geq u_i(\sigma'_i, \sigma^*_{-i}), \forall i \in N, \forall \sigma'_i \in \Sigma_i$ which means that $\sigma_i^*$ is the best response strategy against $\sigma_{-i}^*$ for $\forall i \in N$. According to the unilateral coverage assumption, the dataset covers all strategy profiles $(\sigma_{i}, \sigma_{-i}^*)$ for all $i$ and all $\sigma_{i}$. Then it is easy to verify which strategy for player $i$ is the best response strategy against $\sigma_{-i}^*$ based on the dataset. In other words, we have enough information about $(\sigma_i, \sigma_{-i}^*), \forall \sigma_i$ which is sufficient to verify that $\sigma_i^*$ is the best response strategy of $\sigma_{-i}^*$. In this way, we can verify the best response strategy for every player. Due to the uniqueness of the equilibrium strategy, the strategy $\sigma^*$ would also be the equilibrium strategy of the original game. We can give an example to further explain it. Consider another dataset $D^{'}$ for the game $M_1$ which is generated using the strategy profile $\sigma_D^{'}$ and $\sigma_D^{'}$ is set to be the uniform distribution on these three deterministic strategy profiles $\sigma^{1} = (\sigma^1_1, \sigma^1_2) = (\{S_1: a_1\}, \{S_2: b_1\})$, $\sigma^{2} = (\sigma^2_1, \sigma^2_2) = (\{S_1: a_2\}, \{S_2: b_1\})$ and $\sigma^{3} = (\sigma^1_1, \sigma^2_2) = (\{S_1: a_1\}, \{S_2: b_2\})$. We can easily verify that the dataset satisfies the unilateral coverage assumption for the game $M_1$ and the game generated based on the dataset $D^{'}$ (Figure \ref{fig:my_label}) has a unique equilibrium strategy, $\sigma^{1} = (\sigma^1_1, \sigma^1_2) = (\{S_1: a_1\}, \{S_2: b_1\})$. Then we can find that whatever the missing value in the game is the equilibrium of the game would not change and is the same as the equilibrium strategy of the original game. 
    Therefore, based on the above analysis, under the strong assumption (equilibrium uniqueness), the unilateral coverage assumption would be the minimal dataset coverage assumption.
\end{proof}

The above theorem proves that under the strong assumption (equilibrium uniqueness), the dataset satisfying the unilateral coverage assumption is sufficient for the computation of equilibrium strategy under the OEF setting. However, in the general OEF setting, to guarantee convergence under the dataset satisfying unilateral coverage assumption, it may need a more powerful algorithm that can solve the non-uniqueness of the equilibrium problem. We left it as future work.


So far, we have provided the minimal dataset coverage assumption for our model-based framework to converge to the equilibrium strategy. Then we move to analyze our proposed datasets and their influences on our proposed OEF algorithm. Here, we first provide two assumptions on our proposed datasets based on the generation process of the dataset. 

Since we only use the NE strategy to generate the expert dataset, we can have the following assumption.
\begin{assumption} 
The expert dataset only covers the NE strategy, i.e., the strategy profile used to generate the expert dataset is the NE strategy.
\label{expert}
\end{assumption}
Note that according to the above assumption, we can find that although the expert dataset satisfies the single strategy coverage assumption, it is more strict than the single strategy coverage assumption since the expert dataset only covers the NE strategy. From the empirical results on the expert dataset, we found that the model-based algorithm indeed cannot converge to the NE strategy. However, the behavior cloning algorithm can get a good strategy on the expert dataset since it can mimic the strategy used to generate the expert dataset, i.e., the NE strategy.   

The random dataset is sampled by the uniform strategy. Therefore, it would involve all the state transitions and we can have the following assumption for the random dataset.
\begin{assumption} 
The random dataset satisfies the uniform dataset coverage assumption, i.e., for $\forall s_t$ and $\forall a \in A(s_t)$, $(s_t, a, s_{t+1})$ is covered by the random dataset.
\label{random}
\end{assumption}
Since the random dataset satisfies the uniform dataset coverage assumption, according to Theorem \ref{uniform_theorem}, the random dataset is sufficient for our model-based algorithm to compute the NE strategy. From the empirical results, we can find that the model-based algorithm performs best under the random dataset, which verifies that the random dataset is sufficient for computing the NE strategy. Next, we will provide more analysis of the relationship between the algorithm and the dataset.   

From the empirical analysis, we find that the performance of the model-based algorithm mainly depends on the gap between the trained environment model and the actual game environment. It means that if the trained environment model can recover all the dynamics of the actual game, then the performance is good. Otherwise, the performance is worse. Since our model-based framework can generalize existing equilibrium finding algorithms to the context of the OEF setting and the performance of the existing equilibrium finding algorithm would also determine the convergence of the equilibrium strategy, we assume that there always exists an equilibrium finding algorithm for any game which can converge to the equilibrium strategy in the following proof. Then we have the following theorem. 
\begin{theorem}
Assuming that the environment model is well-trained on the offline dataset, the model-based framework can converge to equilibrium strategy under the random dataset satisfying Assumption \ref{random} and cannot guarantee to converge under the expert dataset satisfying Assumption \ref{expert}. 
\label{mb_theorem}
\end{theorem}
\begin{proof}
Since the environment model is well-trained on the offline dataset, the environment model can fully represent the information of the offline dataset. If the random dataset is the offline dataset, the game defined by the trained environment model is the same as the actual game. The reason is that every state transition is covered by the random dataset according to Assumption \ref{random}. Then the strategy learned by our model-based equilibrium finding algorithm is the approximate equilibrium strategy of the actual game due to the convergence property of the original equilibrium finding algorithm. Therefore, the model-based framework can converge to an equilibrium strategy under the random dataset satisfying Assumption \ref{random}. 

If the offline dataset is the expert dataset, then the dataset only covers these state transitions related to the NE strategy according to Assumption \ref{expert}. Therefore, the state transition of the actual game may not be covered by the expert dataset. The environment model trained based on the expert dataset would produce different transition information on these states not shown in the dataset compared with the actual game. It would cause a gap between the trained environment model and the actual game. Although the model-based framework can learn an approximate equilibrium strategy of the game defined by the environment model, there is no guarantee that the learned strategy is the equilibrium strategy of the actual game. 
\end{proof}

Theorem \ref{mb_theorem} is consistent with our previous conclusion that single strategy coverage is insufficient for NE identification, and dataset coverage satisfying Assumption \ref{uniform} is sufficient for NE identification according to Theorem \ref{uniform_theorem}. And our empirical results also verify these conclusions. The model-based framework performs best under the random dataset and worst under the expert dataset.

Although the expert dataset satisfies the single strategy coverage, the expert dataset assumption is more strict than the single strategy coverage. We find that the behavior cloning algorithm can perform well on the expert dataset. Therefore, to offset the drawback of the model-based algorithm under the expert dataset, we propose to combine the behavior cloning (BC) technique. From the introduction of the BC technique, we know that the BC can mimic the behavior policy in the dataset. Therefore, we have the following theorem describing the power of the BC technique.

\begin{theorem}
Assuming that the behavior cloning policy is well-trained on the offline dataset, the behavior cloning technique can get the equilibrium strategy under the expert dataset satisfying Assumption \ref{expert}, and cannot get the equilibrium strategy under the random dataset satisfying Assumption \ref{random}.
\label{bc_theorem}
\end{theorem}
\begin{proof}
The assumption that the behavior cloning policy is well-trained on the offline dataset means that the behavior cloning policy can precisely mimic the behavior strategy used to generate the offline dataset. If the offline dataset is the expert dataset, according to Assumption \ref{expert}, the behavior strategy used to generate the expert dataset is the NE strategy. Therefore, applying the behavior cloning algorithm on the expert dataset can get an NE strategy. 

If the offline dataset is the random dataset, according to the generation process of the random dataset and Assumption \ref{random}, the behavior strategy used to generate the random dataset is a uniform strategy. Therefore, the behavior cloning algorithm can only get a uniform strategy instead of the equilibrium strategy under the random dataset. 
\end{proof}

Our experimental results also show the same outcomes as Theorem \ref{bc_theorem}. The performance of the behavior cloning technique mainly depends on the quality of the behavior strategy used to generate the offline dataset. Therefore, the behavior cloning technique can perform well under the expert dataset. 
Based on the above two theorems, we propose our OEF algorithm, BC+MB, by combining the above two techniques with different weights to improve the performance under these datasets with unknown behavior strategies. 

\begin{theorem}
Under the assumptions in Theorems \ref{mb_theorem} and \ref{bc_theorem}, BC+MB can compute the equilibrium strategy under either the random dataset satisfying Assumption \ref{random} or the expert dataset satisfying Assumption \ref{expert}.
\label{mb_bc_theorem}
\end{theorem}
\begin{proof}
In the BC+MB algorithm, the weight of the BC policy is represented by $\alpha$. The weight of the MB policy is $1 - \alpha$.
The $\alpha$ ranges from 0 to 1. When under the random dataset satisfying Assumption \ref{random}, let $\alpha$ equal 0. Then the policy of BC+MB would equal to MB policy, i.e., the policy trained using the model-based algorithm. According to Theorem \ref{mb_theorem}, the model-based framework can converge to an equilibrium strategy under the random dataset satisfying Assumption \ref{random}. Therefore, BC+MB can also converge to an equilibrium strategy under the random dataset satisfying Assumption \ref{random}.

When under the expert dataset satisfying Assumption \ref{expert}, let $\alpha$ equal to 1. Then the policy of BC+MB would be equal to BC policy, i.e., the policy trained by behavior cloning algorithm. Similarly, according to Theorem \ref{bc_theorem}, BC+MB can get an equilibrium strategy in the expert dataset satisfying Assumption \ref{expert}.
\end{proof}

Let's move to a more general case in which the offline dataset is generated by a behavior strategy $\sigma$. Then we have the following theorems under the general case.

\begin{theorem}
Assuming that the offline dataset $\mathcal{D}_{\sigma}$ generated by the behavior strategy $\sigma$ covers $(s_t, a, s_{t+1}), \forall s_t, a \in A(s_t)$ and the environment model is well-trained on $\mathcal{D}_{\sigma}$, the model-based framework can converge to an equilibrium strategy that performs equal even better than $\sigma$. 
\label{mb_theorem_2}
\end{theorem}
\begin{proof}
According to the proof of Theorem \ref{mb_theorem}, since every state transition of the actual game is covered by $\mathcal{D}_{\sigma}$, the trained environment model would be the same as the actual game under the assumption that the environment model is well-trained on the offline dataset. Then according to Theorem \ref{mb_theorem}, the model-based framework can converge to an equilibrium strategy. If $\sigma$ used to generate the dataset is not the equilibrium strategy, then the model-based framework can get a better strategy (equilibrium strategy) than $\sigma$. And if $\sigma$ is an equilibrium strategy, then the strategy trained using a model-based framework would perform equal to $\sigma$.
\end{proof}

\begin{theorem}
Assuming that the behavior cloning policy is well-trained on the offline dataset $\mathcal{D}_{\sigma}$ generated by the behavior strategy $\sigma$, the performance of behavior cloning policy $\sigma^{bc}$ would be as good as the performance of $\sigma$. 
\label{bc_theorem_2}
\end{theorem}
\begin{proof}
According to the Assumption \ref{bc_theorem}, behavior cloning can precisely mimic the behavior strategy in the offline dataset. Therefore, $\sigma^{bc}$ would be same as $\sigma$. Consequently, the performance of $\sigma^{bc}$ would have the same performance as $\sigma$.
\end{proof}

\begin{theorem}
Assuming that the environment model and the behavior cloning policy are well-trained, under the offline dataset $\mathcal{D}_{\sigma}$ generated using $\sigma$, BC+MB can get an equal or better strategy than $\sigma$.
\end{theorem}
\begin{proof}
Following the proof of Theorem \ref{mb_bc_theorem}, let $\alpha$ equal 1. Then BC+MB would reduce to BC. Then according to Theorem \ref{bc_theorem_2}, the performance of BC policy is at least as good as $\sigma$. Therefore, BC+MB can get a strategy that is at least as good as the behavior strategy $\sigma$. 

In another extreme case in which $\mathcal{D}_{\sigma}$ covers $(s_t, a, s_{t+1}), \forall s_t, a \in A(s_t)$, let $\alpha$ equal to 0. Then BC+MB would reduce to MB. Then according to Theorem \ref{mb_theorem_2}, the MB policy performs equal to or better than $\sigma$. Therefore, in this case, BC+MB can get an equal or better strategy than $\sigma$.
\end{proof}

In conclusion, under the above assumptions, BC+MB can perform at least equal to the behavior strategy used to generate the offline dataset. The improvement over the behavior strategy mainly depends on the performance of the model-based algorithm under the offline dataset. 

\clearpage
\section{Implementation Details}
\label{app:implementation}
\begin{figure}[ht]
\centering
\vspace{-5pt}
\includegraphics[width=0.6\textwidth]{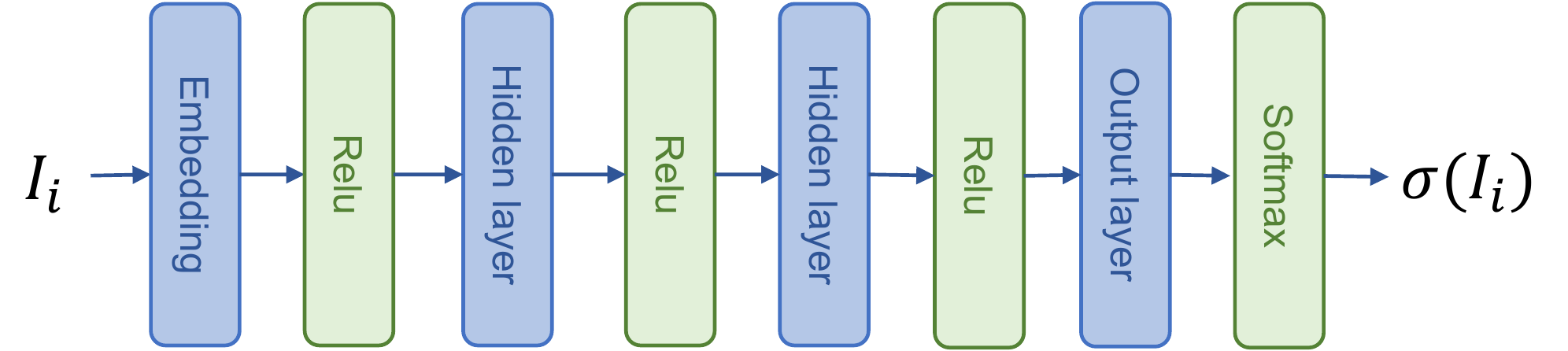}
\caption{Structure of Behavior Cloning policy network.}
\label{fig:bc_model}
\vspace{-5pt}
\end{figure}

\textbf{Behavior Cloning.} Behavior cloning (BC) is a method that mimics the behavior policy in the dataset. Behavior cloning technique is used frequently in offline RL~\cite{fujimoto2021minimalist}. In the OEF setting, we can also use the BC technique to learn a behavior cloning strategy of every player from the offline dataset. More specifically, we can use the imitation learning algorithm to train a policy network $\sigma_i$ parameterized by $\theta$ for every player $i$ to predict the strategy given any information set $I_i$. Only the information sets and action data are needed when training the behavior cloning strategy. We employ the cross-entropy loss as the training loss, defined as
\begin{align}
    \mathcal{L}_{bc} = \mathbb{E}_{(I_i, a) \sim D} [l(a, \sigma_i(I_i;\theta))] = - \mathbb{E}_{(I_i, a) \sim D} [a \cdot \log(\sigma_i(I_i;\theta))],
\end{align}
where $a$ is the action one-hot representation. Figure \ref{fig:bc_model} shows the structure of the behavior cloning policy network. Because equilibrial strategies in most information sets are non-trivial probability distributions, we apply a softmax layer after the output layer to obtain the final mixed strategy.

\textbf{Model-based Framework.} Next, we introduce our instantiate offline model-based algorithms: OEF-PSRO and OEF-CFR, which are adaptions from two widely-used online equilibrium finding algorithms PSRO and Deep CFR, and OEF-JPSRO, which is an adaption from JPSRO. These three algorithms perform on the well-trained environment model $E$. We first introduce the OEF-PSRO algorithm, and the whole flow is shown in Algorithm \ref{Alg:one}. Firstly, we need the well-trained environment model $E$ as input and initialize policy sets $\Pi$ for all players using random strategies. Then, we need to estimate a meta-game matrix by computing expected utilities for each joint strategy profile $\pi \in \Pi$. In vanilla PSRO, to get the expected utility for $\pi$, it needs to perform the strategy $\pi$ in the actual game simulator. However, the simulator is missing in the OEF setting. Therefore, we use the well-trained environment model $E$ to replace the game simulator to provide the information needed in the algorithm. Then we initialize meta-strategies using a uniform strategy. Next, we need to compute the best response policy oracle for every player and add the best response policy oracle to their policy sets. When training the best response policy oracle using DQN or other reinforcement learning algorithms, we sample the training data based on the environment model $E$. After that, we compute missing entries in the meta-game matrix and calculate meta-strategies for the meta-game. To calculate the meta-strategy $\sigma$ of the meta-game matrix, we can use the Nash solver or $\alpha$-rank algorithm. Here, we use the $\alpha$-rank algorithm as the meta solver because our algorithm needs to solve multi-player games. Finally, we repeat the above process until satisfying the convergence conditions. Since the process of JPSRO is similar to PSRO except for the best response computation and meta distribution solver, OEF-JPSRO is also similar to OEF-PSRO. We do not cover OEF-JPSRO in detail here.

\begin{algorithm}[tb]
   \caption{Offline Equilibrium Finding - Policy-Space Response Oracles}
   \label{Alg:one}
\begin{algorithmic}
   \STATE {\bfseries Input:} Trained environment model $E$
   \STATE Initial policy sets $\Pi$ for all players\;
    \STATE Compute expected utilities $U^{\Pi}$ for each joint $\pi\in\Pi$ \textbf{based on the environment model $E$}\;
   \STATE Initialize mate-strategies $\sigma_i$ = \textsc{Uniform}($\Pi_i$) \;
   \REPEAT
   \FOR{player $i \in [1,.., n]$}
   \FOR{best response episodes $p \in [0, ..., t]$}
   \STATE Sample $\pi_{-i} \sim \sigma_{-i}$\;
   \STATE Train best response oracle $\pi_i^{\prime}$ over $\rho \sim (\pi_i^{\prime}, \pi_{-i})$, which \textbf{samples on the environment model $E$}\;
   \ENDFOR
   \STATE Compute missing entries in $U^{\Pi}$ from $\Pi$ \textbf{based on the environment model $E$}\;
   \STATE Compute a meta-strategy $\sigma$ from $U^{\Pi}$ using $\alpha$-rank algorithm;
   \ENDFOR
   \UNTIL{Meet convergence condition}
   \STATE {\bfseries Output:}current solution strategy $\sigma_i$ for player $i$
\end{algorithmic}
\end{algorithm}

\begin{algorithm}[tb]
   \caption{Offline Equilibrium Finding - Deep Counterfactual Regret Minimization}
   \label{deepcfr}
\begin{algorithmic}
   \STATE {\bfseries Input:} Trained environment model $E$
   \STATE Initialize regret network $R(I, a|\theta_{r,p})$ for every player $p$; 
    \STATE Initialize average strategy network $S(I, a|\theta_{\pi, p})$ for every player $p$; 
   \STATE Initialize regret memory $M_{r,p}$ and strategy memory $M_{\pi,p}$ for every player $p$;
   \FOR{CFR Iteration $t=1$ to $T$}
\FOR{player $p \in [1, ..., n]$ }
\FOR{traverse episodes $k \in [1, ..., K]$}
\STATE TRVERSE($\phi$, $p$, $\theta_{r,p}$, $\theta_{\pi, -p}$,$M_{r,p}$, $M_{\pi, -p}$, $E$);
\# use sample algorithm to traverse game tree and record regret and strategy
\ENDFOR
\STATE Train $\theta_{r,p}$ from scratch based on regret memory $M_{r,p}$\;
\ENDFOR
\ENDFOR
\STATE Train $\theta_{\pi,p}$ based on strategy memory $M_{\pi,p}$ for every player $p$;

\STATE {\bfseries Output:}$\theta_{\pi,p}$ for every player $p$
\end{algorithmic}
\end{algorithm}

\begin{algorithm}[t]
\caption{TRVERSE($s, p,$ $\theta_{r, p}$, $\theta_{\pi, -p}$,$M_{r,p}$, $M_{\pi, -p}$, $E$)-External Sampling Algorithm}
\label{traverse}
\begin{algorithmic}
\IF {$s$ is terminal state}
\STATE Get the utility $u_i(s)$ from environment model $E$\;
\STATE {\bfseries Output:} $u_i(s)$
\ELSIF{$s$ is a chance state}
\STATE Sample an action $a$ based on the probability $\sigma_c(s)$, which is obtained from model $E$\;
\STATE $s' = E(s, a)$\;
\STATE {\bfseries Output:} TRAVERSE($s', p,$ $\theta_{r, p}$, $\theta_{\pi, -p}$,$M_{r,p}$, $M_{\pi, -p}$, $E$)
\ELSIF{$P(s) = p$}{
\STATE $I \leftarrow s[p]$; \# game state is formed by information sets of every player \;
\STATE $\sigma(I) \leftarrow$ strategy of $I$ computed using regret values $R(I, a|\theta_{r, p})$ based on regret matching\;
\FOR{$a \in A(s)$}
\STATE $s' = E(s, a)$\;
\STATE $u(a) \leftarrow$ TRAVERSE($s', p,$ $\theta_{r, p}$, $\theta_{\pi, -p}$,$M_{r,p}$, $M_{\pi, -p}$, $E$)\;
\ENDFOR
\STATE $u_{\sigma} \leftarrow \sum_{a \in A(s)} \sigma(I,a)u(a)$\;
\FOR{$a \in A(s)$}
\STATE $r(I, a) \leftarrow u(a) - u_{\sigma}$\;
\ENDFOR
\STATE Insert the infoset and its action regret values $(I, r(I))$ into regret memory $M_{r,p}$\;
\STATE {\bfseries Output:} $u_{\sigma}$}
\ELSE
\STATE $I \leftarrow s[p]$\;
\STATE $\sigma(s) \leftarrow$ strategy of $I$ computed using regret value $R(I,a|\theta_{r, -p})$ based on regret matching\;
\STATE Insert the infoset and its strategy $(I, \sigma(s))$ into strategy memory $M_{\pi, -p}$\;
\STATE Sample an action $a$ from the probability distribution $\sigma(s)$\;
\STATE $s' = E(s, a)$\;
\STATE {\bfseries Output:} TRAVERSE($s', p,$ $\theta_{r, p}$, $\theta_{\pi, -p}$,$M_{r,p}$, $M_{\pi, -p}$, $E$)\;
\ENDIF
\end{algorithmic}
\end{algorithm}

Algorithm~\ref{deepcfr} shows the process of OEF-CFR. It also needs the well-trained environment model $E$ as input. We first initialize regret and strategy networks for every player and then initialize regret and strategy memories for every player. Then we need to update the regret network for every player. To do this, we can perform the traverse function to collect corresponding training data. The traverse function can be any sampling-based CFR algorithm. Here, we use the external sampling algorithm. Note that we need to perform the traverse function on the game tree. In OEF-CFR, the trained environment model can replace the game tree. Therefore, the trained environment model is the input of the traverse function. Algorithm~\ref{traverse} shows the process of the traverse function. In this traverse function, we collect the regret training data of the traveler, and the strategy training data of other players are also gathered. After performing the traverse function several times, the regret network is updated using the regret memory. We need to repeat the above processes $n$ iterations. Then the average strategy network for every player is trained based on its corresponding strategy memory. Finally, the trained average strategy networks are output as the approximate NE strategy. 




\clearpage
\section{Additional Experimental Results}
\label{app:results}
In this part, we provide experimental results on other different games. First, we provide the experimental results of the behavior cloning method and model-based framework (OEF-CFR) based on hybrid datasets, and then the results of our OEF algorithm (BC+MC) are given. We also test our OEF algorithm on a two-player Phantom Tic-Tac-Toe game using the learning dataset. Finally, we provide the ablution study and the setting of hyper-parameters used in our experiments. 

\begin{figure}[ht]
\centering
\subfigure[Kuhn Poker-3 Player]{
\includegraphics[width=0.18\textwidth]{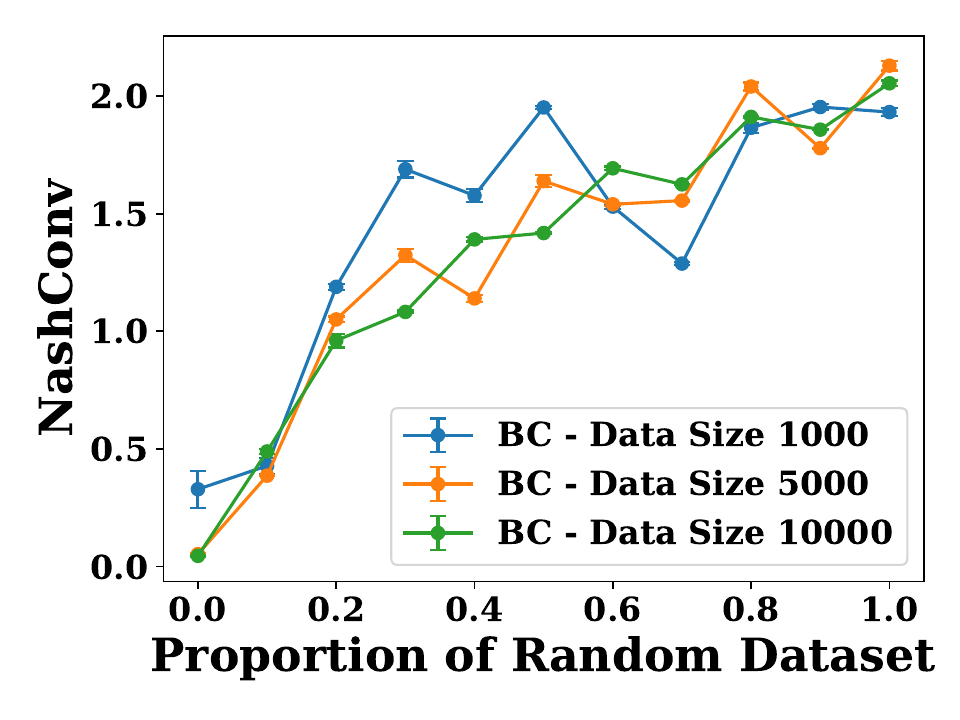}}
\subfigure[Kuhn Poker-4 Player]{
\includegraphics[width=0.18\textwidth]{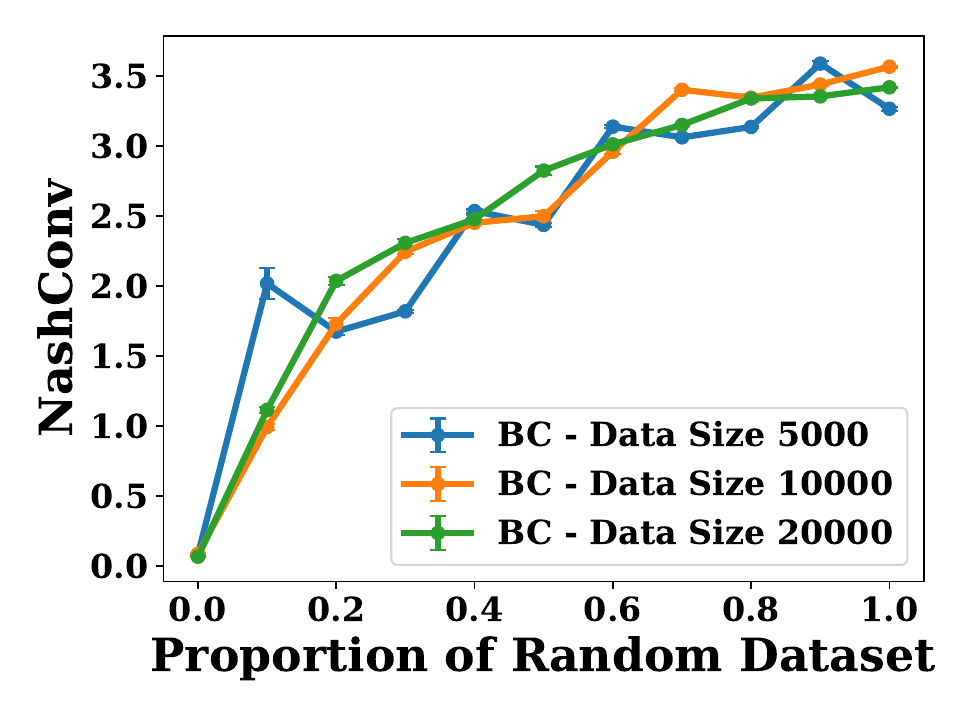}}
\subfigure[Kuhn Poker-5 Player]{
\includegraphics[width=0.18\textwidth]{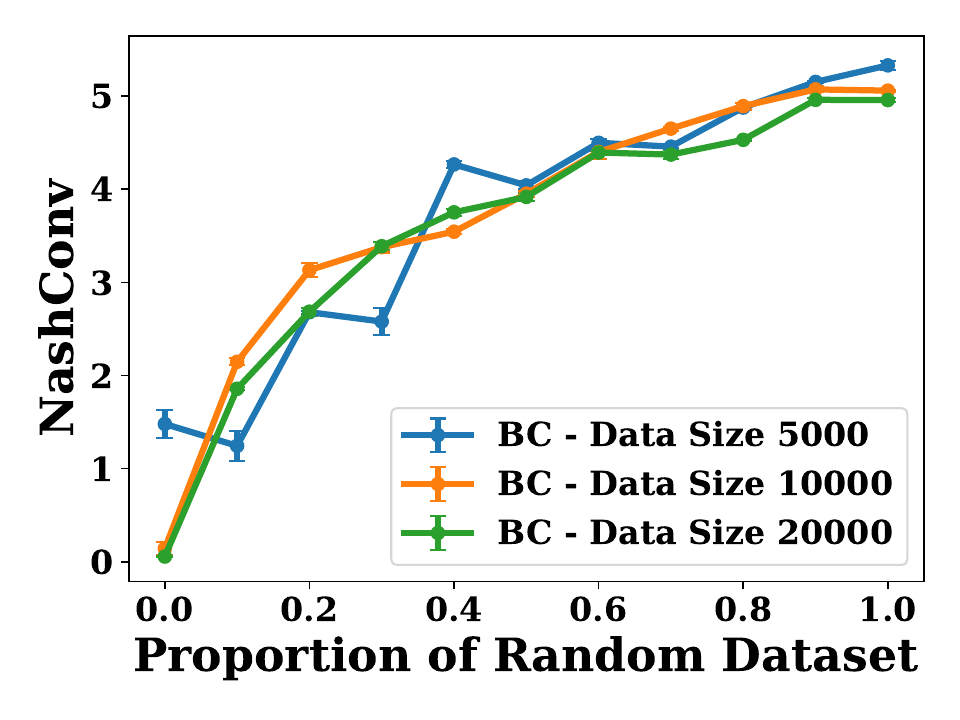}}
\subfigure[Leduc Poker-3 Player]{
\includegraphics[width=0.18\textwidth]{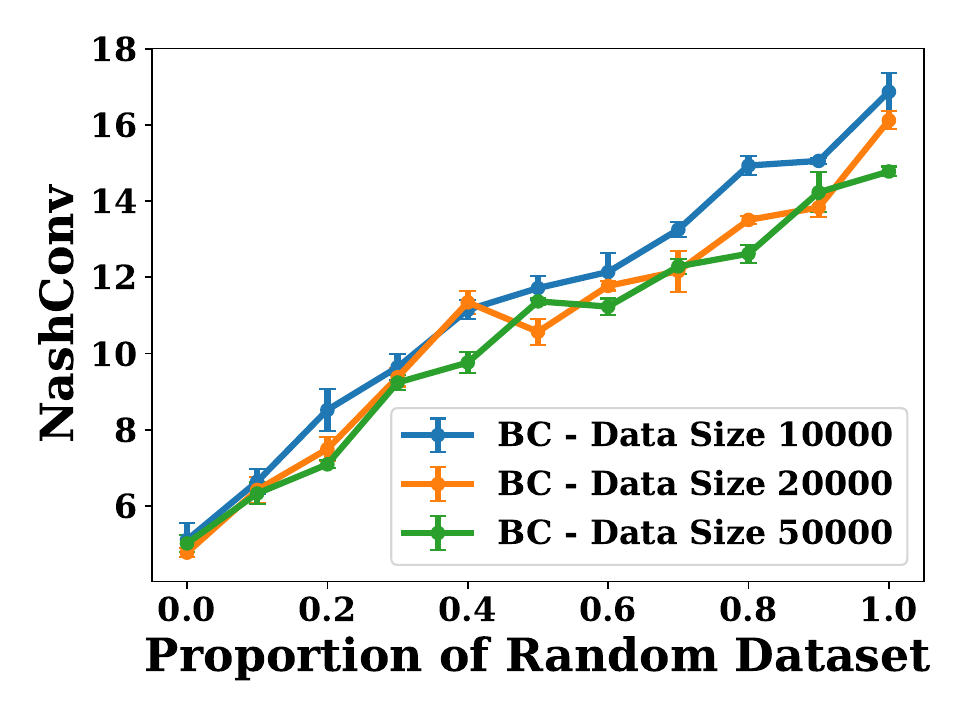}}
\subfigure[Liar's Dice-2 Player]{
\includegraphics[width=0.18\textwidth]{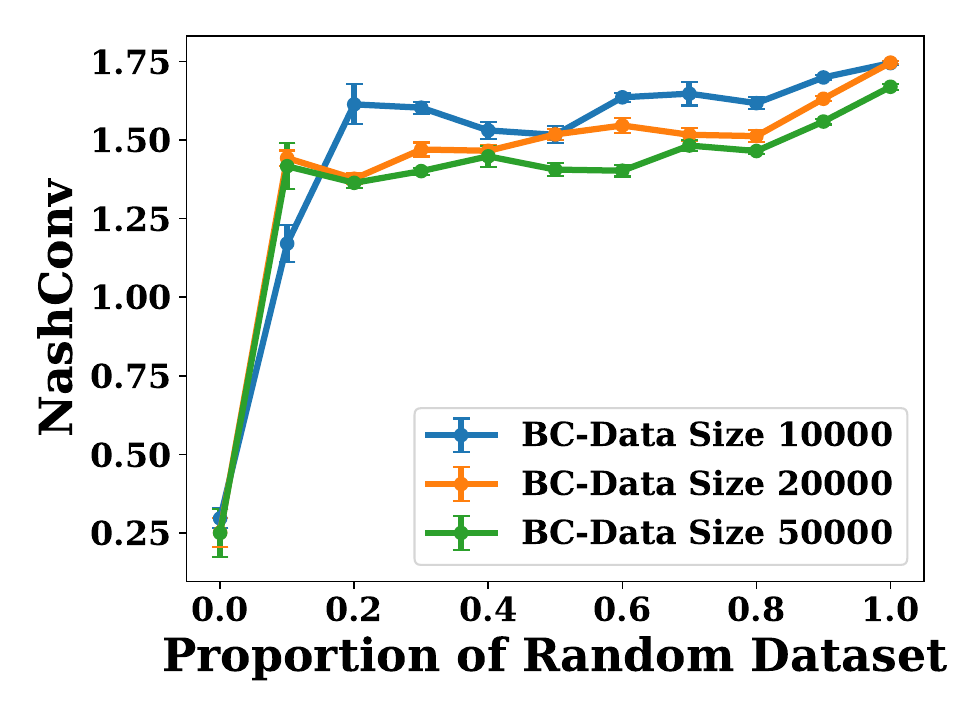}}
\caption{Experimental results for the BC method}
\label{ex_1}
\end{figure}

Figure \ref{ex_1} shows the results of the behavior cloning technique on several multi-player poker games and one two-player Liar's Dice game. It shows that as the proportion of random datasets increases, the performance decreases. It is consistent with the results of previous experiments. 

\begin{figure}[ht]
\centering
\subfigure[Kuhn Poker-3 Player]{
\includegraphics[width=0.18\textwidth]{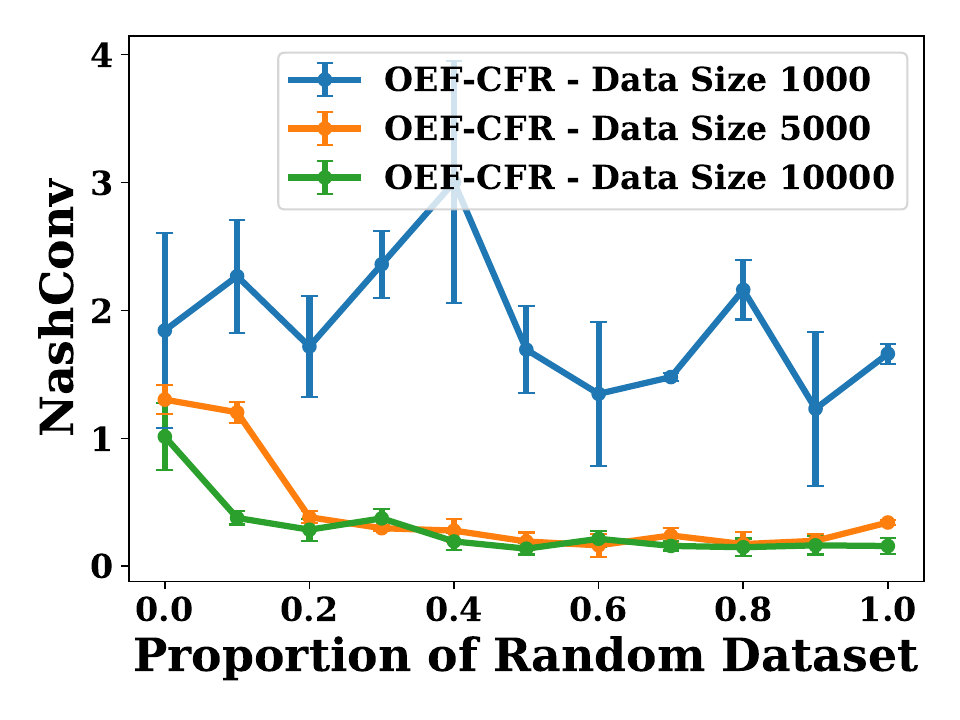}}
\subfigure[Kuhn Poker-4 Player]{
\includegraphics[width=0.18\textwidth]{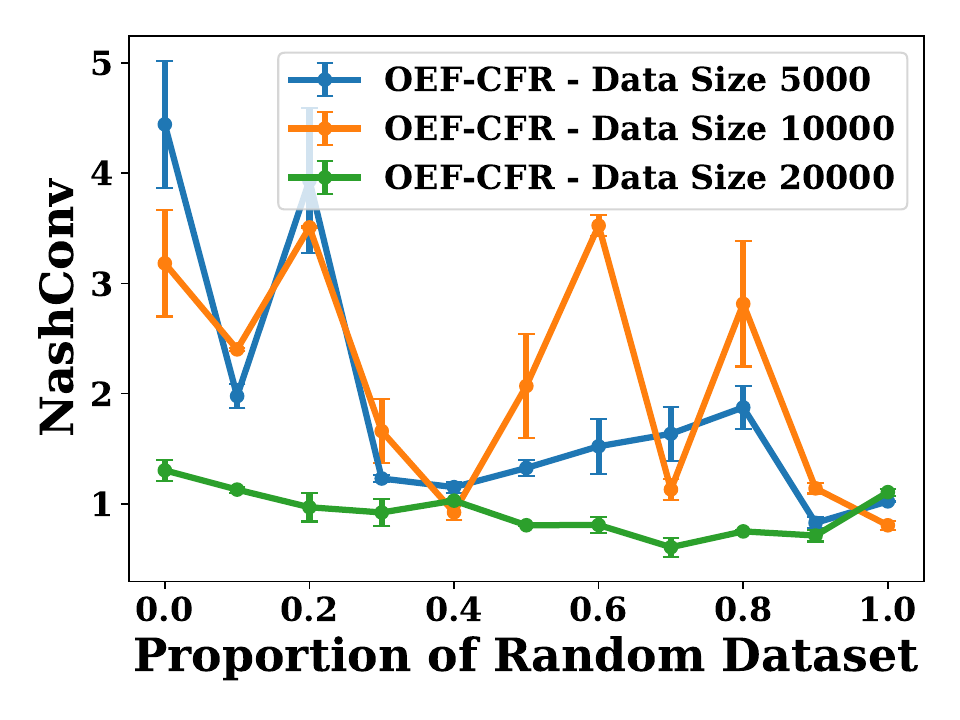}}
\subfigure[Kuhn Poker-5 Player]{
\includegraphics[width=0.18\textwidth]{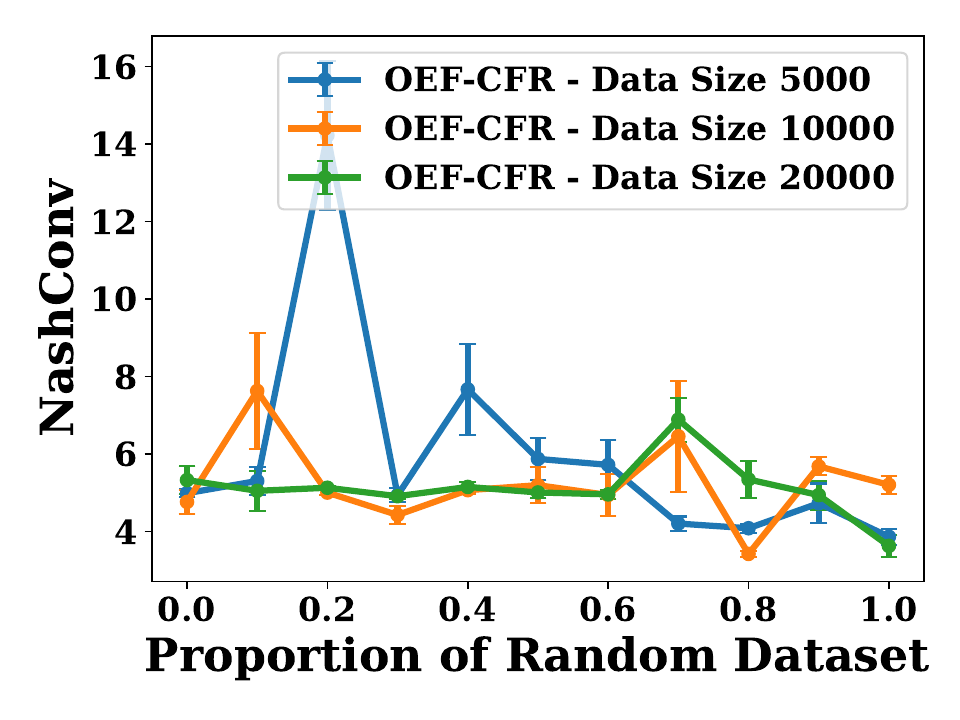}}
\subfigure[Leduc Poker-3 Player]{
\includegraphics[width=0.18\textwidth]{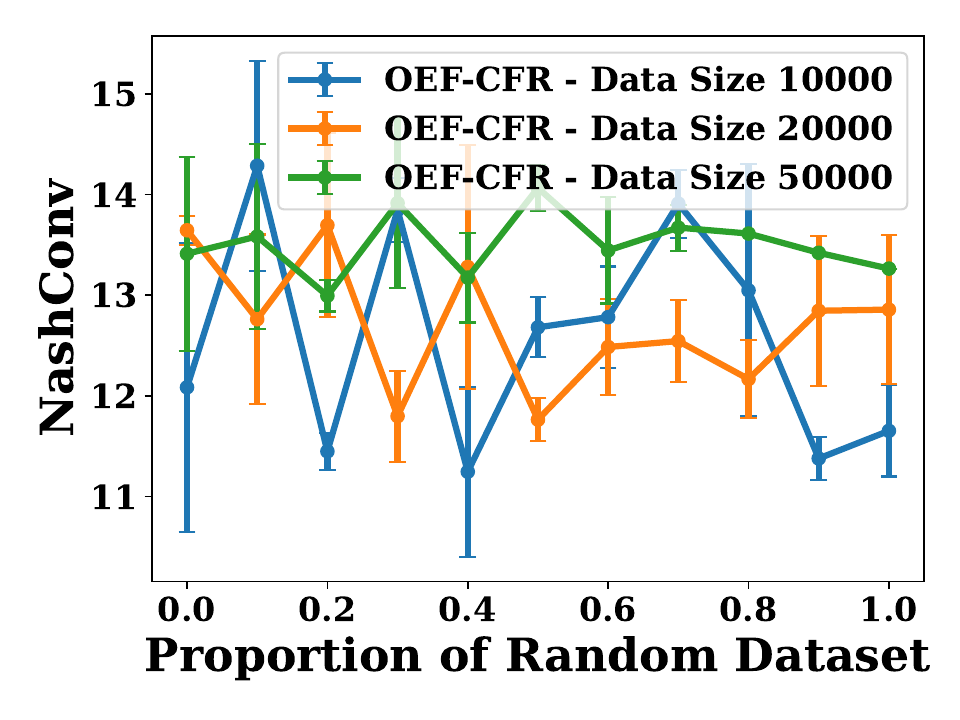}}
\subfigure[Liar's Dice-2 Player]{
\includegraphics[width=0.18\textwidth]{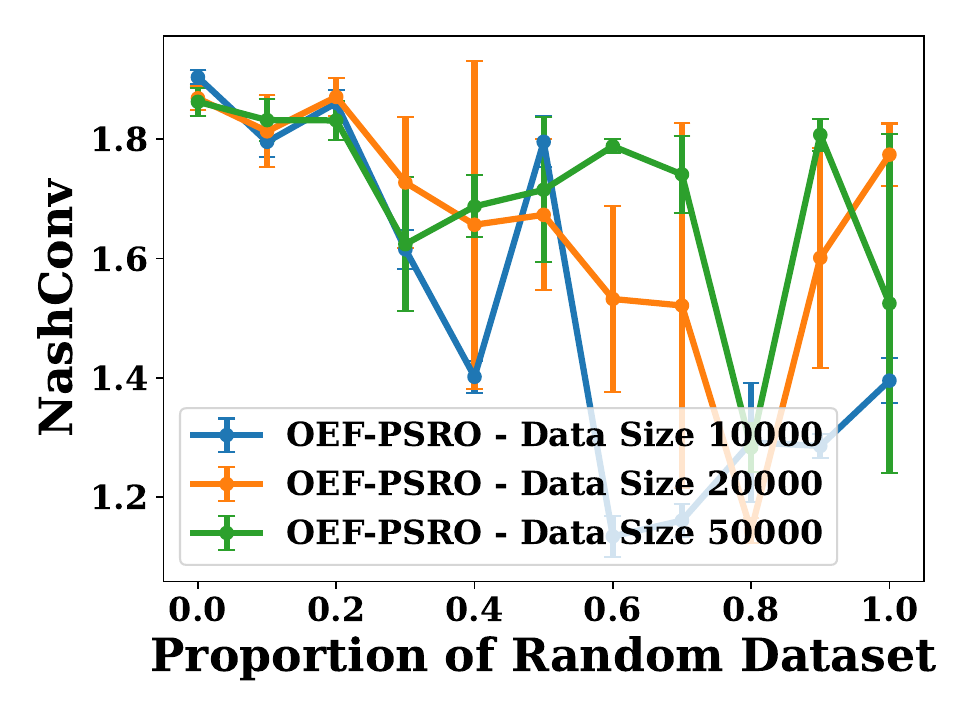}}
\caption{Experimental results for the MB method}
\label{iiiiii}
\end{figure}

\begin{figure}[htb]
\centering
\subfigure[Kuhn Poker-3 Player]{
\includegraphics[width=0.23\textwidth]{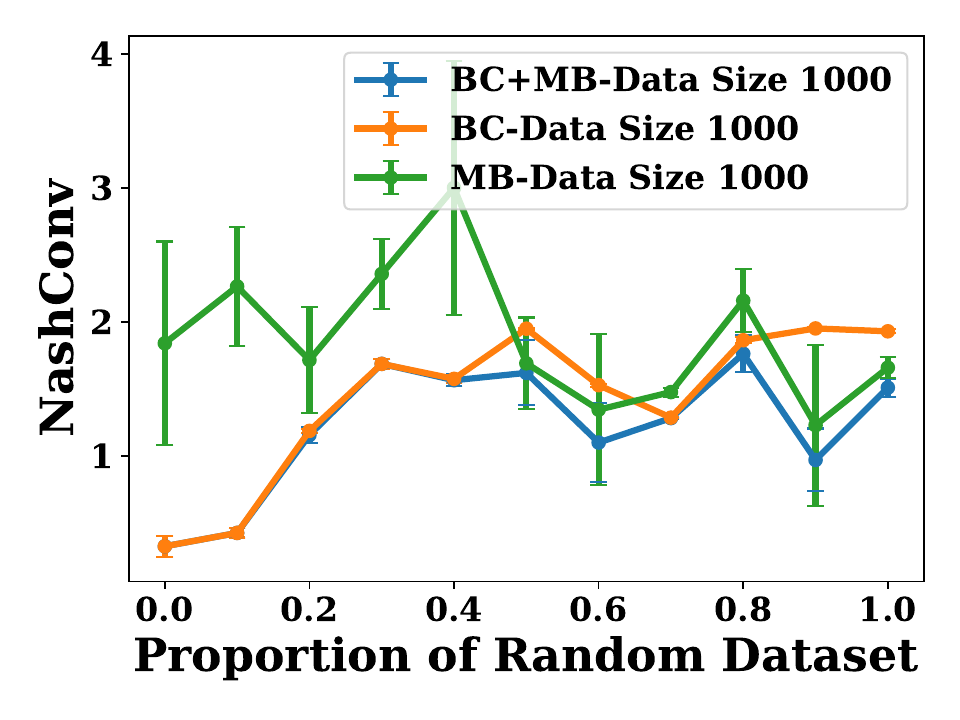}
\label{1}}
\subfigure[Kuhn Poker-3 Player]{
\includegraphics[width=0.23\textwidth]{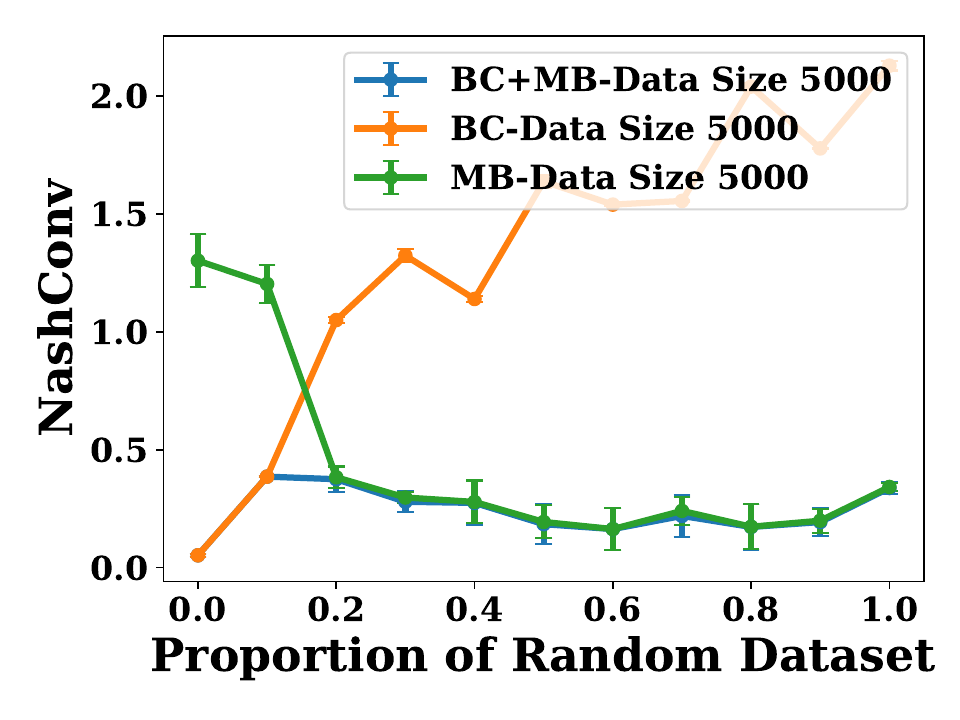}}
\subfigure[Kuhn Poker-4 Player]{
\includegraphics[width=0.23\textwidth]{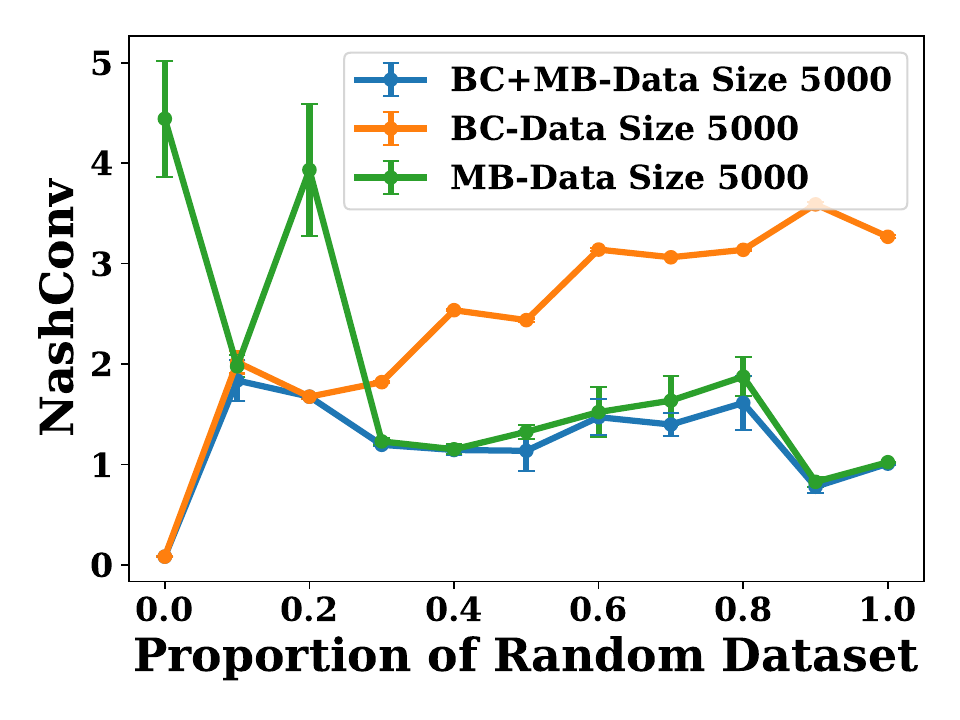}}
\subfigure[Kuhn Poker-4 Player]{
\includegraphics[width=0.23\textwidth]{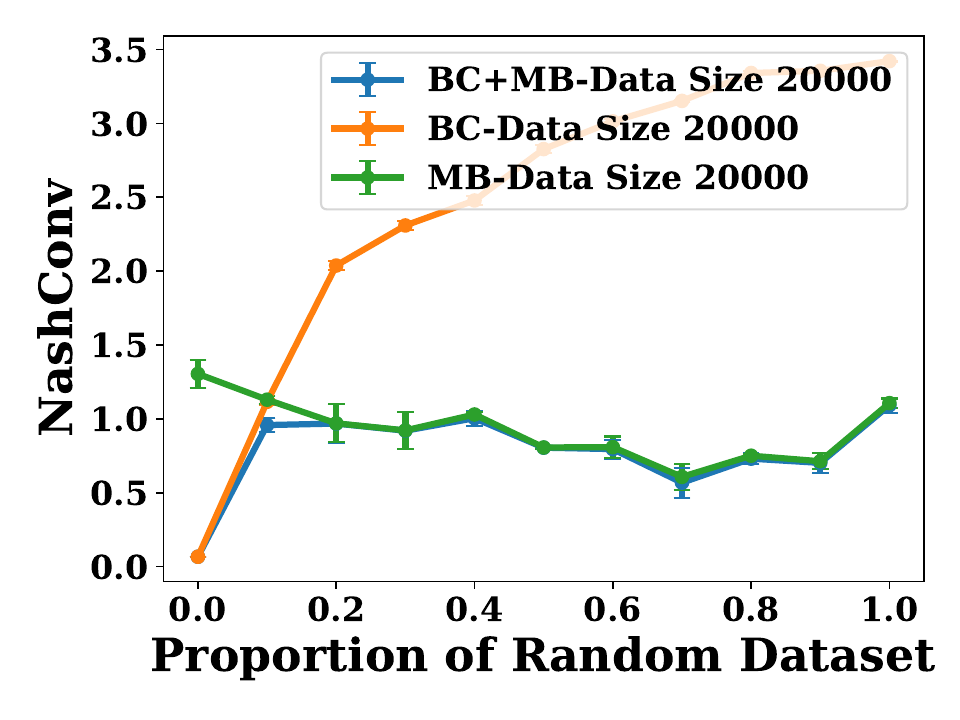}}
\subfigure[Kuhn Poker-5 Player]{\includegraphics[width=0.23\textwidth]{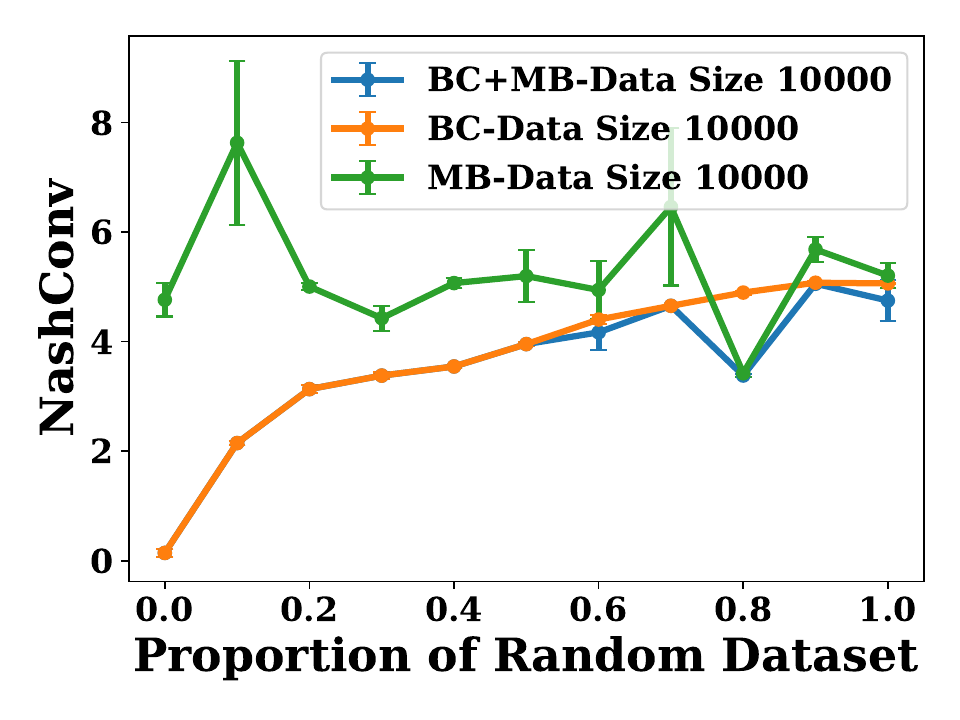}}
\subfigure[Kuhn Poker-5 Player]{\includegraphics[width=0.23\textwidth]{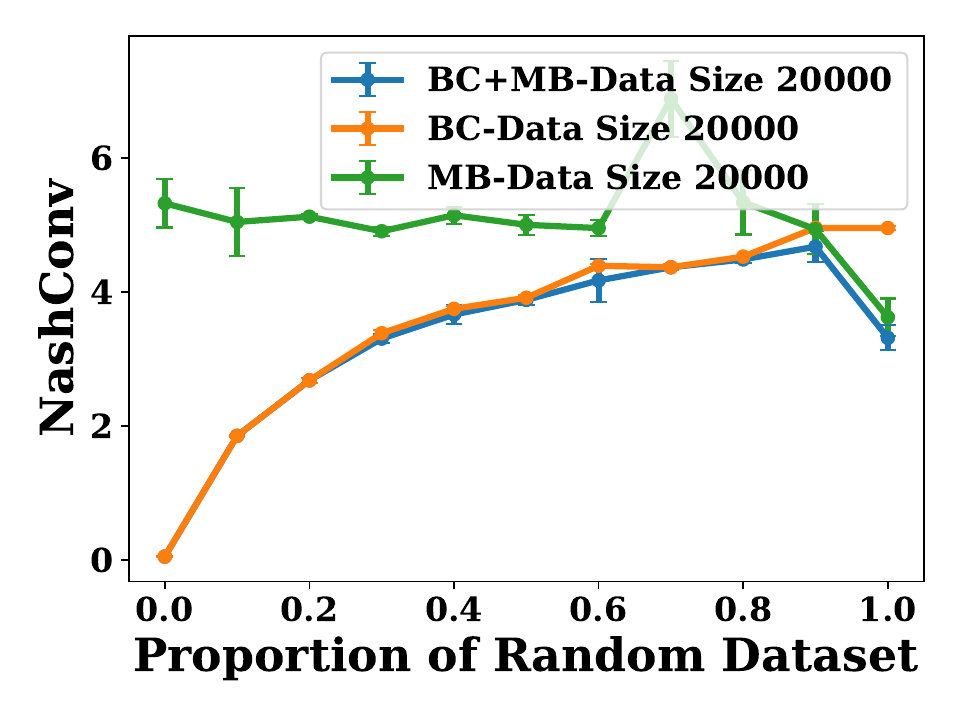}}
\subfigure[Leduc Poker-3 Player]{\includegraphics[width=0.23\textwidth]{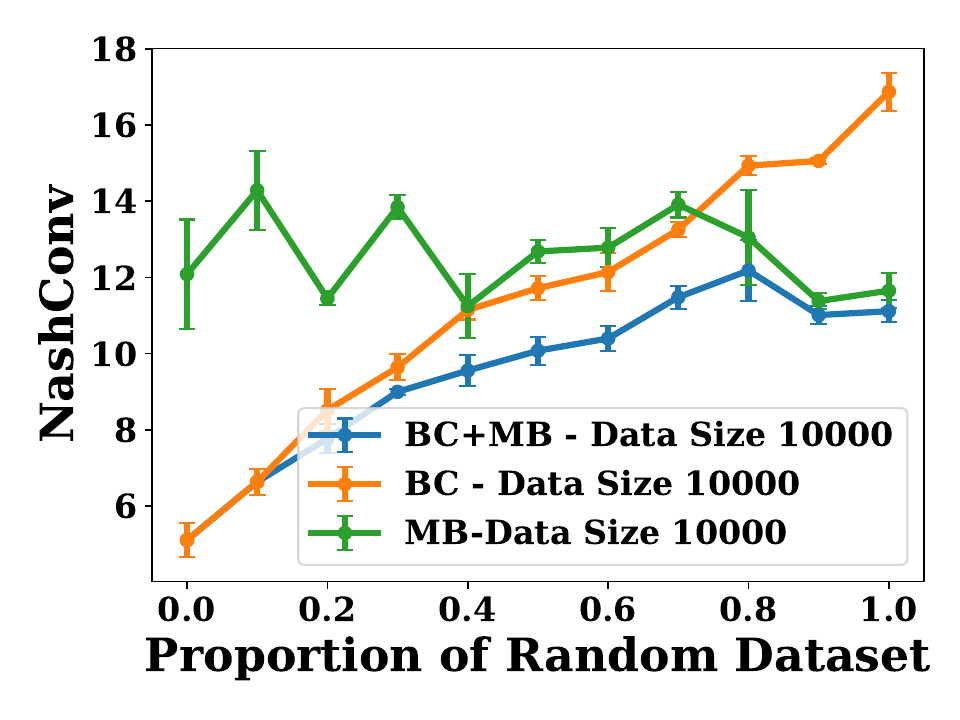}}
\subfigure[Leduc Poker-3 Player]{\includegraphics[width=0.23\textwidth]{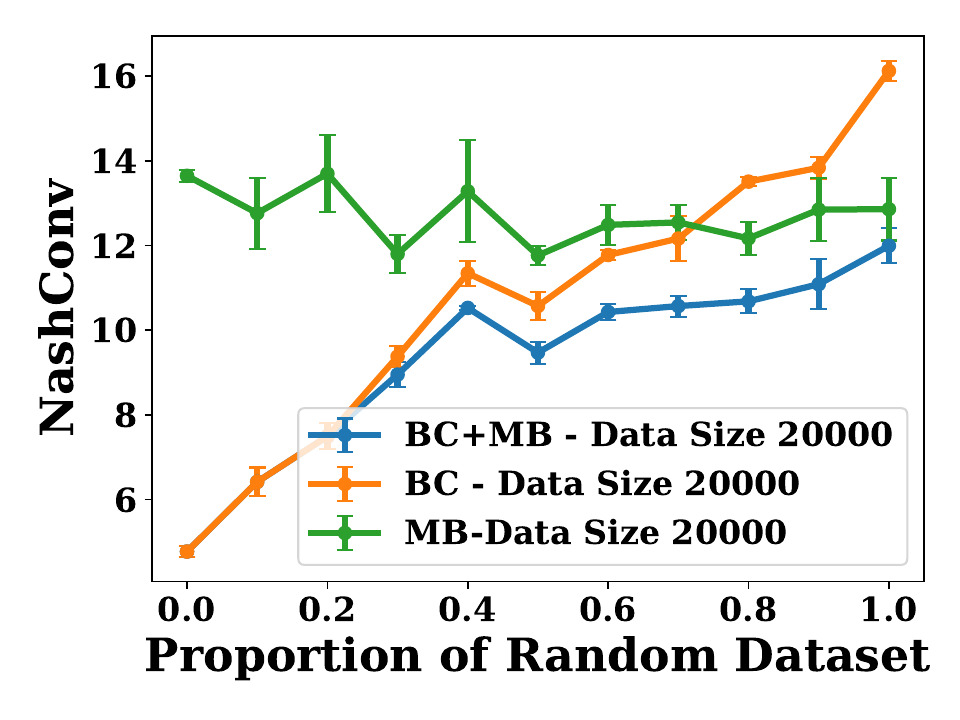}}
\subfigure[Liar's Dice-2 Player]{\includegraphics[width=0.30\textwidth]{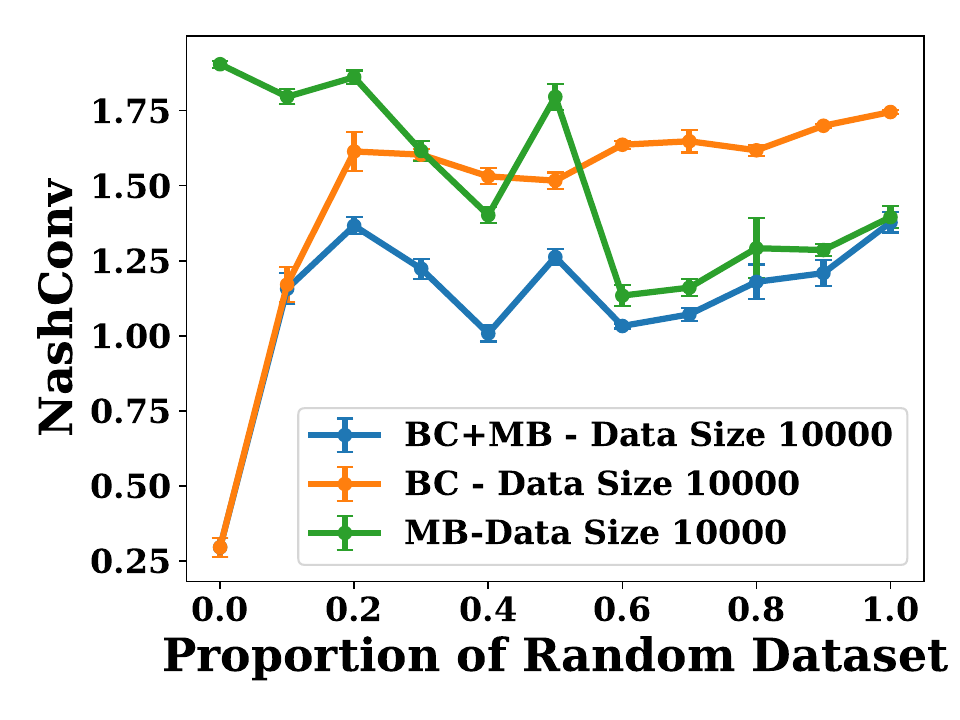}}
\subfigure[Liar's Dice-2 Player]{\includegraphics[width=0.30\textwidth]{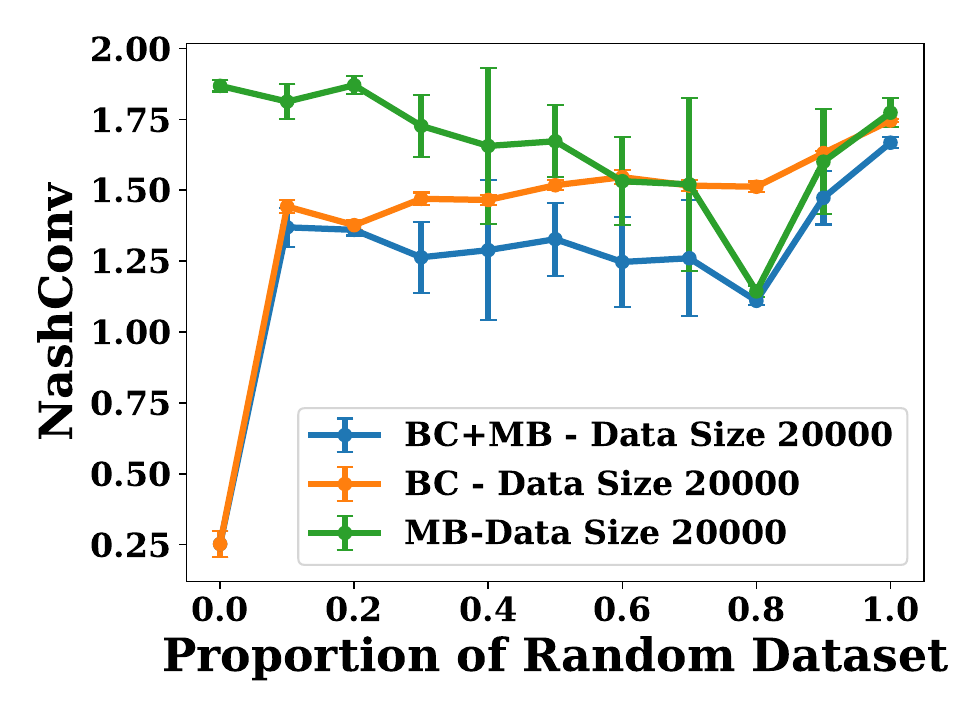}
\label{2}}
\subfigure[Phantom ttt-2 Player]{\includegraphics[width=0.32\textwidth]{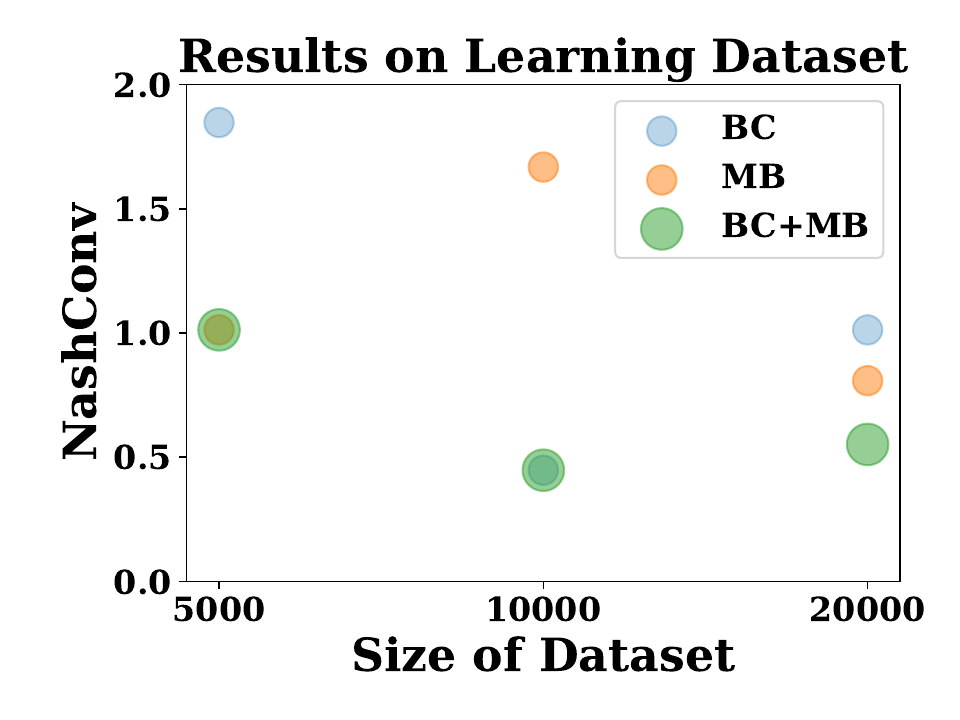}
\label{3}}
\caption{Experimental results for the benchmark algorithm BC+MB}
\label{bc_mb}
\end{figure}

\begin{figure}[ht]
\centering
\subfigure[Kuhn Poker-3 Player]{
\includegraphics[width=0.18\textwidth]{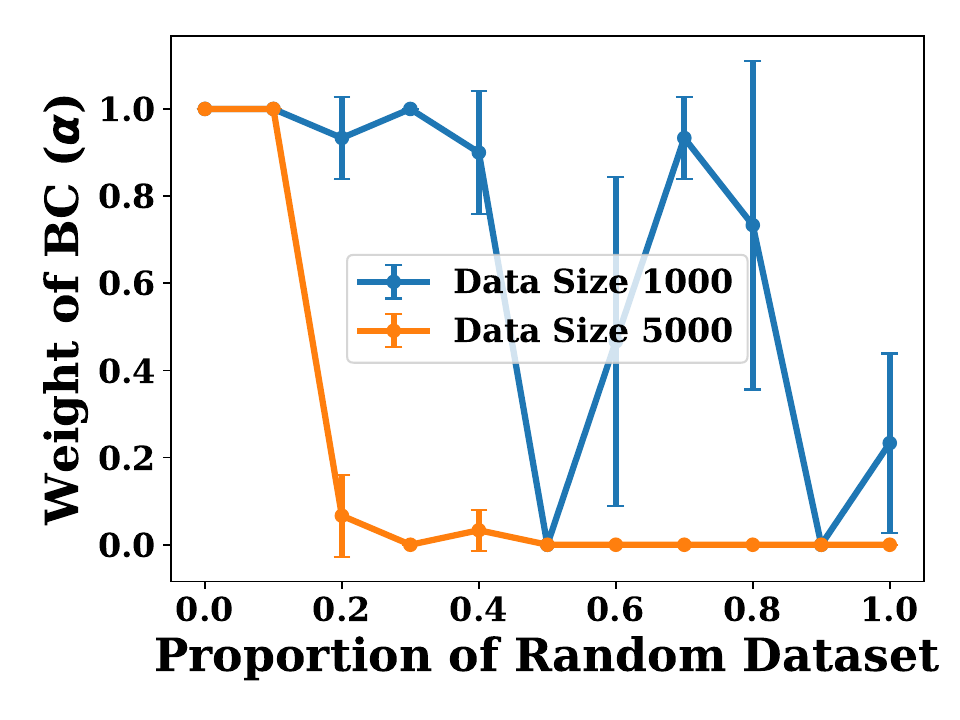}}
\subfigure[Kuhn Poker-4 Player]{
\includegraphics[width=0.18\textwidth]{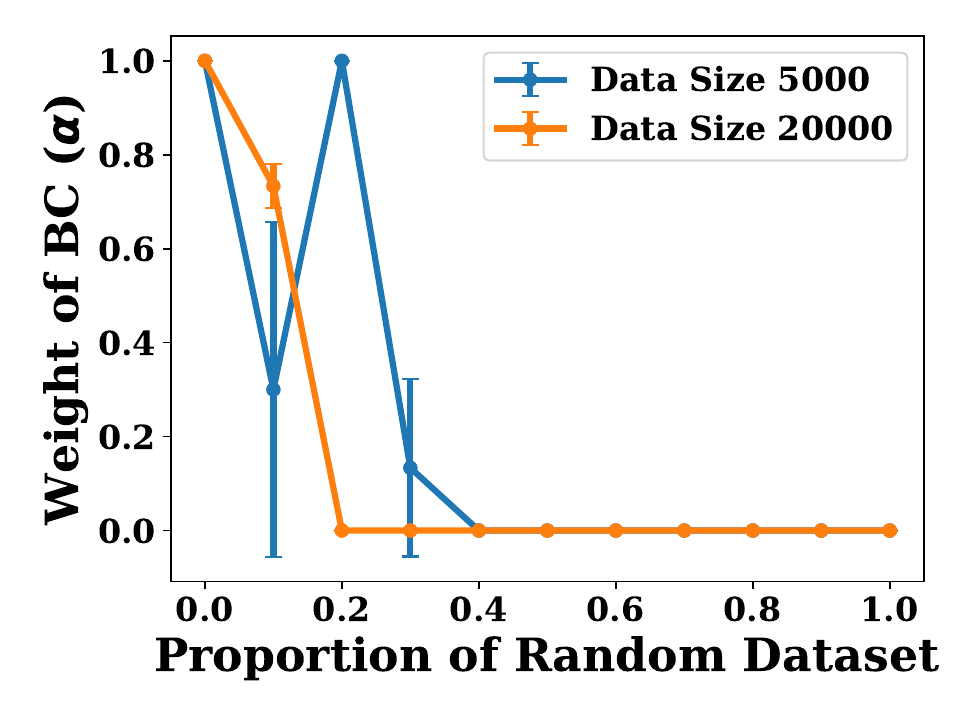}}
\subfigure[Kuhn Poker-5 Player]{
\includegraphics[width=0.18\textwidth]{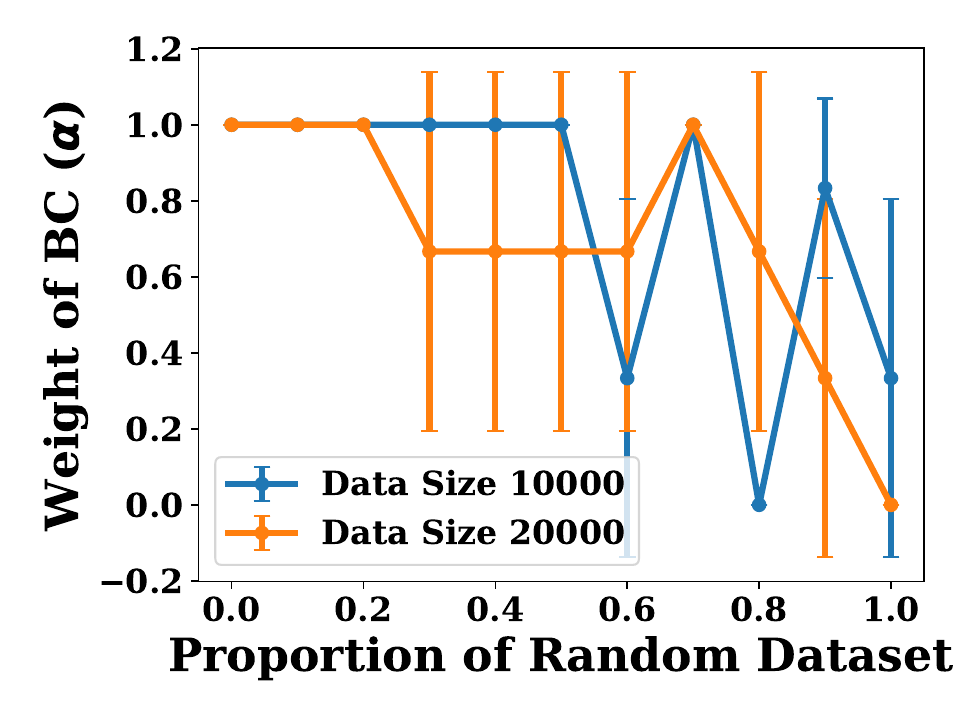}}
\subfigure[Leduc Poker-3 Player]{
\includegraphics[width=0.18\textwidth]{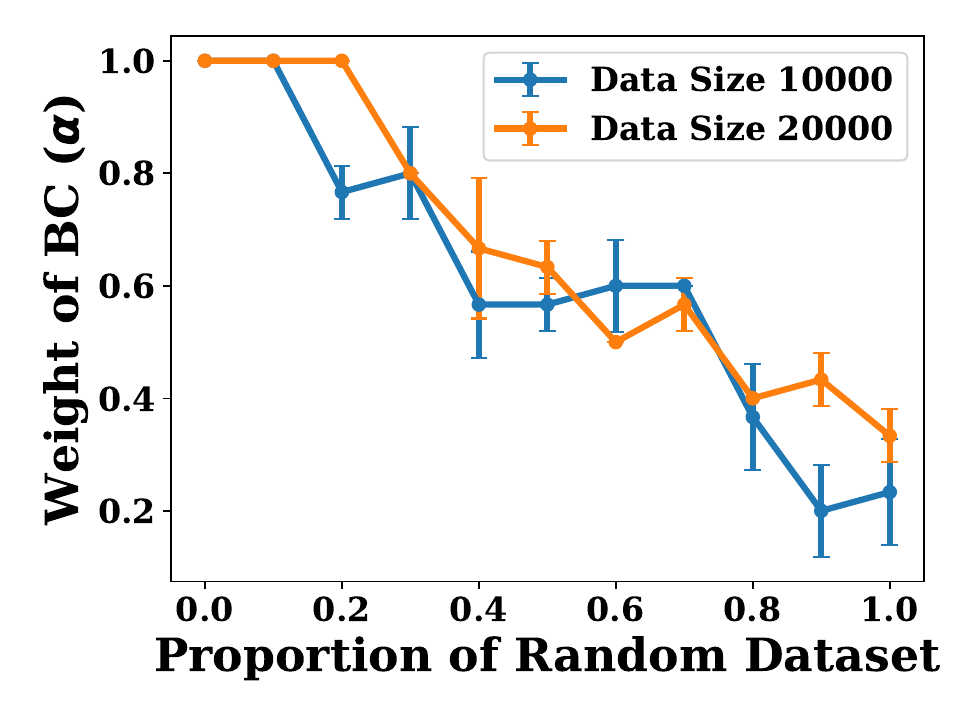}}
\subfigure[Liar's Dice-2 Player]{
\includegraphics[width=0.18\textwidth]{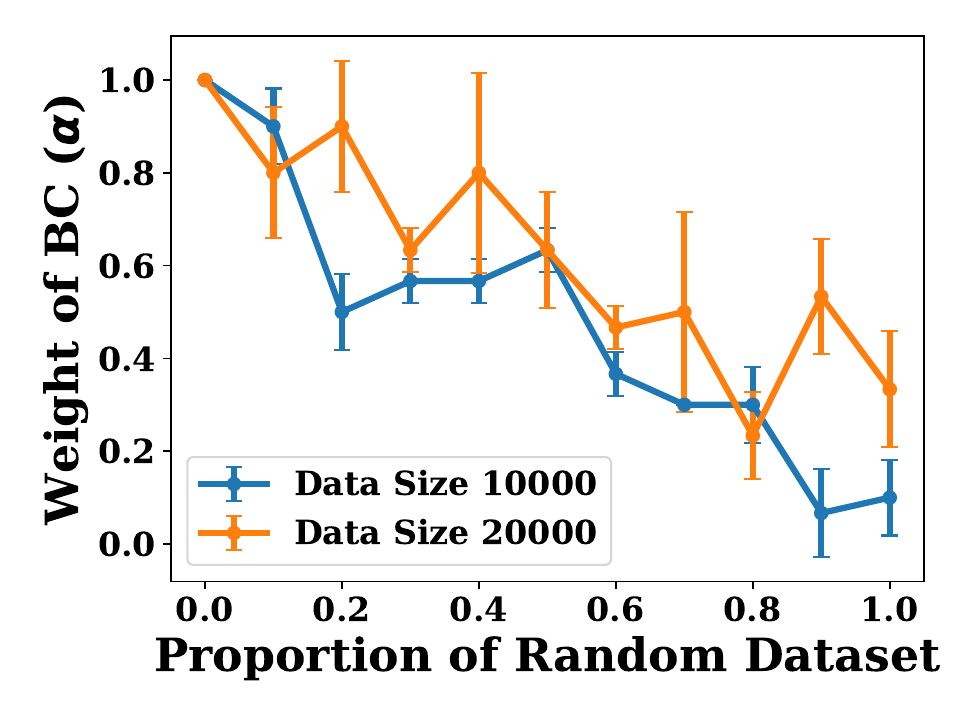}}
\caption{Experimental Results for Proper Weight}
\label{min_weight}
\end{figure}

The experimental results of the model-based framework (OEF-CFR) on these games are shown in Figure~\ref{iiiiii}. Since the strategy learned by OEF-CFR is not a joint strategy, we only use NashConv to measure how it is close to NEs in these multiple-player games. From these results, we found that the performance of the model-based framework is not stable in these games but still shows a slight decrease tendency with the increase of the proportion of the random dataset. Note that the CFR-based algorithm has no theoretical guarantee of convergence in multiple-player games. Therefore, OEF-CFR also can not guarantee to converge to the NE strategy. And the performance of the model-based framework also depends on the trained environment model. Therefore, the bad performance may be caused by the not well-trained environment model or the bad performance of the CFR-based algorithm in multiple-player games. Therefore, learning a good enough strategy is a big challenge in these multiple-player games under the OEF setting.

Figure~\ref{1}-\ref{2} show the experimental results of BC+MB on various games. 
We also test our OEF algorithm BC+MB in the Phantom Tic-Tac-Toe game based on the learning dataset (Figure \ref{3}). The NashConv values in Phantom Tic-Tac-Toe are approximate results since the best response policy is trained using DQN, and the utilities are obtained by simulation. It shows that the BC+MB performs better than BC and MB, which implies that our combination method can perform well in any game under any unknown dataset. The proper weights in the BC+MB algorithm under different datasets are shown in Figure~\ref{min_weight}. It shows a similar tendency as previous experiments.

\begin{figure}[h]
\centering
\subfigure[Kuhn-2 Player]{
\includegraphics[width=0.23\textwidth]{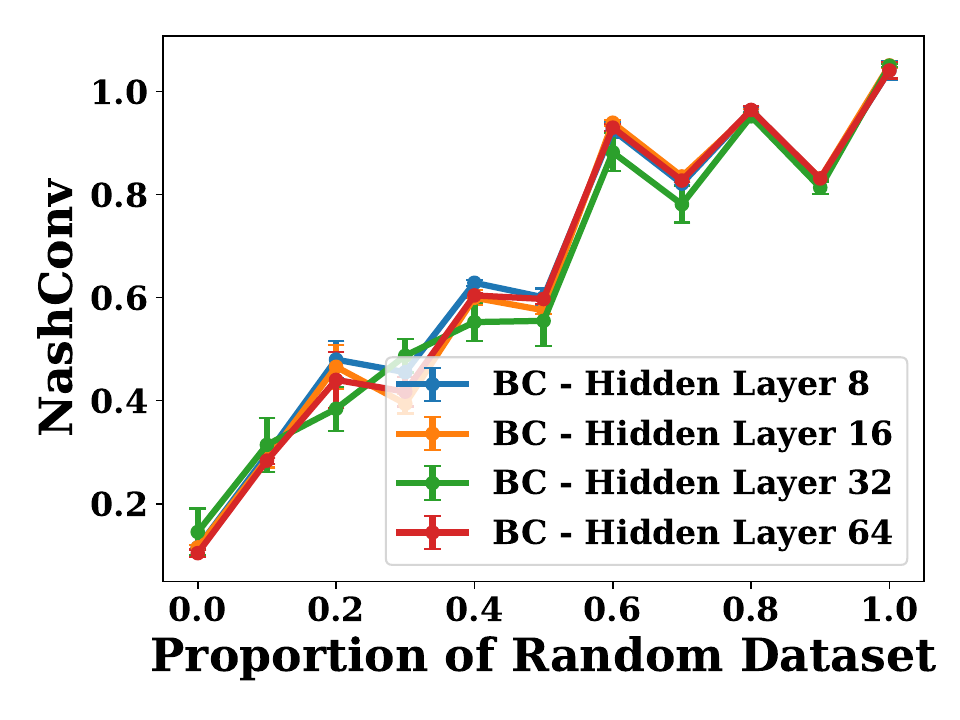}}
\subfigure[Kuhn-2 Player]{
\includegraphics[width=0.23\textwidth]{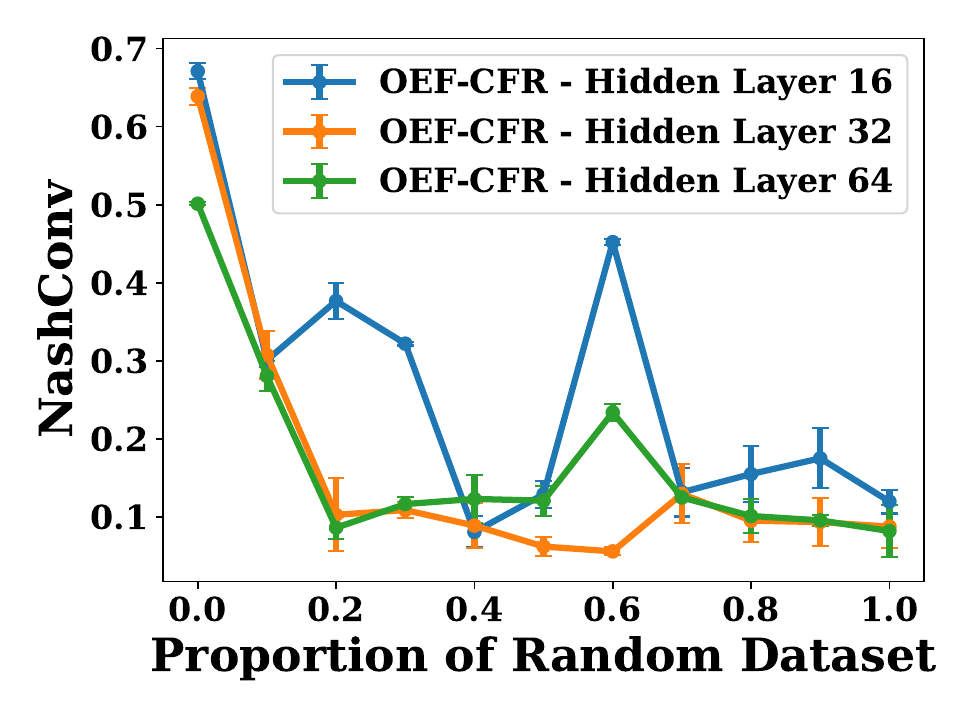}}
\subfigure[Leduc-2 Player]{
\includegraphics[width=0.23\textwidth]{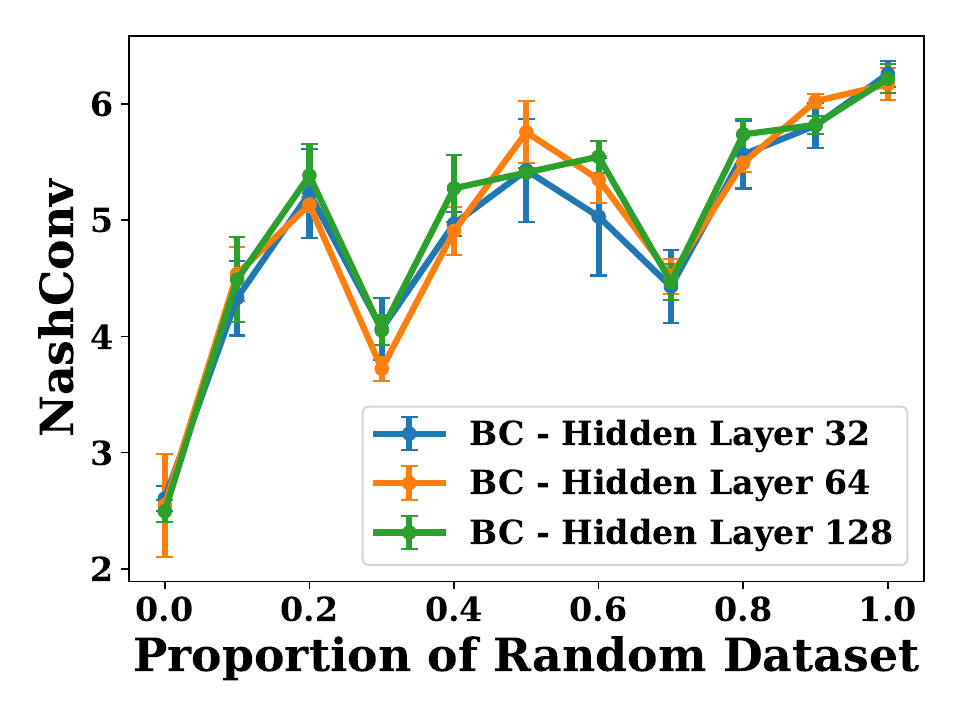}}
\subfigure[Leduc-2 Player]{
\includegraphics[width=0.23\textwidth]{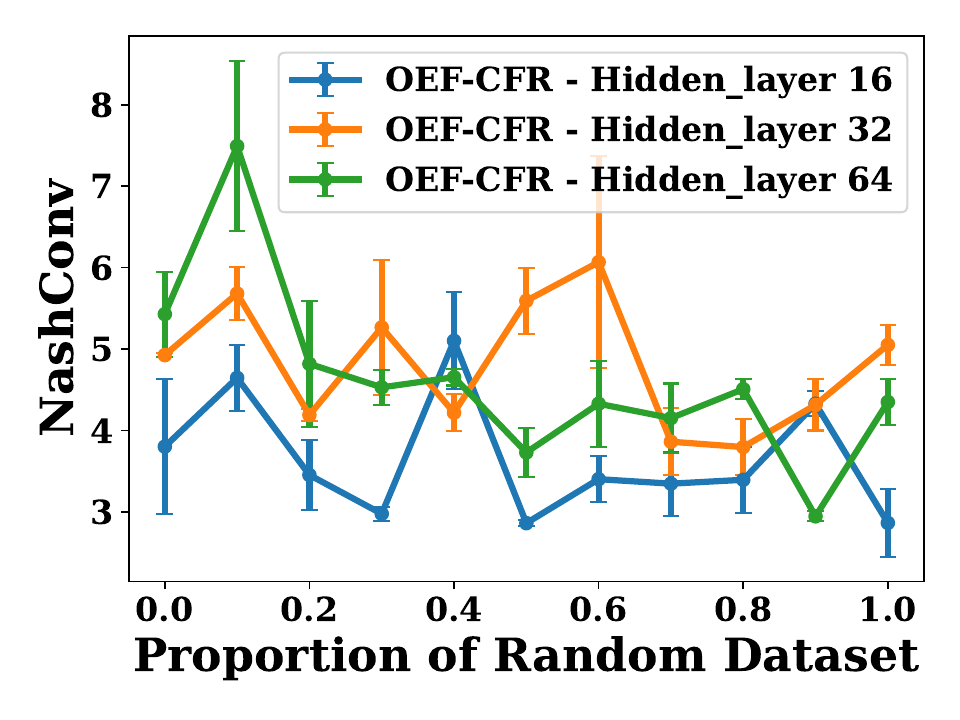}}
\caption{Ablation results for different hidden layer size}
\label{ablation}
\end{figure}

\begin{figure}[h]
\centering
\subfigure[Kuhn-2 Player]{
\includegraphics[width=0.23\textwidth]{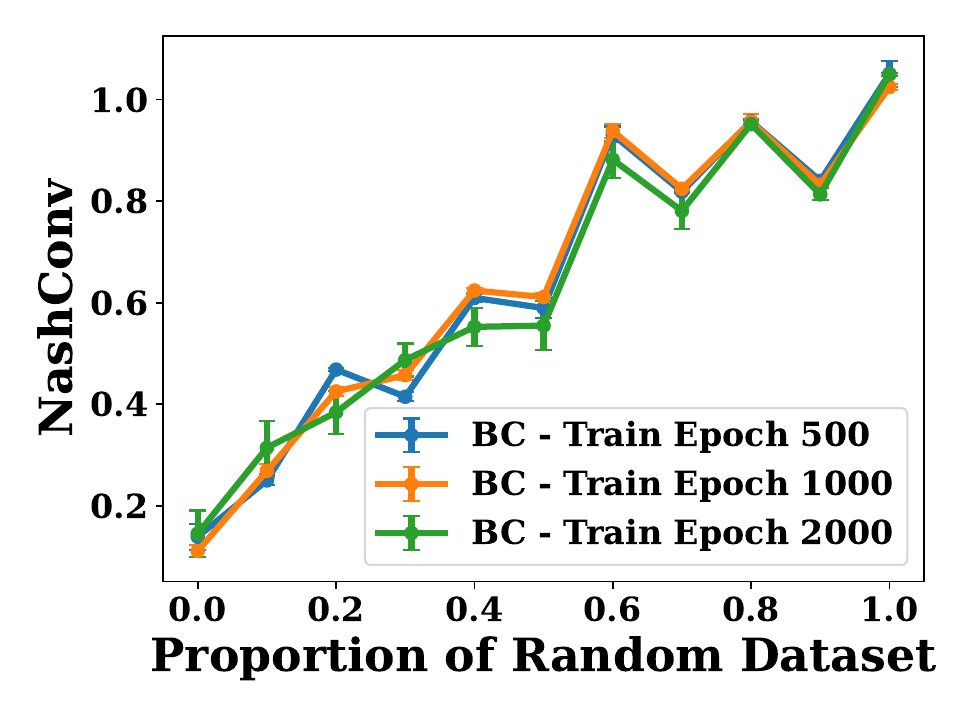}}
\subfigure[Kuhn-2 Player]{
\includegraphics[width=0.23\textwidth]{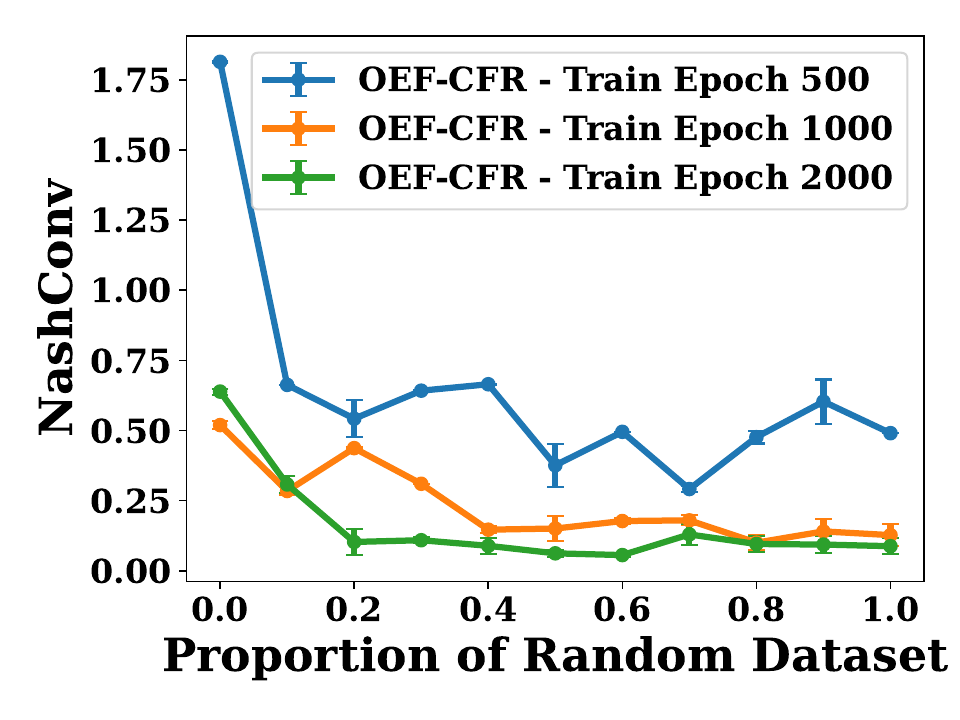}}
\subfigure[Leduc-2 Player]{
\includegraphics[width=0.23\textwidth]{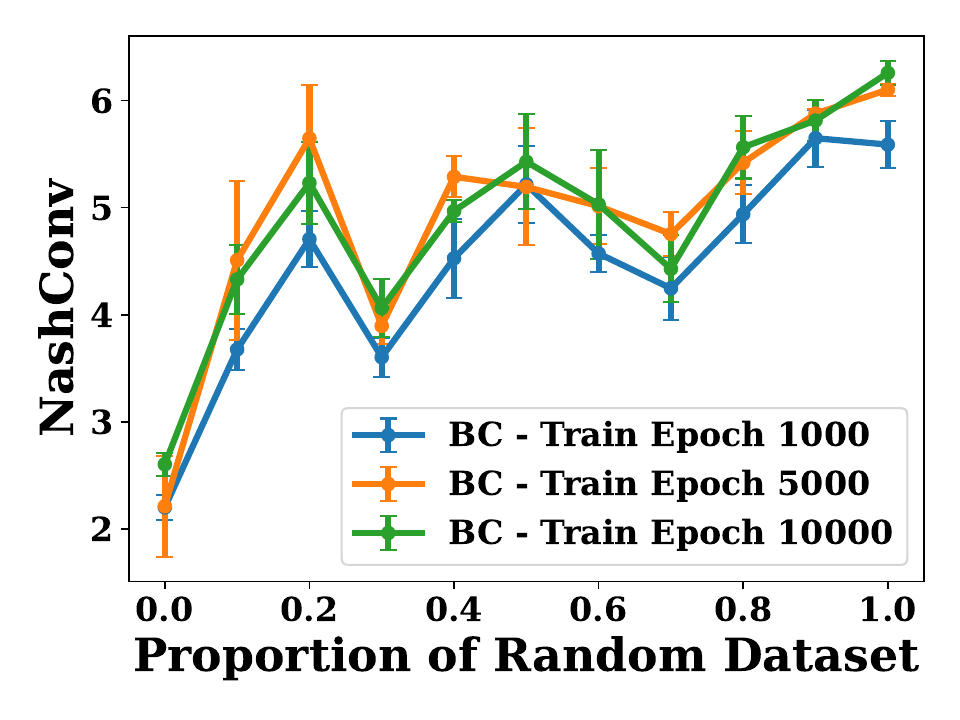}}
\subfigure[Leduc-2 Player]{
\includegraphics[width=0.23\textwidth]{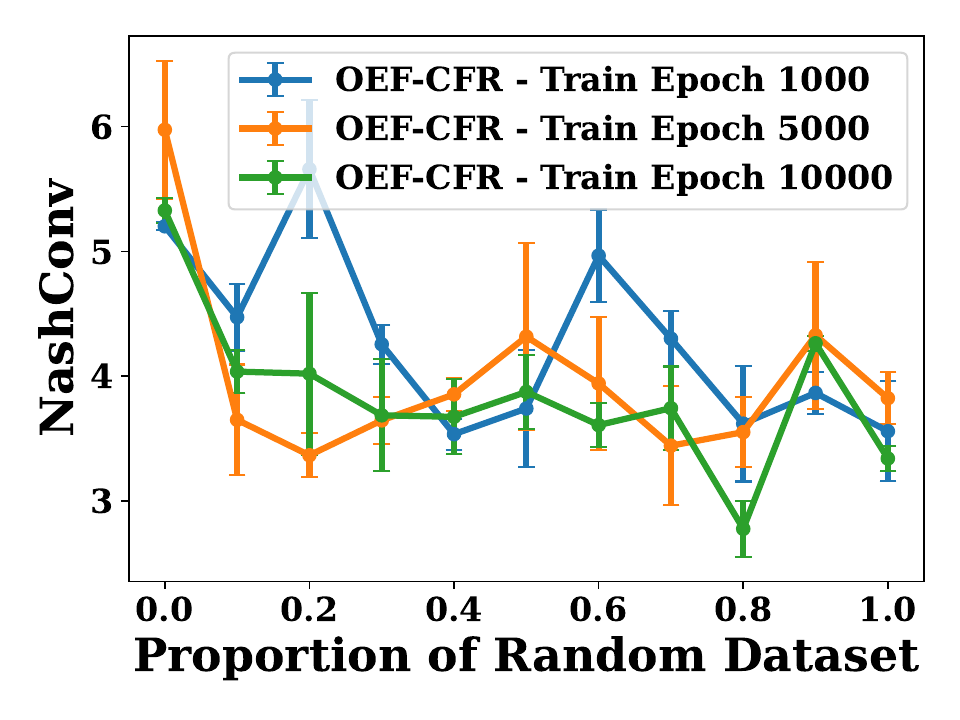}}
\caption{Ablation results for different train epoch}
\label{ablation_train}
\end{figure}

\clearpage
\textbf{Ablation Study.} To figure out the influence of hyperparameters, we conduct some ablation experiments on two-player Kuhn poker and Leduc poker games. 
We consider different model structures with various hidden layers. Specifically, for the 2-Player Kuhn poker game, we use different environment models with 8, 16, 32, and 64 hidden layers. For the 2-Player Leduc poker game, which is a more complicated game, the numbers of hidden layers for different models are 32, 64, and 128. 
Besides, we train the environment models for different epochs to evaluate the robustness of our approach.
Figures \ref{ablation}-\ref{ablation_train} show these ablation results. We can find that the hidden layer size and the number of training epochs have little effect on the performance of the BC algorithm. These results further verify that the performance of the BC algorithm mainly depends on the quality of the dataset. As we know that the performance of the model-based framework mainly depends on the trained environment model. Since the size of the hidden layer and the number of training epochs will influence the training phase of the environment model, the size of the hidden layer and the number of train epochs have a slight influence on the performance of the model-based framework. As long as the size of the hidden layer and the number of training epochs can guarantee that the environment model is well-trained, the performance of the model-based framework will not be affected.

\clearpage

\textbf{Parameter Setting.} We list the parameters used to train the behavior cloning policy and environment model for all games used in our experiments in Table \ref{tab:1} and Table \ref{tab:2}. 

\begin{table}[ht]
\centering
\caption{Parameters for Behavior Cloning algorithm}
\label{tab:1}
\begin{tabular}{c|ccccccccc}
\toprule
Games & Data size & Hidden layer & Batch size & Train epoch \\
\midrule
2-player Kuhn poker & 500  & 32 & 32 & 1000 \\
2-player Kuhn poker & 1000 & 32 & 32 & 2000 \\
2-player Kuhn poker & 5000  & 32 & 32 & 2000 \\
 \midrule
3-player Kuhn poker & 1000  & 32 & 32 & 5000 \\
3-player Kuhn poker & 5000  & 32 & 32 & 5000 \\
3-player Kuhn poker & 10000  & 64 & 128 & 5000 \\
 \midrule
4-player Kuhn poker & 5000  & 64 & 64 & 5000 \\
4-player Kuhn poker & 10000  & 64 & 128 & 5000 \\
4-player Kuhn poker & 20000  & 64 & 128 & 5000 \\
 \midrule
5-player Kuhn poker & 5000  & 64 & 64 & 5000 \\
5-player Kuhn poker & 10000  & 64 & 128 & 5000 \\
5-player Kuhn poker & 20000  & 64 & 128 & 5000 \\
 \midrule
2-player Leduc poker & 10000  & 128 & 128 & 10000 \\
2-player Leduc poker & 20000  & 128 & 128 & 10000 \\
2-player Leduc poker & 50000  & 128 & 128 & 10000 \\
 \midrule
3-player Leduc poker & 10000  & 128 & 128 & 10000 \\
3-player Leduc poker & 20000  & 128 & 128 & 10000 \\
3-player Leduc poker & 50000  & 128 & 128 & 10000 \\
 \midrule
Liar's Dice & 10000  & 64 & 64 & 5000 \\
Liar's Dice & 20000  & 64 & 128 & 5000 \\
Liar's Dice & 50000  & 64 & 128 & 5000 \\
 \midrule
Phantom Tic-Tac-Toe & 5000  & 128 & 128 & 5000 \\
Phantom Tic-Tac-Toe & 10000  & 128 & 128 & 5000 \\
Phantom Tic-Tac-Toe & 20000  & 128 & 128 & 5000 \\
\bottomrule
\end{tabular}
\end{table}

\begin{table}[t]
\centering
\caption{Parameters for training Environment Model}
\label{tab:2}
\begin{tabular}{c|ccccccccc}
\toprule
Games & Data size & Hidden layer & Batch size & Train epoch \\
 \midrule
2-player Kuhn poker & 500  & 32 & 32 & 1000 \\
2-player Kuhn poker & 1000 & 32 & 32 & 2000 \\
2-player Kuhn poker & 5000  & 32 & 32 & 2000 \\
 \midrule
3-player Kuhn poker & 1000  & 32 & 32 & 2000 \\
3-player Kuhn poker & 5000  & 32 & 32 & 5000 \\
3-player Kuhn poker & 10000  & 64 & 128 & 5000 \\
 \midrule
4-player Kuhn poker & 5000  & 64 & 64 & 5000 \\
4-player Kuhn poker & 10000  & 64 & 128 & 5000 \\
4-player Kuhn poker & 20000  & 64 & 128 & 5000 \\
 \midrule
5-player Kuhn poker & 5000  & 64 & 64 & 5000 \\
5-player Kuhn poker & 10000  & 64 & 128 & 5000 \\
5-player Kuhn poker & 20000  & 64 & 128 & 5000 \\
 \midrule
2-player Leduc poker & 5000  & 64 & 64 & 5000 \\
2-player Leduc poker & 10000  & 64 & 64 & 5000 \\
2-player Leduc poker & 20000  & 128 & 128 & 10000 \\
 \midrule
3-player Leduc poker & 10000  & 128 & 128 & 10000 \\
3-player Leduc poker & 20000  & 128 & 128 & 10000 \\
3-player Leduc poker & 50000  & 128 & 128 & 10000 \\
 \midrule
Liar's Dice & 10000  & 64 & 64 & 5000 \\
Liar's Dice & 20000  & 64 & 128 & 5000 \\
Liar's Dice & 50000  & 64 & 128 & 5000 \\
 \midrule
Phantom Tic-Tac-Toe & 5000  & 128 & 128 & 5000 \\
Phantom Tic-Tac-Toe & 10000  & 128 & 128 & 5000 \\
Phantom Tic-Tac-Toe & 20000  & 128 & 128 & 5000 \\
\bottomrule
\end{tabular}
\end{table}


\end{document}